\documentclass[11pt]{article}
\usepackage{authblk}
\pdfoutput=1
\usepackage{geometry}
\usepackage{amsmath,amssymb,mathrsfs}
\usepackage{color}
\usepackage{graphicx}
\usepackage{subfigure}
\usepackage{soul}
\usepackage{bbding}
\usepackage{lineno,hyperref}
\usepackage{multirow, booktabs, stmaryrd}
\modulolinenumbers[5]
\usepackage{threeparttable} 
\usepackage{latexsym}
\usepackage[T1]{fontenc}
\usepackage[utf8]{inputenc}
\usepackage{authblk}

\title{A Dimension-Augmented Physics-Informed Neural Network (DaPINN) with High Level Accuracy and Efficiency}
\author[1]{Weilong Guan}
\author[1]{Kaihan Yang}
\author[1]{Yinsheng Chen}
\author[1]{Zhong Guan\thanks{Corresponding author:guanzh23@mail.sysu.edu.cn}}
\affil[1]{School of Microelectronics Science and Technology, Sun Yat-Sen University, Zhuhai, 519082, China}
\date{}

\begin{document}

\maketitle

\begin{abstract}
Physics-informed neural networks (PINNs) have been widely applied in different fields due to their effectiveness in solving partial differential equations (PDEs). However, the accuracy and efficiency of PINNs need to be considerably improved for scientific and commercial use. To address this issue, we systematically propose a novel dimension-augmented physics-informed neural network (DaPINN), which simultaneously and significantly improves the accuracy and efficiency of the PINN. In the DaPINN model, we introduce inductive bias in the neural network to enhance network generalizability by adding a special regularization term to the loss function. Furthermore, we manipulate the network input dimension by inserting additional sample features and incorporating the expanded dimensionality in the loss function. Moreover, we verify the effectiveness of power series augmentation, Fourier series augmentation and replica augmentation, in both forward and backward problems. In most experiments, the error of DaPINN is 1$\sim$2 orders of magnitude lower than that of PINN. The results show that the DaPINN outperforms the original PINN in terms of both accuracy and efficiency with a reduced dependence on the number of sample points. We also discuss the complexity of the DaPINN and its compatibility with other methods.

\noindent {\it keywords}: PINN, Dimension-augmented, Data Enhancement, PDEs
\end{abstract}
\maketitle
\section{Introduction}
Partial differential equations (PDEs) are commonly used to describe various phenomena in science and engineering problems. PDE numerical solutions have received considerable attention from scholars, and many numerical methods have been proposed, including the spectral method \cite{javidi2006numerical}, variational iterative method \cite{moghimi2007variational}, and finite element method \cite{chernyshenko2015adaptive}. However, these numerical methods still face enormous challenges in solving inverse and high-dimensional problems \cite{ames2014numerical}. To overcome these issues, M. Razzi et al. \cite{dissanayake1994neural,raissi2019physics} proposed the physics-informed neural network (PINN), which introduces deep learning to numerical methods by using general approximations and powerful neural network representation.

The PINN has been applied in fluid mechanics \cite{chen2022flowdnn}, heat conduction \cite{cai2021physics}, materials \cite{jiang2021physics}, and wave fields \cite{shukla2020physics}. However, the accuracy and efficiency of the PINN need to be increased. Several aspects of the PINN can be enhanced, and many researchers have conducted work on this topic. The hyperparameter settings have been improved through sampling methods \cite{kharazmi2021hp}, loss functions \cite{bischof2021multi,wu2021modified,xiang2021self,yu2022gradient}, and activation functions \cite{jagtap2020adaptive}. Moreover, some related work has achieved improvements on high-dimensional disaster problems \cite{sirignano2018dgm} and inverse problems \cite{gorbachenko2016neural,jo2019deep,mishra2022estimates}. In terms of the network structure, specific boundary conditions \cite{lagaris1998artificial,mcfall2009artificial,dong2021method,lagari2020systematic} and feature information \cite{yazdani2020systems,cai2020phase,liu2020multi} have been incorporated. In terms of training methods, the residual-based adaptive refinement (RAR) method proposed in the literature \cite{lu2021deepxde} improves the accuracy by adding sample points in stages. Replacing the network input with Fourier basis functions to impose periodicity has been shown to be an effective optimization method for PDE systems with periodic boundary conditions \cite{lu2021physics,dong2021method}. However, this approach is applicable only in systems with periodic bounds and has not been generalized to more cases. Moreover, few studies have optimized neural networks from the perspective of dimension augmentation. 

The input dimension of the PINN usually depends on the number of equation variables, which is considerably lower than the number of variables in neural networks in other fields, such as natural language processing and computer vision. The input dimension of a network objectively affects the accuracy of the network. For example, in convolutional neural networks (CNNs), degradation in the image resolution reduces model performance \cite{kannojia2018effects}. Hence, the input dimension can be addressed by inserting more features into the input vector to enhance the PINN.

In this work, we propose a dimension-augmented PINN (DaPINN) that improves the accuracy and efficiency of the PINN by systematically manipulating the network input dimension, with the extended dimension bounded by a loss function that considers partial derivatives. The DaPINN model improves the solution accuracy by using power series augmentation and Fourier series augmentation and replica augmentation. Moreover, the DaPINN performs significantly better than the PINN, as the DaPINN requires less training time to achieve the same accuracy and has considerably higher accuracy under the same training conditions.

This paper is organized as follows. In Section 2, we introduce the PINN and DaPINN, as well as several strategies for extending the input dimension. In Section 3, we first prove the effectiveness of the proposed input dimension expansion approach and then show the DaPINN performance and the impact of different expansion methods on model accuracy for various positive and inverse problems. In Section 4, we discuss the computational complexity, network size implications, and method compatibility of DaPINN. Finally, we conclude the paper in Section 5.
\section{Methods}
\label{sec1}
\subsection{Physics-informed neural networks (PINN)} 	
The PINN is a neural network structure for solving partial differential equations that incorporates a network loss function based on the PDE , initial conditions and boundary conditions of the problem. To approximate the PDE solution $u(\mathbf{x})$, the PINN trains a neural network $\mathcal{N}(\mathbf{x})$ to minimize a loss function formulated according to the partial differential equation, initial conditions and boundary conditions.

Consider the following differential equation:
\begin{equation}
f\left(x,t,\mathcal{N},\partial_x\mathcal{N},\partial_t\mathcal{N}\ldots,\lambda\right)=0,\ \ x\in\mathrm{\Omega},\ t\in\left[0,T\right]
\end{equation}
\begin{equation}
\mathcal{N}\left(x,0\right)=g_0\left(x\right),\ \ x\in\mathrm{\Omega}
\end{equation}
\begin{equation}
\mathcal{N}\left(x,t\right)=g_\mathrm{\Gamma}\left(t\right),x\in\partial\Omega,t\in\left[0,T\right]
\end{equation}
where $\emph{x}$ is a spatial coordinate; $\emph{t}$ is the time; $\emph{f}$ denotes the residual of the partial differential equation, including the differential operator $([\partial x_u,\partial t_u,\ldots])$ and the parameter $(\boldsymbol{\lambda}=[\lambda_1 ,\lambda_2 ,\ldots])$; $u(x,t)$ is the solution of the PDE with the given initial boundary value condition, where the initial value condition is $g_0(x)$, and the boundary value condition is $g_\Gamma\left(t\right)$ (can be type I, type II, or type III boundary conditions); $\Omega$ and $\partial\Omega$ denote the spatial domain and boundary, respectively. The PINN uses a fully connected feedforward neural network (FNN) with multiple hidden layers to approximate the solution of the PDE $\emph{u}$. The spatial and temporal coordinates (x,t) are used as inputs. The hidden layer parameters of the $k^{th}$ layer are denoted as $z^k$. A neural network with depth of L can be represented as:
\begin{equation}
z^0=\left(x,t\right)
\end{equation}
\begin{equation}
z^k=\sigma\left(W^kz^{k-1}+b^k\right),1\le k\le L-1
\end{equation}
\begin{equation}
z^k=W^kz^{k-1}+b^k,k=L
\end{equation}

The PINN formulates the solution of the PDE system as an optimization problem by iterating the neural network weights. Our objective is to minimize the loss function $\mathcal{L}$ by optimizing trainable parameters $\boldsymbol{\theta}$:

\begin{equation}
\mathcal{L}=\omega_f\mathcal{L}_f\left(\boldsymbol{\theta},\boldsymbol{\lambda};\mathcal{T}_f\right)+\omega_b\mathcal{L}_b\left(\boldsymbol{\theta},\boldsymbol{\lambda};\mathcal{T}_b\right)+\omega_i\mathcal{L}_i\left(\boldsymbol{\theta},\boldsymbol{\lambda};\mathcal{T}_i\right)
\end{equation}
In the above equation, $\mathcal{L}_f$, $\mathcal{L}_b$ and $\mathcal{L}_i$ are the residuals of the differential equation,  initial conditions and boundary conditions, respectively. $\mathcal{T}_f$, $\mathcal{T}_b$ and $\mathcal{T}_i$ are the sample points in the domain, initial state and on boundary. $\omega_1$, $\omega_2$, and $\omega_3$ are the weight coefficients corresponding to the different loss terms.
\begin{figure}[htbp]
  \centering
  \subfigure[]{
  \begin{minipage}[t]{0.5\linewidth}
  \centering
  \includegraphics[width=80mm]{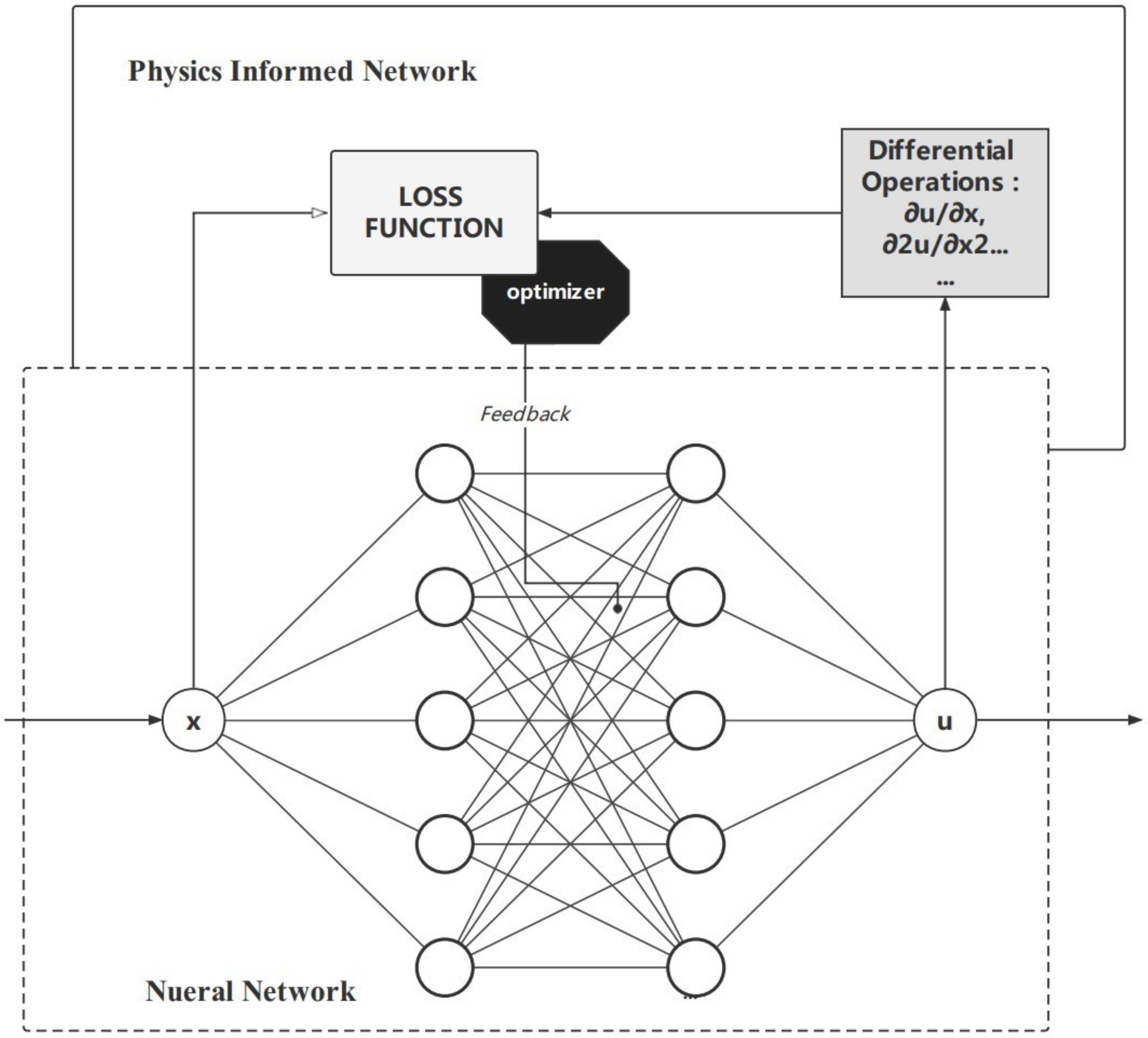}
  \end{minipage}%
  }%
  \subfigure[]{
  \begin{minipage}[t]{0.5\linewidth}
  \centering
  \includegraphics[width=80mm]{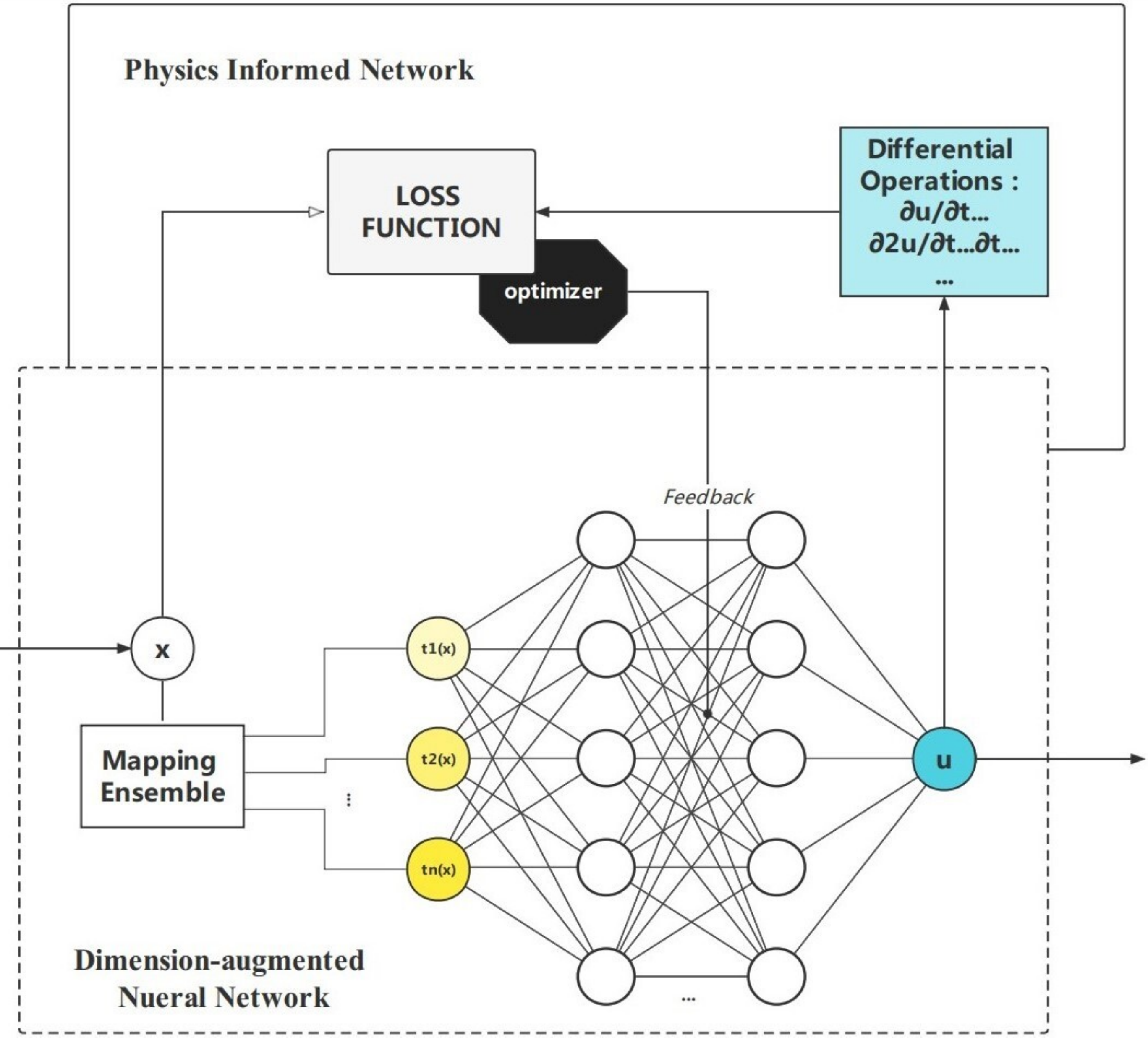}
  \end{minipage}%
  }%
  \centering
  \caption{ (a) Structure of conventional physics-informed neural networks (PINN). (b) Structure of dimension-augmented physics-informed neural networks (DaPINN).}
\end{figure}
\subsection{Dimension-augmentated PINN} 	
In PINNs, we only use the physical coordinate $\mathbf{x}$ as the input to construct the neural network $\mathcal{N}\left(\mathbf{x}\right)$ to fit the objective function $u\left(\mathbf{x}\right)$. Here, we introduce a set of mappings:
\begin{equation}
t_1,\ t_2\ldots t_n: X \rightarrow T_1,T_2, \ldots,T_{n}; \tau_{i}=t_i\left(x\right),\ i=1,\ldots,n.
\end{equation}

Then, the same method can be applied with $\tau_1,\tau_2, \ldots,\tau_{n}$, the images of $\mathbf{x}$ under mapping functions $t_1,t_2, \ldots,t_{n}$, used as the input to construct a new neural network $\mathcal{N}\left(\tau_1,\tau_2\ldots \tau_{n}\right)$. The new neural network has a input dimension of n, compared to PINN where the input dimension is only related to the size of $\mathbf{x}$. If the appropriate mapping is chosen, the new network, which is denoted as $\mathcal{N}\left(\boldsymbol{\tau}\right)=\mathcal{N}\left(t_1\left(x\right),\ldots,t_n\left(x\right)\right)$ can also approximate the original objective function $u\left(x\right)$ after iterating (Fig. 1.).

The loss function of the DaPINN model is defined as equation (7), where the residual term and the boundary/initial error in the PDE can be re-expressed as
\begin{equation}
\mathcal{L}_f\left(\boldsymbol{\theta},\boldsymbol{\lambda};\mathcal{T}_f\right)  =  \frac{1}{{\left|\mathcal{T}_f\right|}}\times{\sum_{\mathbf{x}\in \mathcal{T}_f}\left|f\left(\tau_1,\tau_2,\ldots;\frac{\partial\mathcal{N}}{\partial \tau_1},\frac{\partial\mathcal{N}}{\partial \tau_2},\ldots;\frac{\partial^2\mathcal{N}}{\partial \tau_1\partial \tau_1},\frac{\partial^2\mathcal{N}}{\partial \tau_1\partial \tau_2},\ldots;\ldots;\boldsymbol{\lambda}\right)\right|^2}
\end{equation}
\begin{equation}
\mathcal{L}_{b,i}\left(\boldsymbol{\theta},\boldsymbol{\lambda};\mathcal{T}_{b,i}\right)  = \frac{1}{\left|\mathcal{T}_{b,i}\right|}\times\sum_{\mathbf{x}\in T_{b,i}}\left|\mathcal{N}\left(\tau_1,\tau_2,\ldots,\tau_n;\boldsymbol{\theta}\right)-u\left(\tau_1,\tau_2,\ldots,\tau_n;\boldsymbol{\theta}\right)\right|^2
\end{equation}
For example, for the 1D Laplace equation: $\Delta u(x)=0$, by using mappings $t_1:\ x\rightarrow x,\ t_2:\ x\rightarrow x^2$, a two-input neural network can be constructed as $\mathcal{N}(\tau_1,\tau_2)$. It is straightforward to express the residual term of the differential equation as
\begin{multline}
\mathcal{L}_f=\frac{1}{{\left|\mathcal{T}_{f}\right|}}\sum_{\mathbf{x}\in T_{f}}{\Delta \mathcal{N}\left(\tau_1,\tau_2\right)}
\\
=\frac{1}{{\left|\mathcal{T}_{f}\right|}}\sum_{\mathbf{x}\in \mathcal{T}_f}\left|\left(\frac{\partial^2\mathcal{N}\left(\tau_1,\tau_2\right)}{\partial{\tau_1}^2}+4\tau_1\frac{\partial^2\mathcal{N}\left(\tau_1,\tau_2\right)}{\partial \tau_1\partial \tau_2}+4\tau_2\frac{\partial^2\mathcal{N}\left(\tau_1,\tau_2\right)}{{\partial \tau_2}^2}+2\frac{\partial\mathcal{N}\left(\tau_1,\tau_2\right)}{{\partial \tau_2}}\right)\right|^2
\end{multline}
The residual terms $\mathcal{L}_{b}$ and $\mathcal{L}_{i}$ of the boundary and initial conditions are consistent with those of the conventional PINN.

On the one hand, this dimension augmentation approach provides a way to optimize the topology of PINNs, making the width of the input and hidden layers more balanced. On the other hand, DaPINN directly introduces physical coordinate features into the neural network through the input layer rather than allowing the neural network to obtain these features. With the same network structure, the dimension augmentation method theoretically increases the input layer width and thus the complexity of the network, which guarantees a higher solution accuracy \cite{raghu2017expressive,pascanu2013number}. In addition, the DaPINN training process is simpler and faster than the training process in traditional methods since some features are provided as input.
From a higher-level perspective, we introduce an inductive bias to the neural network by adding a special regularization term to the loss function, thus enhancing network generalizability.
\subsection{Several input dimension augmentation methods} 	

There are several typical methods for augmenting DaPINNs:

1. Power series augmentation: As previously mentioned, power series augmentation increases the dimensionality of the input quantity with a power series transformation. $t_1:\ x\rightarrow x,\ t_2:\ x\rightarrow x^2\ldots$ This idea is derived from the Taylor expansion of the function:
\begin{equation}
f\left(\mathbf{x}\right)=f\left(\mathbf{x_k}\right)+\left[\mathrm{\nabla f}\left(\mathbf{x_k}\right)\right]^T\left(\mathbf{x}-\mathbf{x_k}\right)+\frac{1}{2}\left[\mathbf{x}-\mathbf{x_k}\right]^TH\left(\mathbf{x_{k^\prime}}\right)\left[\mathbf{x}-\mathbf{x_k}\right]\ ...
\end{equation}
where $H\left(x\right)$ is the Hessian matrix of the function.

2. Fourier series augmentation: Another possible dimensional expansion method is Fourier series augmentation, i.e., introducing the mappings $t_1:x\rightarrow x$,\ $t_2:x\rightarrow sin\left(\frac{2\pi nx}{T} \right)$,\ $t_3:x\rightarrow cos\left(\frac{2\pi nx}{T}\right)$, yielding the neural network $\mathcal{N}\left(x,sin(\frac{2\pi nx}{T}),cos(\frac{2\pi nx}{T})\right)$ (where T depends on the maximum period of the periodic function in the problem). Since any periodic function can be written as a sum of trigonometric functions, the function can be approximated by a trigonometric series through the Fourier series augmentation formula. The Fourier series augmentation method can be used in problems with periodic boundary conditions.

3. Replica augmentation: Simple copy-paste is a powerful data augmentation method \cite{ghiasi2021simple} that introduces the mapping $t_1:\ x\rightarrow x,\ t_2:\ x\rightarrow x$, yielding the neural network $\mathcal{N}\left(x_1,x_2\right)$ . This expansion method introduces mirrored inputs to provide multiple paths for network approximation, thereby improving model accuracy.
\section{Results}
We present several examples of solving forward and inverse PDE problems using the proposed DaPINN model, in which tanh function is used to activate the network and Adam algorithm is used to train it. In Section 3.1, we use replica and second-order power series augmentation to enhance the PINN input vector to verify the validity of the input augmentation approach. In Section 3.2, we show the effectiveness of our method in solving inverse problems. In Section 3.3, we discuss the high order power series augmentation method. Finally, in Section 3.4, we apply a DaPINN with Fourier series augmentation.

\subsection{Validity of DaPINN} 	
We first demonstrate the validity of the dimension augmentation approaches using a simple 1D Poisson equation and a pedagogical example of a 2D Poisson equation.

In these two problems, we use replica augmentation and second-order power series augmentation.
\subsubsection{1D Poisson equation}
We first consider the following 1D Poisson equation:
\begin{equation}
-\Delta u=\sum_{i=1}^{3}isin{ix}+7sin{7x}+8sin{8x} , x\epsilon\left[0,\pi\right]
\end{equation}
with Dirichlet boundary conditions, namely, $u\left(x=0\right)=0$ and $u\left(x=\pi\right)=\pi $. The analytical solutions is
\begin{equation}
u\left(x\right)=x+\sum_{i=1}^{3}\frac{sinix}{i}+\frac{sin{7x}}{7}+\frac{sin{8x}}{8}
\end{equation}
We construct the neural network N(x), expand the input and change the loss function as follows:
\begin{equation}
\mathcal{L}=\omega_{f}\mathcal{L}_{f}+\omega_{b}\mathcal{L}_{b}
\end{equation}
For replica augmentation, which introduces $\tau_1=\tau_2=x$, $\mathcal{L}_{f}$ and $\mathcal{L}_{b}$ are defined as

\begin{multline}
\mathcal{L}_f=\frac{1}{{\left|\mathcal{T}_f\right|}}\sum_{x\in \mathcal{T}_f}|(\frac{\partial^2\mathcal{N}\left(\tau_1,\tau_2\right)}{\partial{\tau_1}^2}+2\frac{\partial^2\mathcal{N}\left(\tau_1,\tau_2\right)}{\partial \tau_1\partial \tau_2}+\frac{\partial^2\mathcal{N}\left(\tau_1,\tau_2\right)}{{\partial \tau_2}^2}\\
+\sum_{i=1}^{3}isin{ix}+7sin{7x}+8sin{8x})|^2
\end{multline}
\begin{equation}
\mathcal{L}_b=\frac{1}{{\left|\mathcal{T}_b\right|}}\sum_{x\in \mathcal{T}_b}\left|\mathcal{N}\left(\tau_1,\tau_2\right)\right|^2
\end{equation}
For second-order power series augmentation, which introduces  $\tau_1=x, \tau_2=x^2$, $L_{f}$ and $L_{b}$ are defined as

\begin{multline}
\mathcal{L}_f=\frac{1}{{\left|\mathcal{T}_f\right|}}\sum_{x\in \mathcal{T}_f}|(\frac{\partial^2\mathcal{N}(\tau_1,\tau_2)}{\partial \tau_2}+4x\frac{\partial^2\mathcal{N}(\tau_1,\tau_2)}{\partial x\partial x^2}+4x^2\frac{\partial^2\mathcal{N}(\tau_1,\tau_2)}{{\partial \tau_2}^2}+2\frac{\partial\mathcal{N}(\tau_1,\tau_2)}{\partial \tau_2}\\
+\sum_{i=1}^{3}isin{ix}+7sin{7x}+8sin{8x})|^2
\end{multline}
\begin{equation}
\mathcal{L}_b=\frac{1}{{\left|\mathcal{T}_b\right|}}\sum_{x\in \mathcal{T}_b}\left|\mathcal{N}\left(\tau_1,\tau_2\right)\right|^2
\end{equation}
Here, we set $\omega_{f}=\omega_{b}=1$

We compare the performance of DaPINN and PINN with respect to the number of training points and the number of epochs in this experiment, and the results are shown in Fig. 2(a) and (b). When the number of sample points is increased from 11 to 23, the L2 relative error decreases from 12.1\% to 4.0\% with the PINN method, while the L2 relative error of the DaPINN method decreases from 13.2\% to 0.38\% using replica augmentation($x$) and 13.5\% to 0.2\% using second-order power series augmentation($x^2$). At 32 training points, the L2 relative error of the PINN method decreases from 31\% to 0.75\% when the number of training epochs is increased from 2000 to 15000, while the L2 relative error of the DaPINN method decreases from 24\% to 0.09\% with the x approach and 4.6\% to 0.04\% with $x^2$. 

We find that the DaPINN with x and the DaPINN with $x^2$ both have significantly smaller L2 relative errors than the PINN. The DaPINN models outperform the PINN because the DaPINNs expands the network input dimension, thereby allowing the neural network to extract features with more information, resulting in a substantially faster convergence rate than PINN. When 26 sample points were used, the results of the PINN and two DaPINNs are shown in Fig. 2(c). The DaPINN with $x^2$ model has better accuracy than the DaPINN with x model in this problem.

We note that the PINN method has a poor fit in the nonlinear region of the function, while the DaPINN with second-order power series augmentation: approach achieves good results because $x^2$ was introduced as an augmentation feature, which enhances the fitting in the nonlinear region.
\begin{figure}[htbp]
\centering
\subfigure[]{
\begin{minipage}[t]{0.5\linewidth}
\centering
\includegraphics[width=60mm]{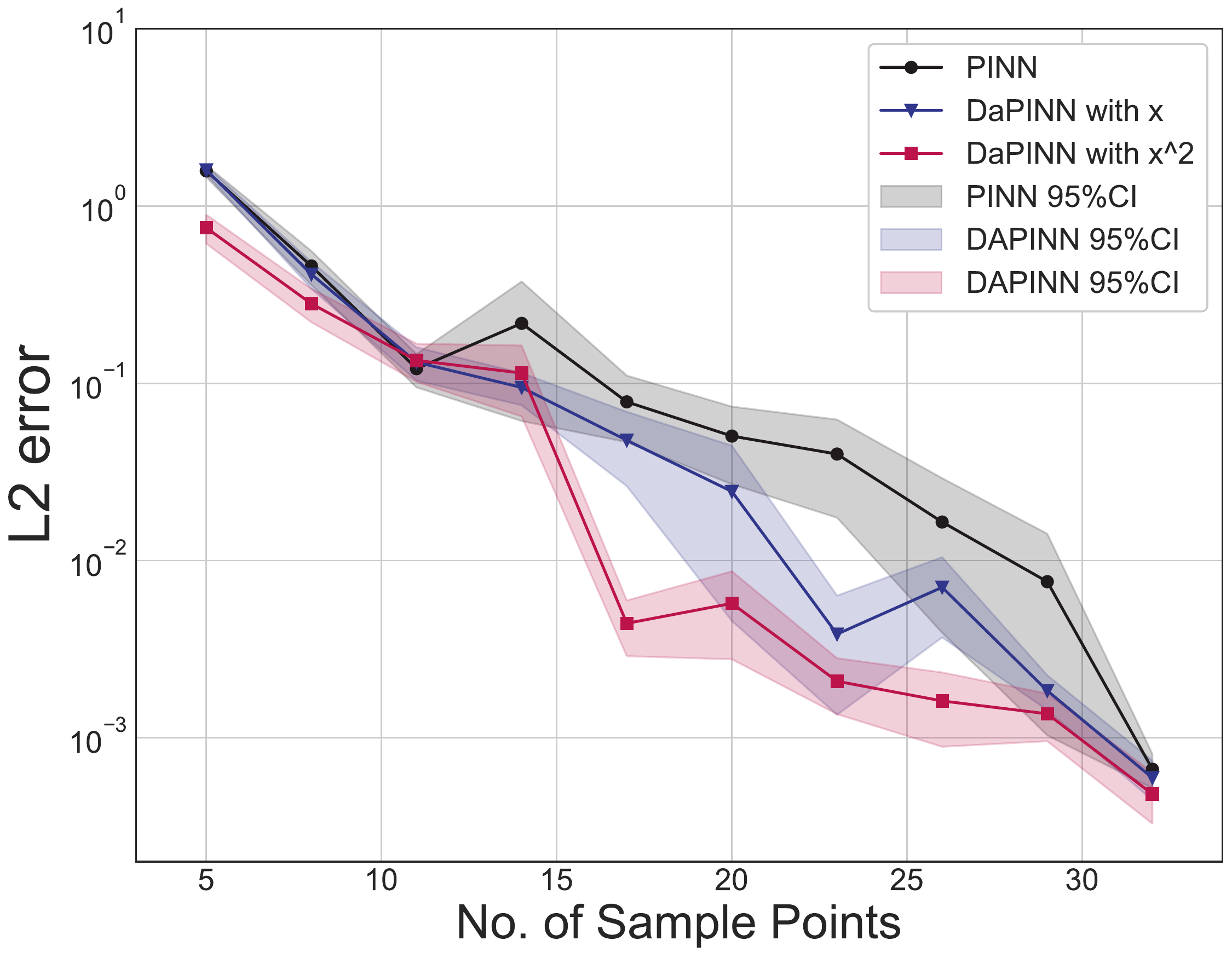}
\end{minipage}%
}%
\subfigure[]{
\begin{minipage}[t]{0.5\linewidth}
\centering
\includegraphics[width=60mm]{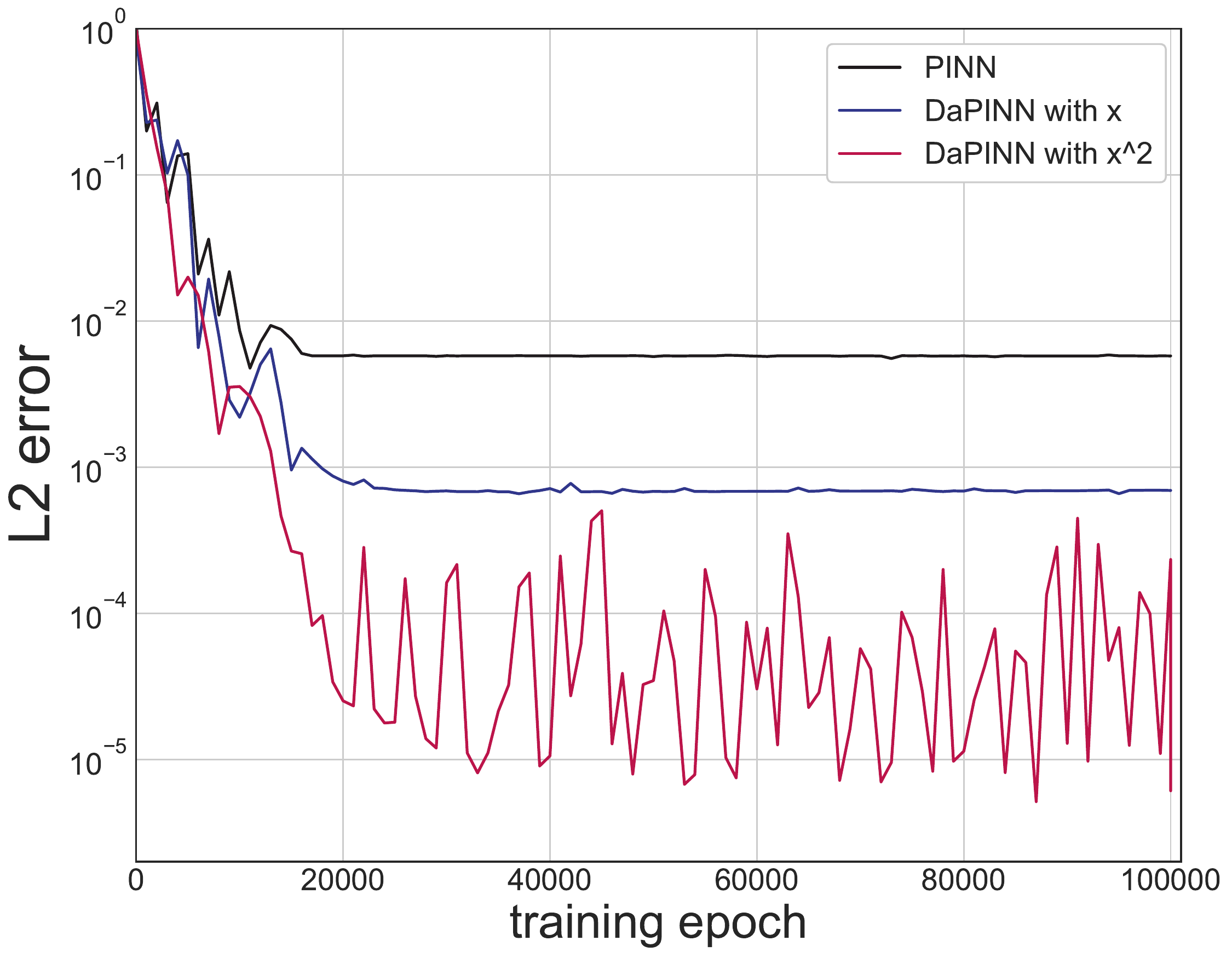}
\end{minipage}%
}%

\subfigure[]{
\begin{minipage}[t]{1\linewidth}
\centering
\includegraphics[width=100mm]{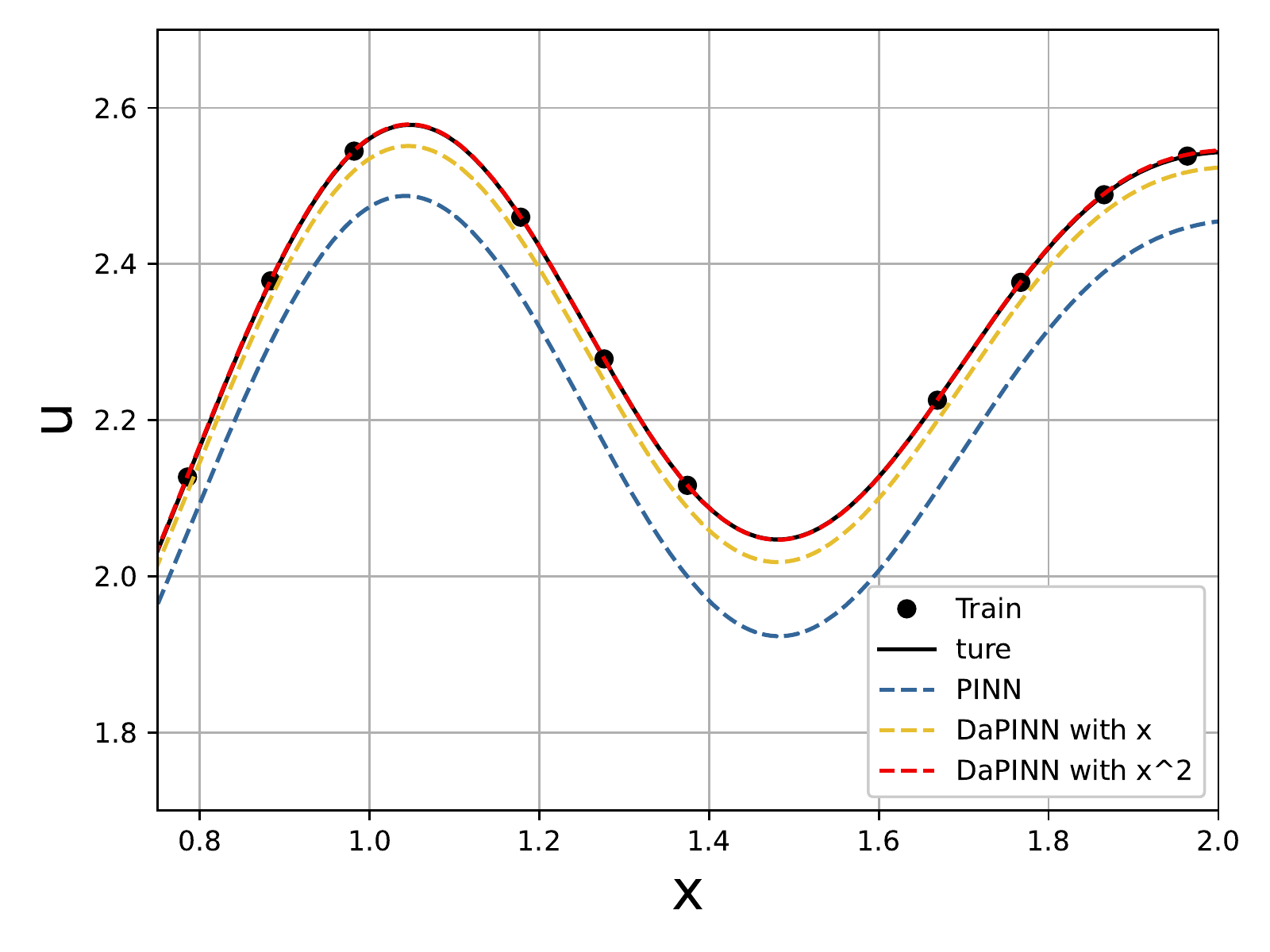}
\end{minipage}%
}%

  \centering
  \caption{Examples in Section 3.1.1: comparison of the PINN, DaPINN with x and DaPINN with $x^2$ models. A network size of $\left [ I, 20, 20, 20, 20, 1 \right ]$ and 10000 epochs with a learning rate of 0.001 are used, where $I=1$ for the PINN and $I=2$ for the DaPINN models.(a) L2 relative error of u versus the number of samples. (b) L2 relative error of u versus the number of training epochs. (c) PINN, DaPINN with x, and DaPINN with $x^2$ prediction function when the sample points are 26.}
\end{figure}
\subsubsection{2D Poisson equation}
Next, we discuss a 2D Poisson equation problem:
\begin{equation}
-\nabla u=f(x,y) ,\space\left(x,y\right)\epsilon\left[0,1\right]^2
\end{equation}
with Dirichlet boundary conditions of $u\left(x,y=0\ or\ x,y=1\right)=0$
In addition, f can be described as
\begin{multline}
f\left(x,y\right)=\frac{{16}^{a}a\left(a\left(1-2x\right)^2-2x^2+2x-1\right)\left(\left(x-1\right)x\left(y-1\right)y\right)^{a}}{\left(x-1\right)^2x^2}
\\
+\frac{{16}^{a}a\left(a\left(1-2y\right)^2-2y^2+2y-1\right)\left(\left(x-1\right)x\left(y-1\right)y\right)^{a}}{\left(y-1\right)^2y^2}
\end{multline}
The analytical solution of the PDE is $u\left(x,y\right)=2^{4a}x^{a}\left(1-x\right)^{a}y^{a}\left(1-y\right)^{a} ,a=10$

In this two-dimensional Poisson equation problem, we construct the neural network N(x) and apply replica augmentation and power series augmentation approach to expand the input dimension.

For replica augmentation, where $\tau_1=x, \tau_2=x, \tau_3=y, \tau_4=y$, the loss function is
\begin{equation}
\mathcal{L}=\omega_{f}\mathcal{L}_{f}+\omega_{b}\mathcal{L}_{b}
\end{equation}
in which $\mathcal{L}_{f}$ and $\mathcal{L}_{b}$ are the PDE residual term and the boundary condition residual term, respectively:

\begin{multline}
\mathcal{L}_f=\frac{1}{\mathcal{T}_{f}}\sum_{\mathbf{x}\in \mathcal{T}_{f}}|(\frac{\partial^2\mathcal{N}\left(\tau_1,\tau_2,\tau_3,\tau_4\right)}{\partial{\tau_1}^2}+2\frac{\partial^2\mathcal{N}\left(\tau_1,\tau_2,\tau_3,\tau_4\right)}{\partial \tau_1\partial \tau_2}+\frac{\partial^2\mathcal{N}\left(\tau_1,\tau_2,\tau_3,\tau_4\right)}{{\partial \tau_2}^2}\\
+\frac{\partial^2\mathcal{N}\left(\tau_1,\tau_2,\tau_3,\tau_4\right)}{\partial{\tau_3}^2}+2\frac{\partial^2\mathcal{N}\left(\tau_1,\tau_2,\tau_3,\tau_4\right)}{\partial \tau_3 \partial \tau_4}
+\frac{\partial^2\mathcal{N}\left(\tau_1,\tau_2,\tau_3,\tau_4\right)}{{\partial \tau_4}^2}+f\left(\tau_1,\tau_3\right))|^2
\end{multline}

\begin{equation}
\mathcal{L}_b=\frac{1}{\mathcal{T}_b}\sum_{\mathbf{x}\in \mathcal{T}_b}\left|\mathcal{N}\left(\tau_1,\tau_2,\tau_3,\tau_4\right)\right|^2
\end{equation}
For the second-order power series augmentation, where $\tau_1=x, \tau_2=x^2, \tau_3=y, \tau_4=y^2$, the loss function is
\begin{equation}
\mathcal{L}=\omega_{f}\mathcal{L}_{f}+\omega_{b}\mathcal{L}_{b}
\end{equation}
in which $\mathcal{L}_{f}$ and $\mathcal{L}_{b}$ are the PDE residual term and boundary condition residual term, respectively:

\begin{multline}
\mathcal{L}_{f}=\frac{1}{\mathcal{T}_{f}}\sum_{\mathbf{x}\in \mathcal{T}_{f}}|(\frac{\partial^2\mathcal{N}\left(\tau_1,\tau_2,\tau_3,\tau_4\right)}{\partial \tau_1^2}+4\tau_1\frac{\partial^2\mathcal{N}\left(\tau_1,\tau_2,\tau_3,\tau_4\right)}{\partial \tau_1\partial \tau_2}\\
+4\tau_2\frac{\partial^2\mathcal{N}\left(\tau_1,\tau_2,\tau_3,\tau_4\right)}{{\partial \tau_2}^2}+2\frac{\partial\mathcal{N}\left(\tau_1,\tau_2,\tau_3,\tau_4\right)}{\partial \tau_2}+\frac{\partial^2\mathcal{N}\left(\tau_1,\tau_2,\tau_3,\tau_4\right)}{\partial \tau_3^2}+4\tau_3\frac{\partial^2\mathcal{N}\left(\tau_1,\tau_2,\tau_3,\tau_4\right)}{\partial \tau_3\partial \tau_4}\\
+4\tau_4\frac{\partial^2\mathcal{N}\left(\tau_1,\tau_2,\tau_3,\tau_4\right)}{{\partial \tau_4}^2}+2\frac{\partial\mathcal{N}\left(\tau_1,\tau_2,\tau_3,\tau_4\right)}{\partial \tau_4}+f\left(\tau_1,\tau_3\right))|^2
\end{multline}

\begin{equation}
\mathcal{L}_{b}=\frac{1}{\mathcal{T}_{b}}\sum_{\mathbf{x}\in \mathcal{T}_{b}}\left|\mathcal{N}\left(\tau_1,\tau_2,\tau_3,\tau_4\right)\right|^2
\end{equation}
Here, we set $\omega_{f}=\omega_{b}=1$
As shown in Fig. 3, DaPINN demonstrates excellent performance. The L2 error of the DaPINN with x decreases to 0.8\% at 450 training points, while the DaPINN with $x^2$ and PINN require 550 and 575 training points, respectively, to achieve the same accuracy. As shown in Fig. 3(b), the DaPINN with $x$ outperforms the DaPINN with $x^2$ at 500 sample points, while the DaPINN with $x^2$ outperforms the PINN. However, when the number of sample points reaches 1000, the DaPINN with $x^2$ outperforms the replica-augmented DaPINN.
\begin{figure}[htbp]
  \centering
  \subfigure[]{
  \begin{minipage}[t]{0.5\linewidth}
  \centering
  \includegraphics[width=60mm]{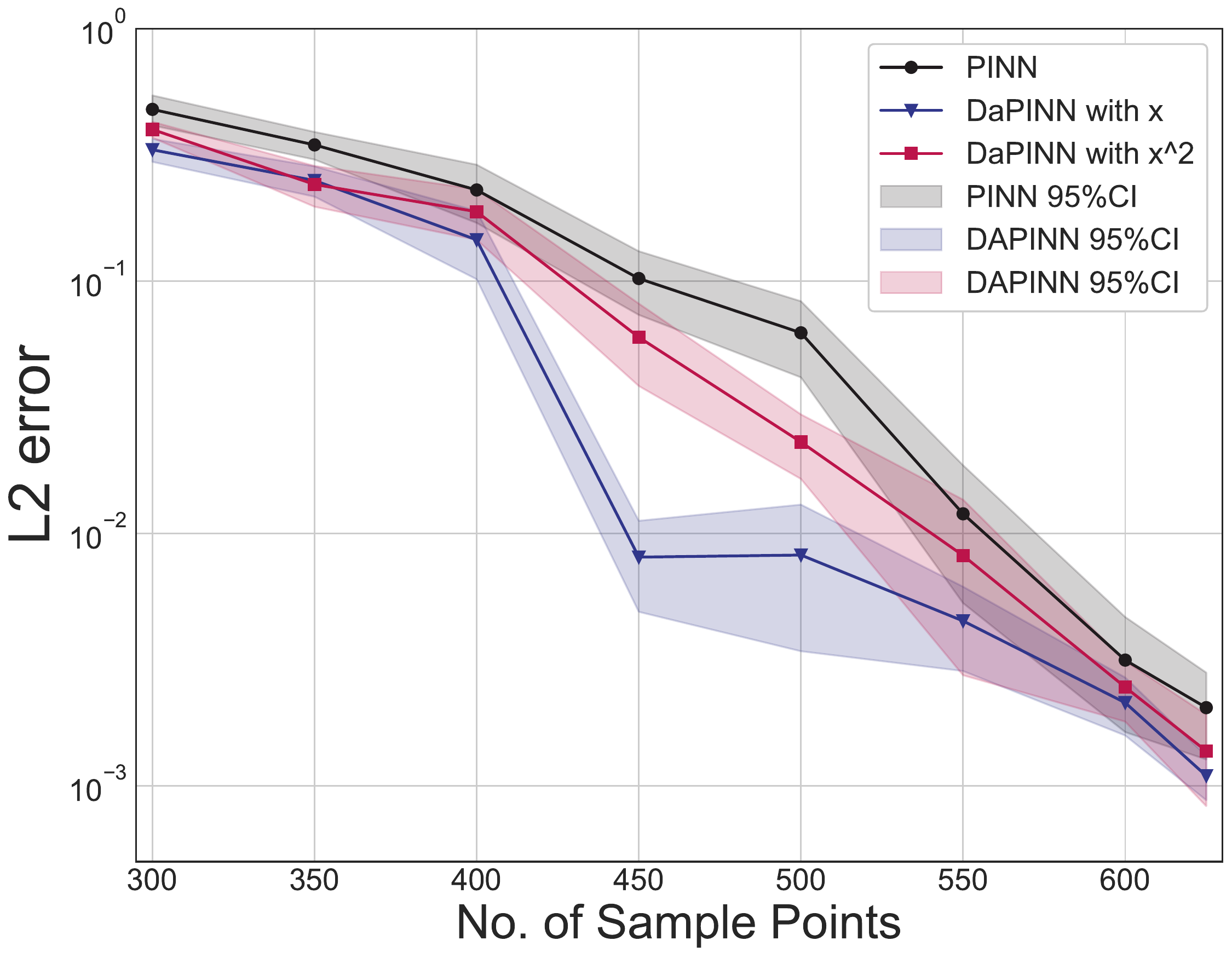}
  \end{minipage}%
  }%
  \subfigure[]{
  \begin{minipage}[t]{0.5\linewidth}
  \centering
  \includegraphics[width=60mm]{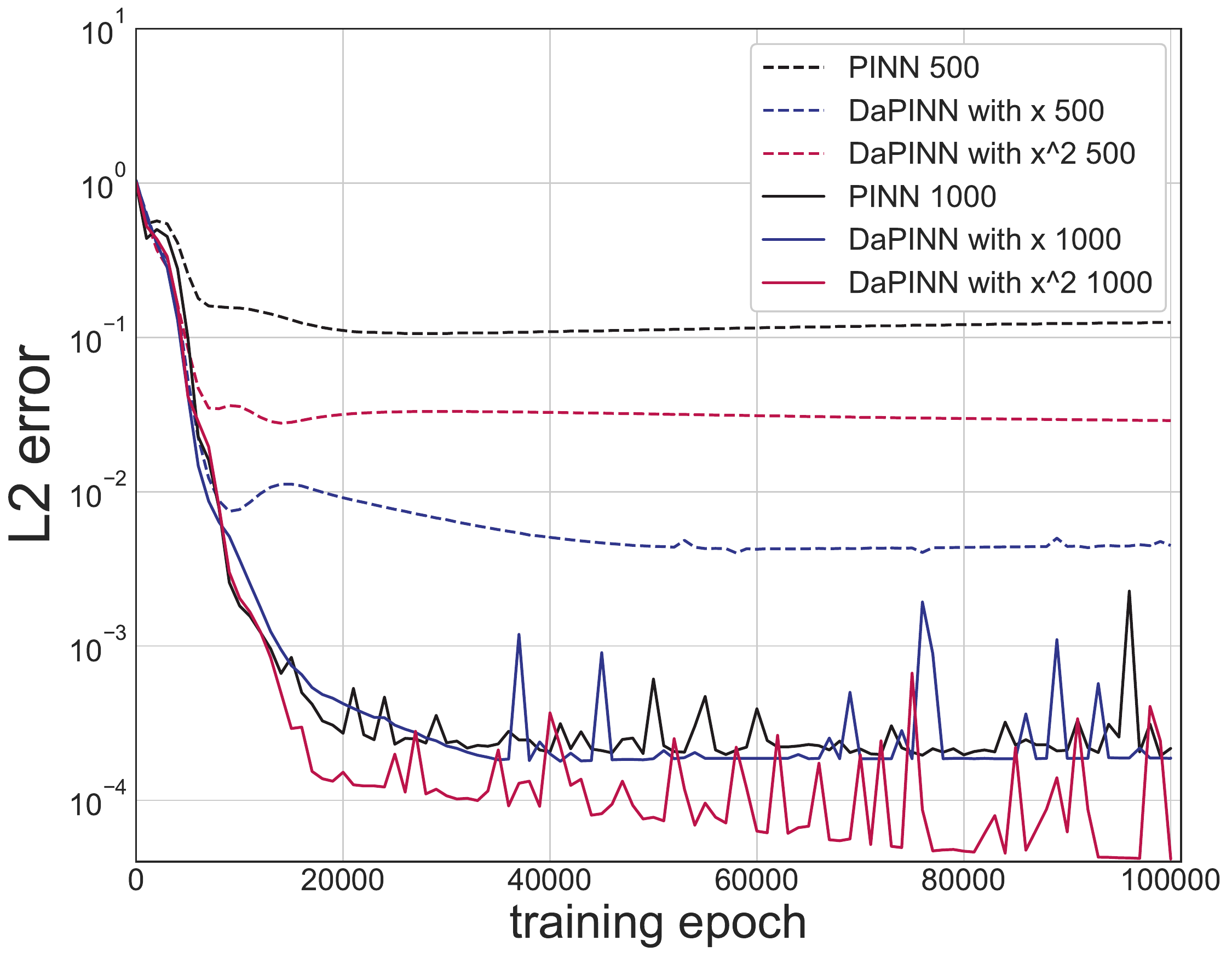}
  \end{minipage}%
  }%
  \centering
  \caption{Example in Section 3.1.2: comparison of the PINN, DaPINN with x and DaPINN with $x^2$ models. A network size of $\left [ I, 50, 50, 1 \right ]$ and 10000 epochs with a learning rate of 0.001 are used, where $I=2$ for the PINN and $I=4$ for the DaPINN. (a) Variation in the L2 relative error of u with the number of samples. (b) Variation in L2 relative error of u with the number of training epochs at 500 and 1000 sample points.}
\end{figure}

Here, we find replica augmentation performs better than the second-order power series augmentation approach when less than 575 sample points are used. With fewer sample points, the network is prone to overfitting, and replica aug increases the training difficulty by introducing exactly mirrored inputs so that the results can be obtained from two identical inputs, while the randomness of the weight parameters ensures the two inputs do not contribute equally to the results. Thus, the model cannot overfit all samples, improving the generalizability. The power series augmentation approach has a weaker generalizability than the replica augmentation method due to the lack of symmetry in the inputs. However, when a sufficient number of sample points is considered, the power series augmentation approach shows better performance in the nonlinear region fitting.

On the one hand, the accuracy of the fully computed DaPINN is significantly better than that of the PINN; on the other hand, the DaPINN requires fewer training epochs than the PINN to achieve the same accuracy. This result demonstrates the generalizability of the DaPINN with replica augmentation approach as well as the good fitting performance of the DaPINN with power series augmentation approach, demonstrating that these DaPINNs achieved higher accuracy with reduced computational costs.
\subsection{Input dimension expansion in inverse problems}
In the previous experiment, we verified the effectiveness of the dimensional expansion method in solving partial differential equation problems, demonstrating that the proposed approaches show significantly better performance than the PINN.

In this subsection, we compare the performance of the DaPINN and PINN models on inverse problems with the 1-dimensional Poisson equation and the diffusion reaction equation to show that DaPINN has higher accuracy and lower computational cost than PINN.
\subsubsection{Poisson (inverse)}
Returning to the 1D Poisson equation in Section 3.1.1, in the following inverse problem, the source function 
\emph{f} 
is no longer given to the neural network. Instead, we compute 
\emph{f} 
by solving the function for a series of points.
\begin{equation}
-\Delta u=f(x), x \in(0,\pi)
\end{equation}
Consider a system with the boundary conditions $u(0) = 0, u(\pi) = \pi$ where the source 
\emph{f} 
has the form
\begin{equation}
f\left(x\right)=\sum_{i=1}^{4}{i\sin{(ix)}}
\end{equation}
The analytic solution of u is
\begin{equation}
u\left(x\right)=\ x+\sum_{i=1}^{4}{\frac{1}{i}\sin{(ix)}}
\end{equation}
After selecting a uniform number of measurement points in $(0,\pi)$, we use two parallel neural networks to approximate 
\emph{f} 
and 
\emph{u} 
and compare the training results of the PINN and DaPINN methods.

Increasing the number of sampling points from 10 to 80 reduces the average error of 
\emph{f} 
by only approximately 30\% (Fig. 4(c) and (d)), while the result still has intolerable computational errors (Fig. 4(a) and (b)). Analogously, increasing the complexity of the network often means increasing the number of neurons, which make it more difficult to train the network so that the result may get worse in many cases (Fig. 4(e)). In fact, the PINN has poor performance when solving this kind of nonlinear inverse problem and its accurancy is hard to be improved by using conventional methods described above.

After applying the DaPINN with $x^2$, the training time reduces as the complexity of the network increases. Not only does the DaPINN have faster convergence rate, but also the DaPINN accuracy exceeds the PINN accuracy (Fig. 4(e)). Furthermore, the error in the DaPINN model is satisfactorily small when only 20 sample points are selected, with a large error only in the steep region near $x=0$ (Fig. 4(a) and (b)). The calculated L2 error of the parameter function 
\emph{f} 
is approximately 1/3 of that in the PINN model. And for the objective function 
\emph{u} 
, the error obtained by the DaPINN is an order of magnitude smaller than that obtained by the PINN (Fig. 4(c) and (d)).

\begin{figure}[htbp]
\centering
\subfigure[]{
\begin{minipage}[t]{0.5\linewidth}
\centering
\includegraphics[width=60mm]{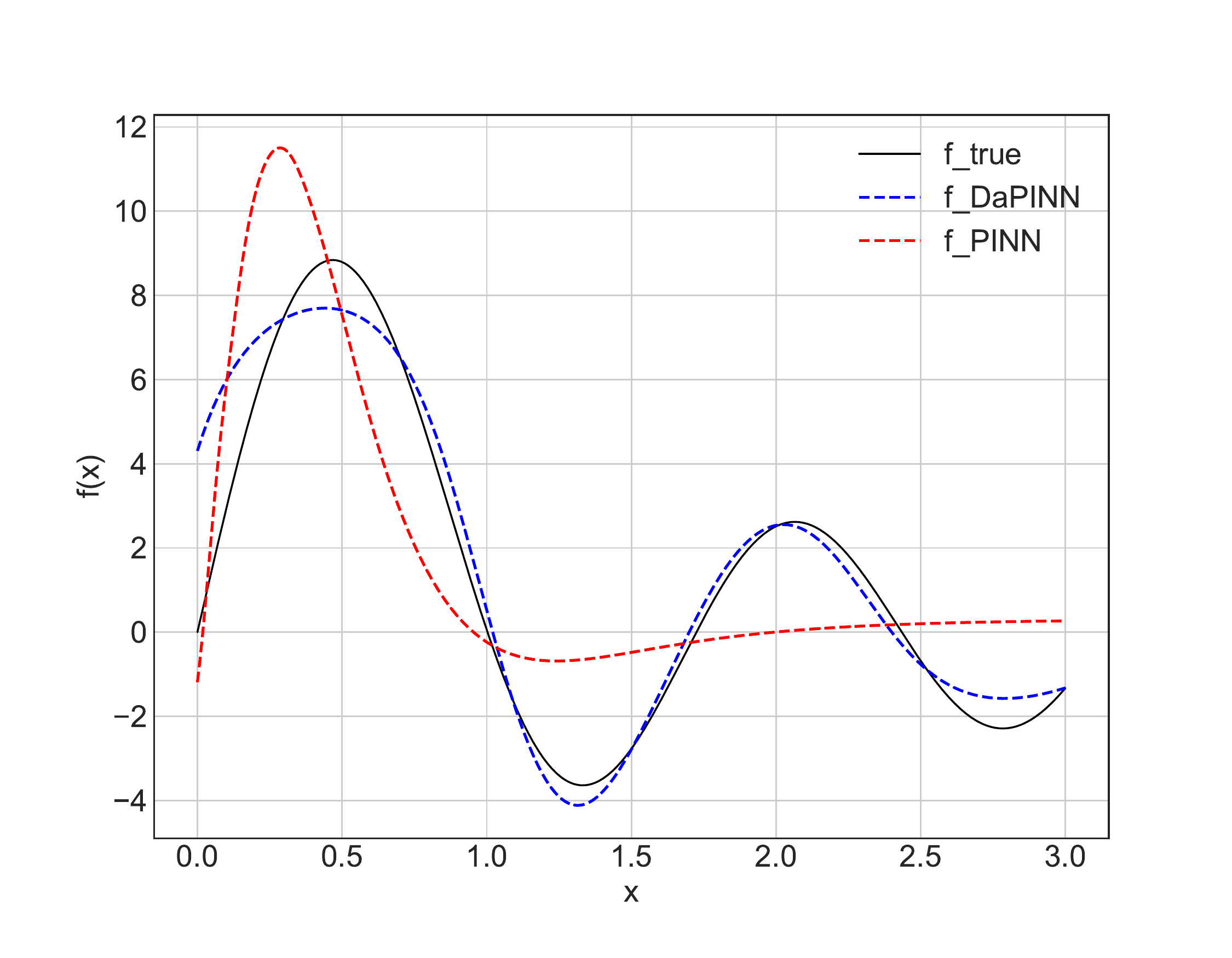}
\end{minipage}%
}%
\subfigure[]{
\begin{minipage}[t]{0.5\linewidth}
\centering
\includegraphics[width=60mm]{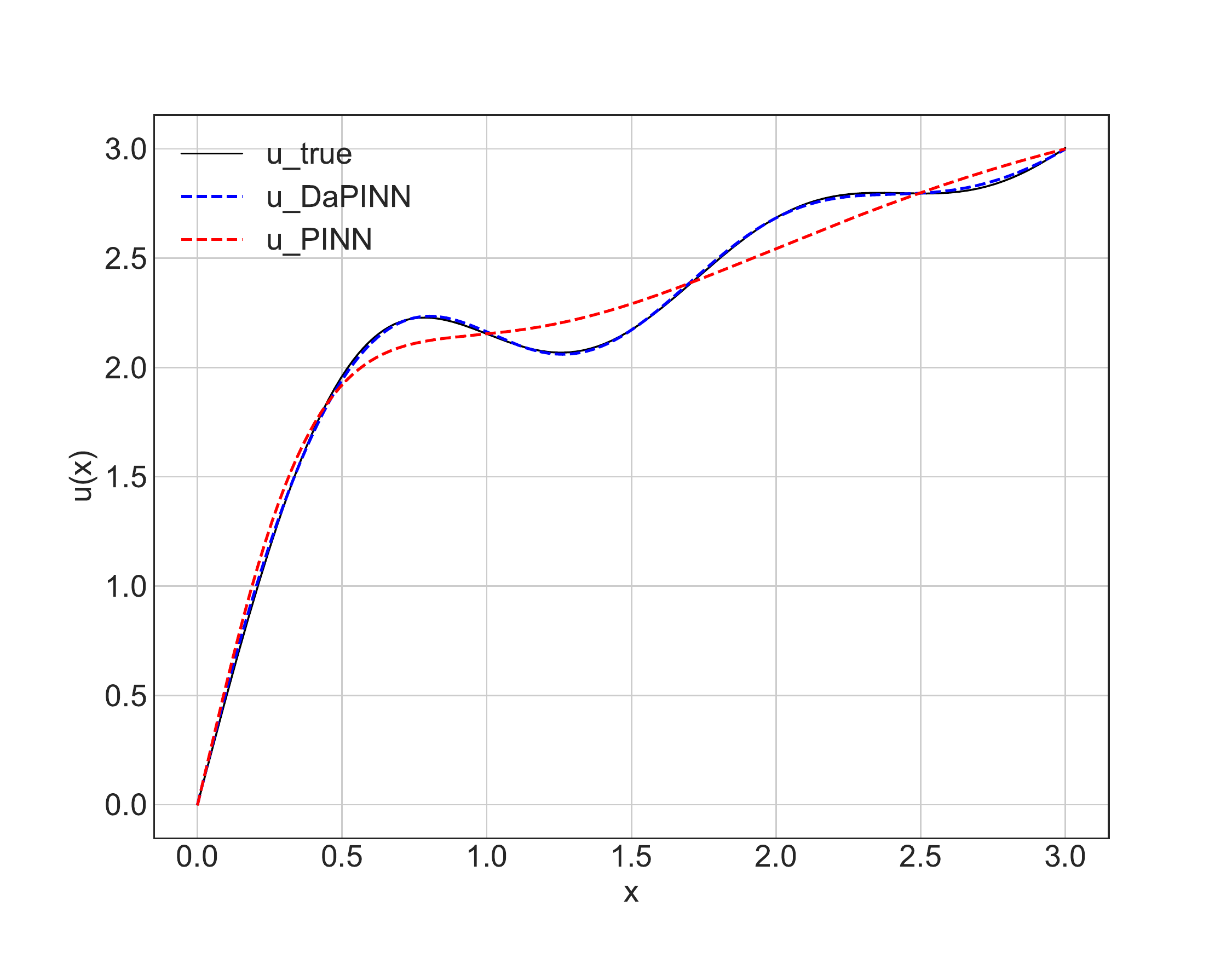}
\end{minipage}%
}%

\subfigure[]{
\begin{minipage}[t]{0.33\linewidth}
\centering
\includegraphics[width=50mm]{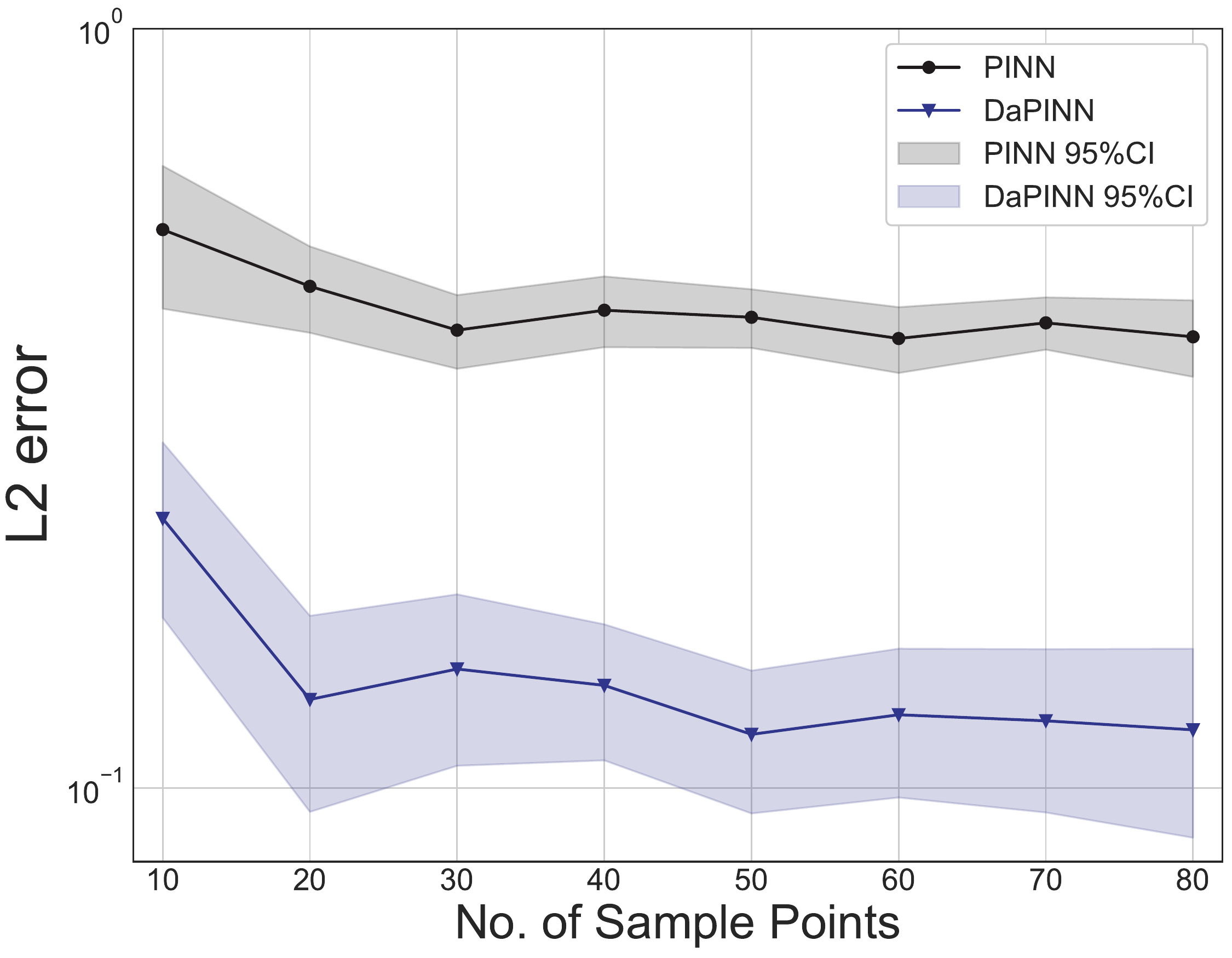}
\end{minipage}%
}%
\subfigure[]{
\begin{minipage}[t]{0.33\linewidth}
\centering
\includegraphics[width=50mm]{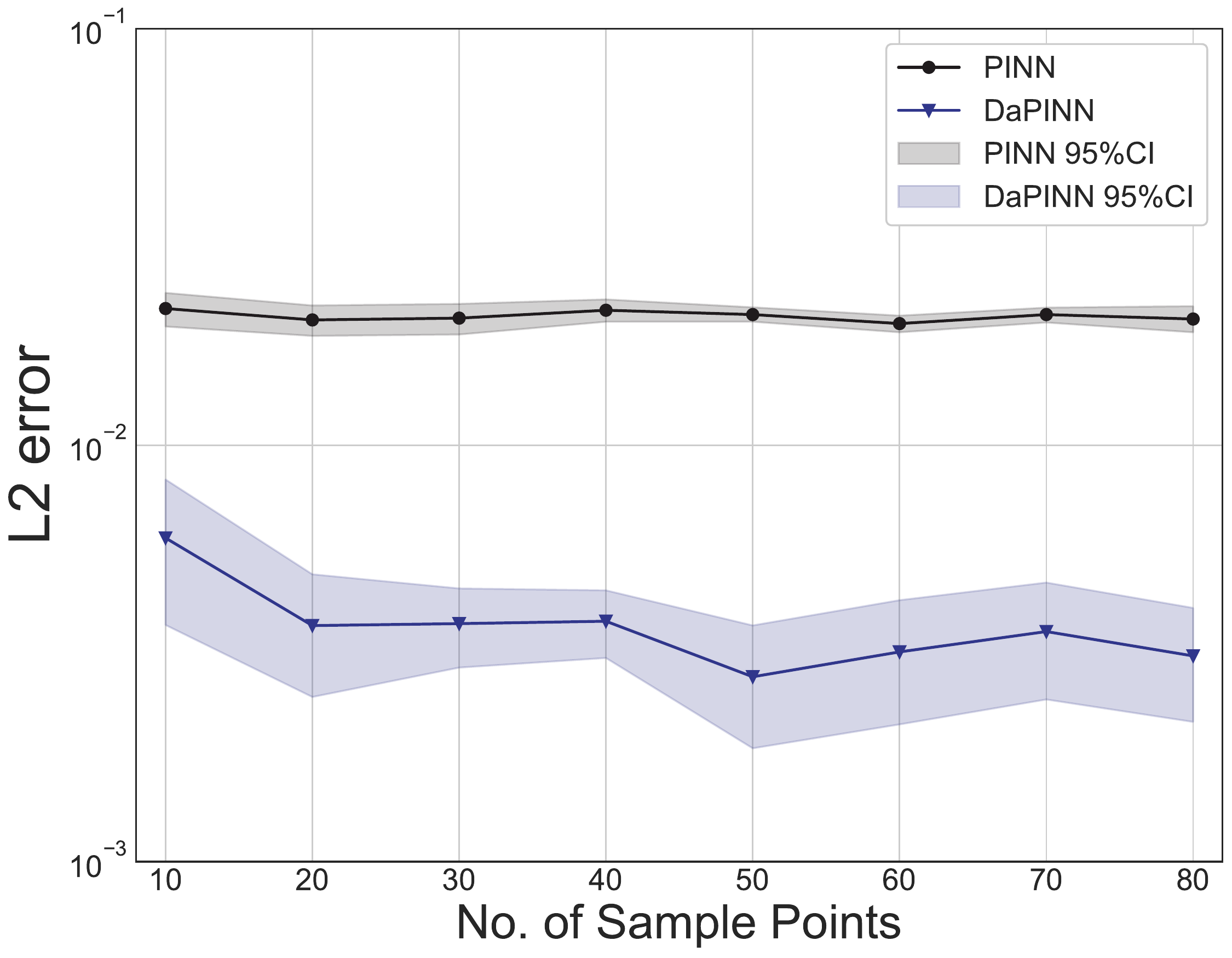}
\end{minipage}%
}%
\subfigure[]{
\begin{minipage}[t]{0.33\linewidth}
\centering
\includegraphics[width=50mm]{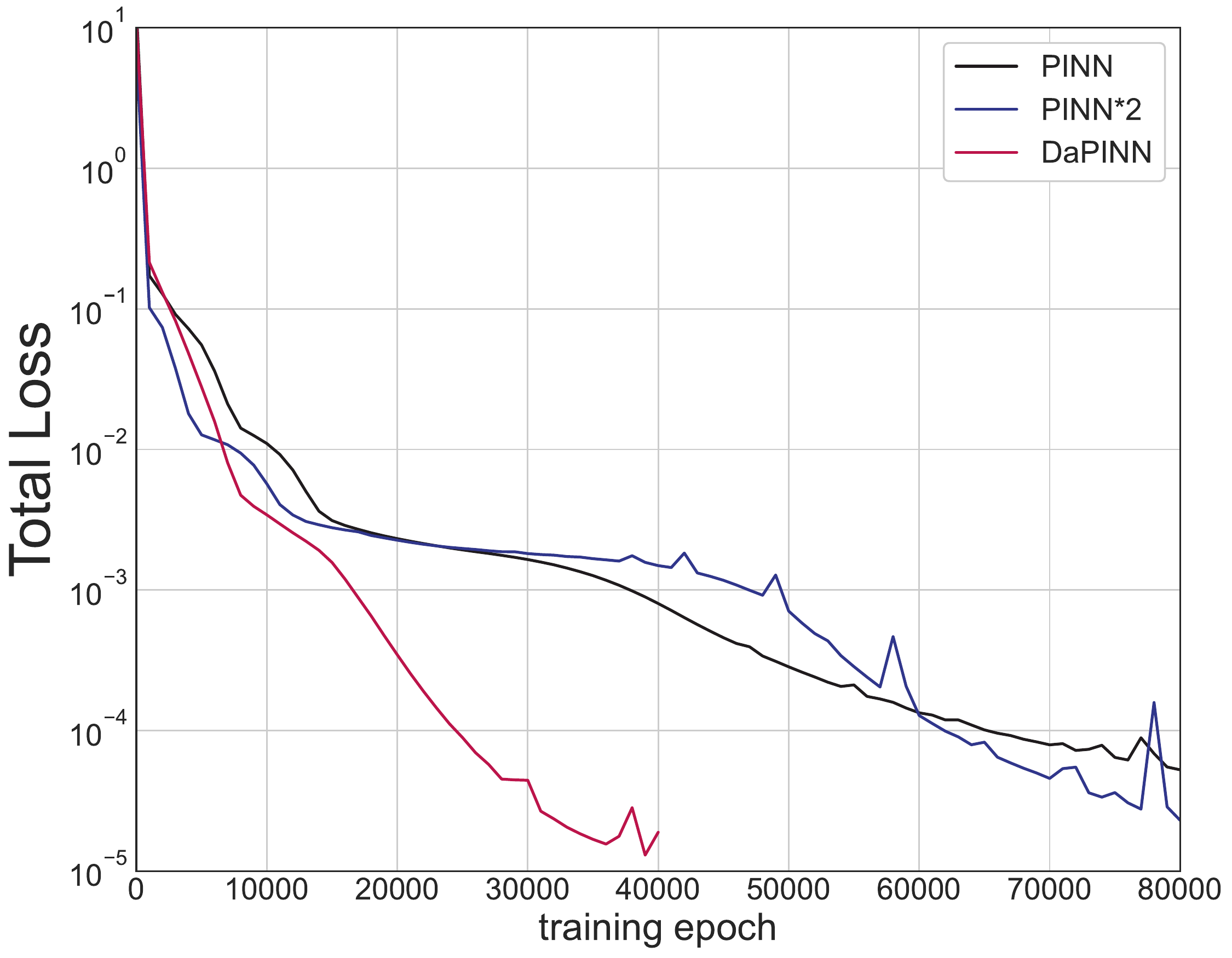}
\end{minipage}%
}%
\centering
\caption{Example in Section 3.2.1: comparison of the PINN and DaPINN with $x^2$ methods. A network size of $\left [ I, 30, 30, 30, 30, 1 \right ]$ and 20000 epochs with a learning rate of 0.001 are used, with $I=1$ for PINN and $I=2$ for DaPINN. (a) Analytic and predictive solutions of the PDE with 80 sample and training points. (b) Analytic and predictive solutions of the parametric function f(x). (c, d) Prediction source $\hat{f}$ versus the predicted solution $\mathcal{N}$ of the L2 error. (e) The number of training epochs for the loss function to reach 1E-5, where PINN*2 represents a PINN with twice the number of neurons.}
\end{figure}

\subsubsection{Thermal diffusion (inverse)}
We next consider another inverse problem in which the unknown parameter is a constant. In a one-dimensional heat conduction model $x\in\left(-1,1\right),\ t\in(0,1)$ the diffusion equation is known:
\begin{equation}
\frac{\partial u}{\partial t}=C\frac{\partial^2u}{{\partial x}^2}+k\left(x,t\right)       
\end{equation}
Here, the heat source $k(x,t)$ has the form:
\begin{equation}
k\left(x,t\right)=e^{-t}\left(\pi^2-1\right)sin{\left(\pi x\right)}.
\end{equation}
We choose thermal diffusivity $C=1$, the boundary condition $k(1,t)=k(-1,t)=0$, and the initial condition $k(x,0)=\sin(\pi x)$ . In addition, the analytical solution of the equation at this point is
\begin{equation}
u\left(x,t\right)=e^{-t}sin{\left(\pi x\right)}
\end{equation}
We next determine the thermal diffusivity $C$ of this system, which can be calculated using several measurement points on the boundary and in the initial (final) state (Fig. 5(a)).

When 80 training points were used for training, the error of parameter $C$ obtained by the DaPINN model quickly converged to within 0.1\%, while the error of the PINN approach was always greater than 1\%. Moreover, when no final state measurement points were used, the computational error of the DaPINN approach was less than the computational error of the PINN model. The PINN computational error was always greater than 10\%, and the results were basically unusable (Fig. 5(b)). When the number of training points was increased from 30 to 114, the PINN parameter error decreased from 26.3\% to 0.9\%, and the L2 error decreased from 24.7\% to 1.2\%. For the DaPINN model, the parameter error decreased from 0.27\% to 0.01\%, and the L2 error decreased from 1.6\% to 0.28\%. In this process, the gap between the two neural networks gradually decreases; however, until the error converges, the advantage of the DaPINN approach is apparent (Fig. 5(c) and (d)).

However, although the total error is reduced, the error of the DaPINN with $x^2$ model is concentrated in the steep region near $t = 0$ (Fig. 5(g) and (h)), which implies a flaw in augmenting the input features with $x^2$ in this problem. Some improvements are provided in Section 3.3.3.
\begin{figure}[htbp]
\centering
\subfigure[]{
\begin{minipage}[t]{0.5\linewidth}
\centering
\includegraphics[width=60mm]{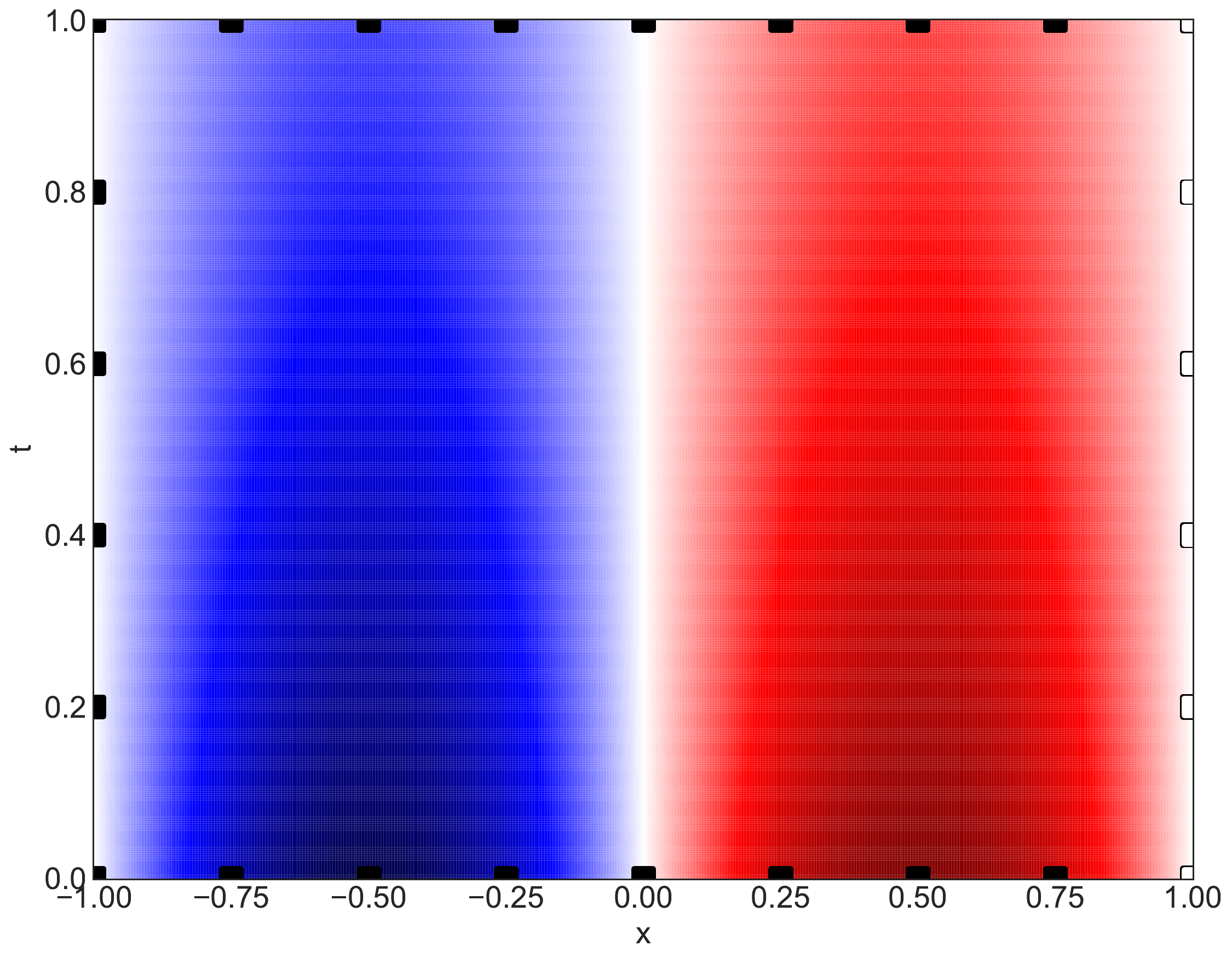}
\end{minipage}%
}%
\subfigure[]{
\begin{minipage}[t]{0.5\linewidth}
\centering
\includegraphics[width=60mm]{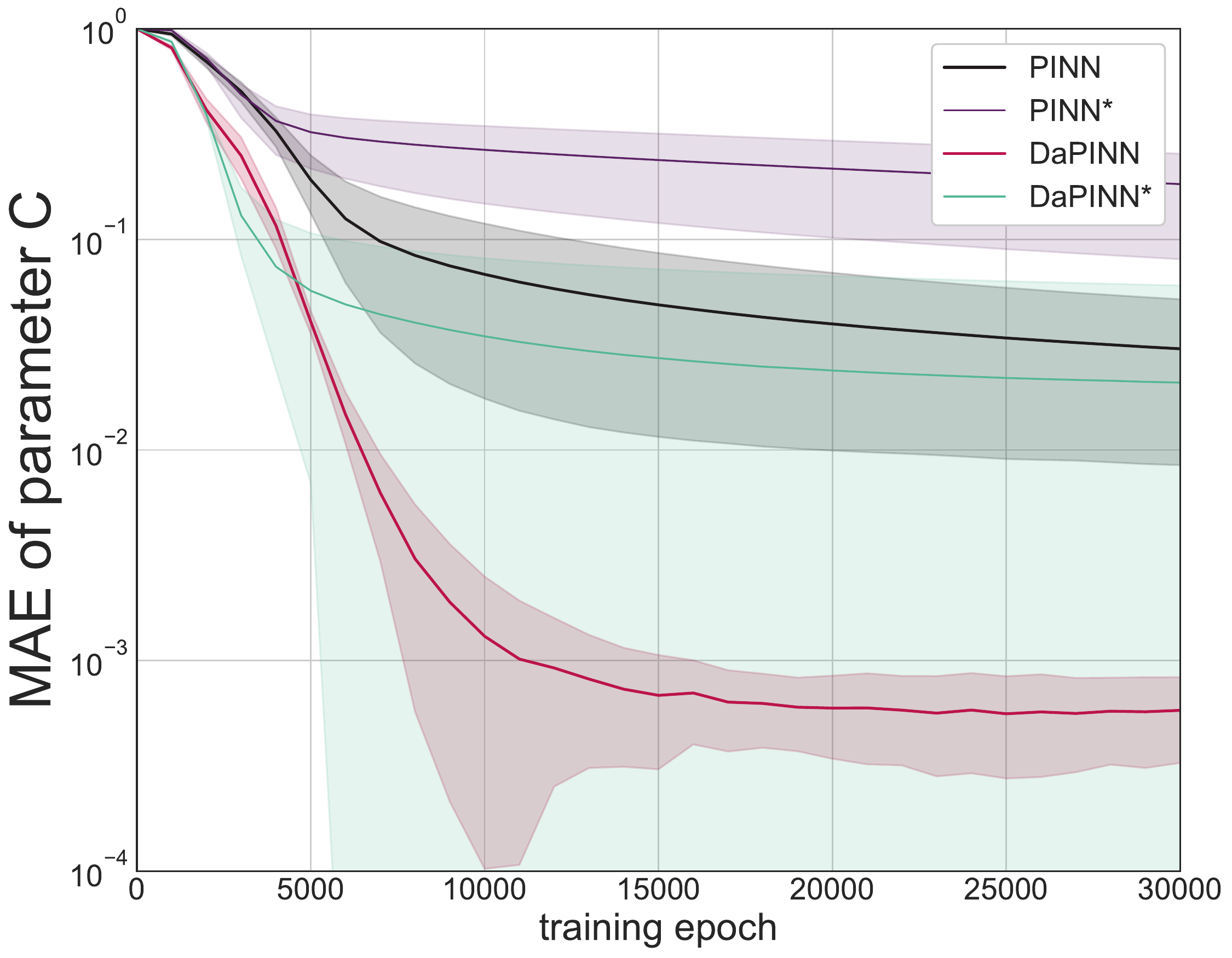}
\end{minipage}%
}%

\subfigure[]{
\begin{minipage}[t]{0.5\linewidth}
\centering
\includegraphics[width=60mm]{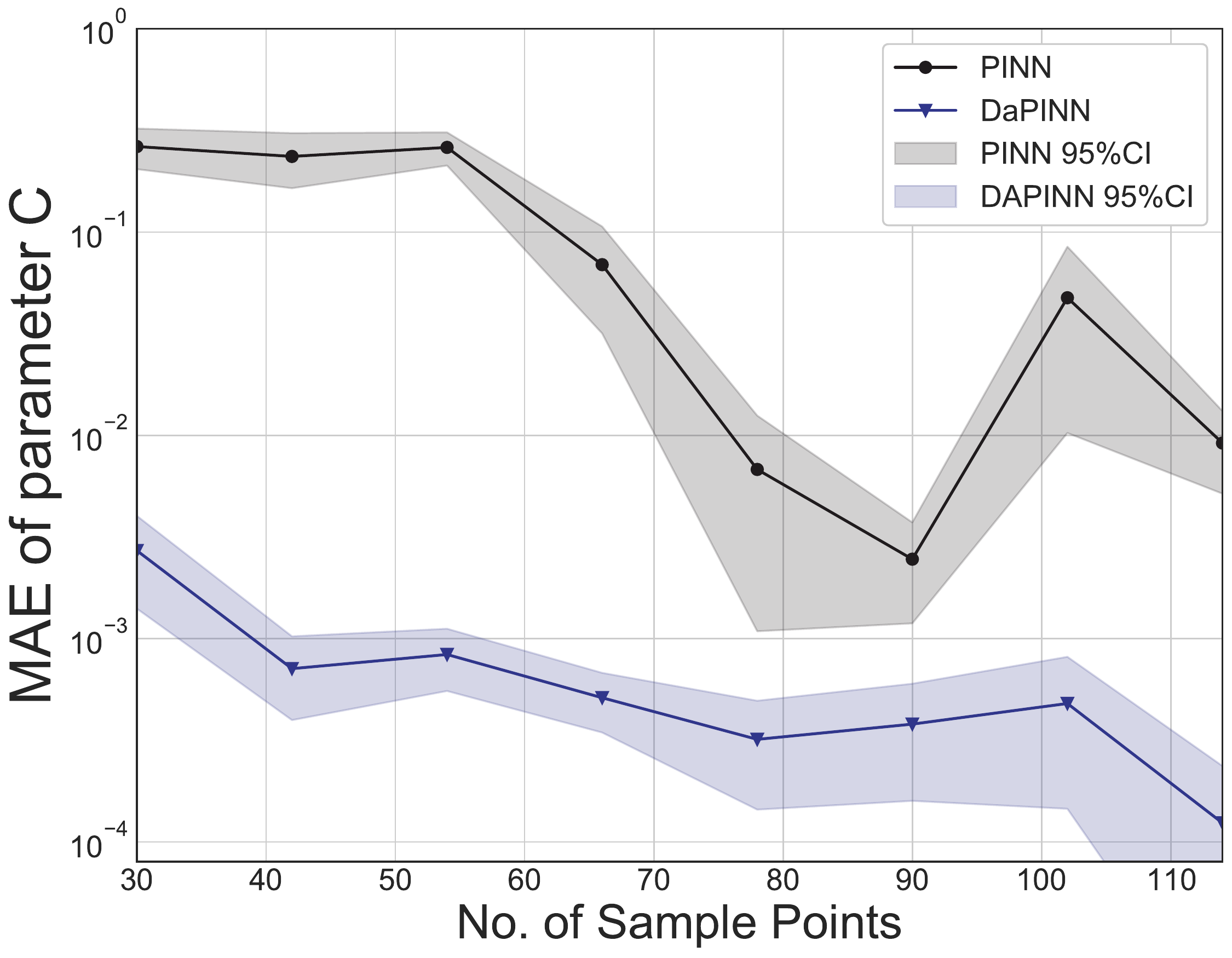}
\end{minipage}%
}%
\subfigure[]{
\begin{minipage}[t]{0.5\linewidth}
\centering
\includegraphics[width=60mm]{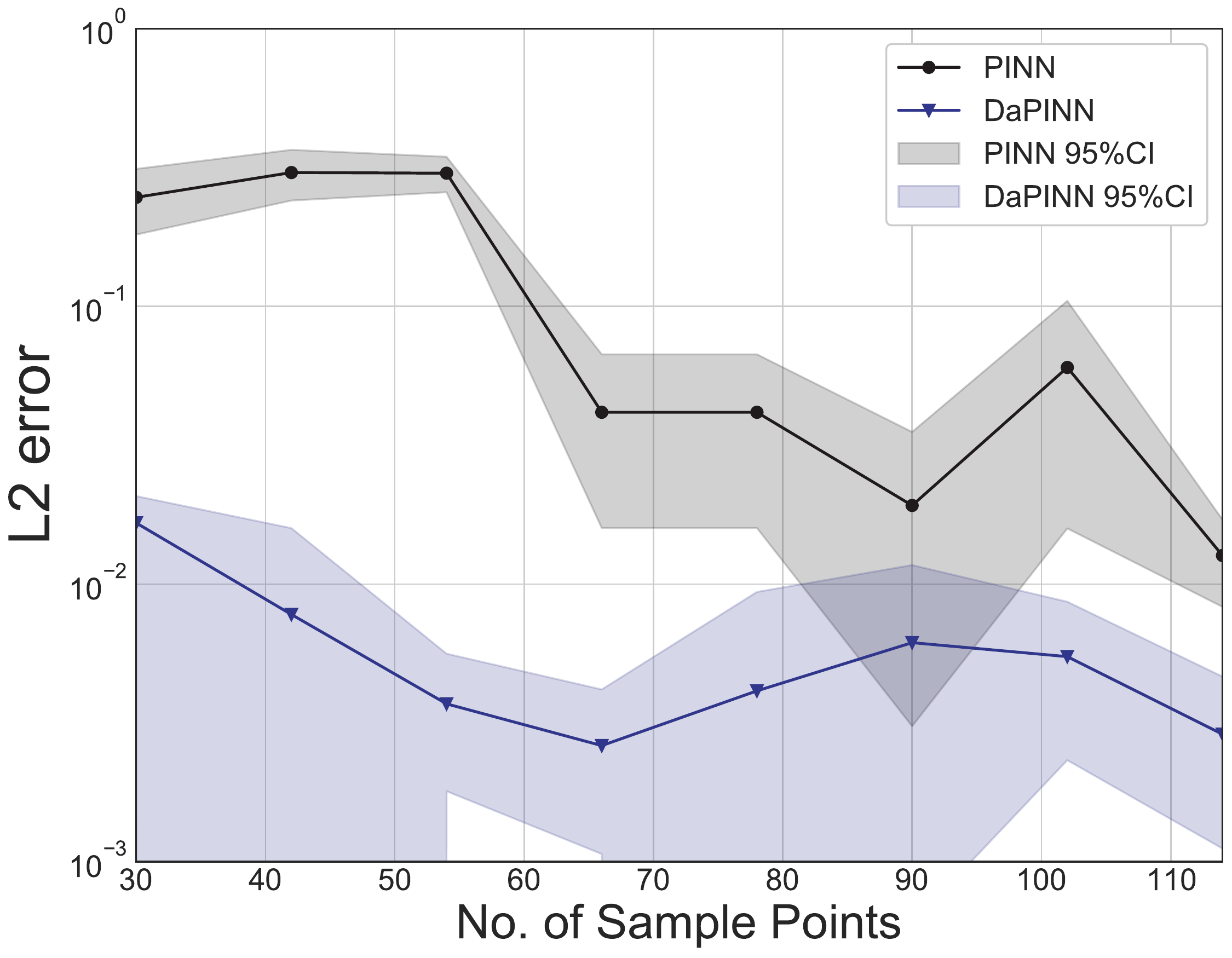}
\end{minipage}%
}%

\subfigure[]{
\begin{minipage}[t]{0.25\linewidth}
\centering
\includegraphics[width=40mm]{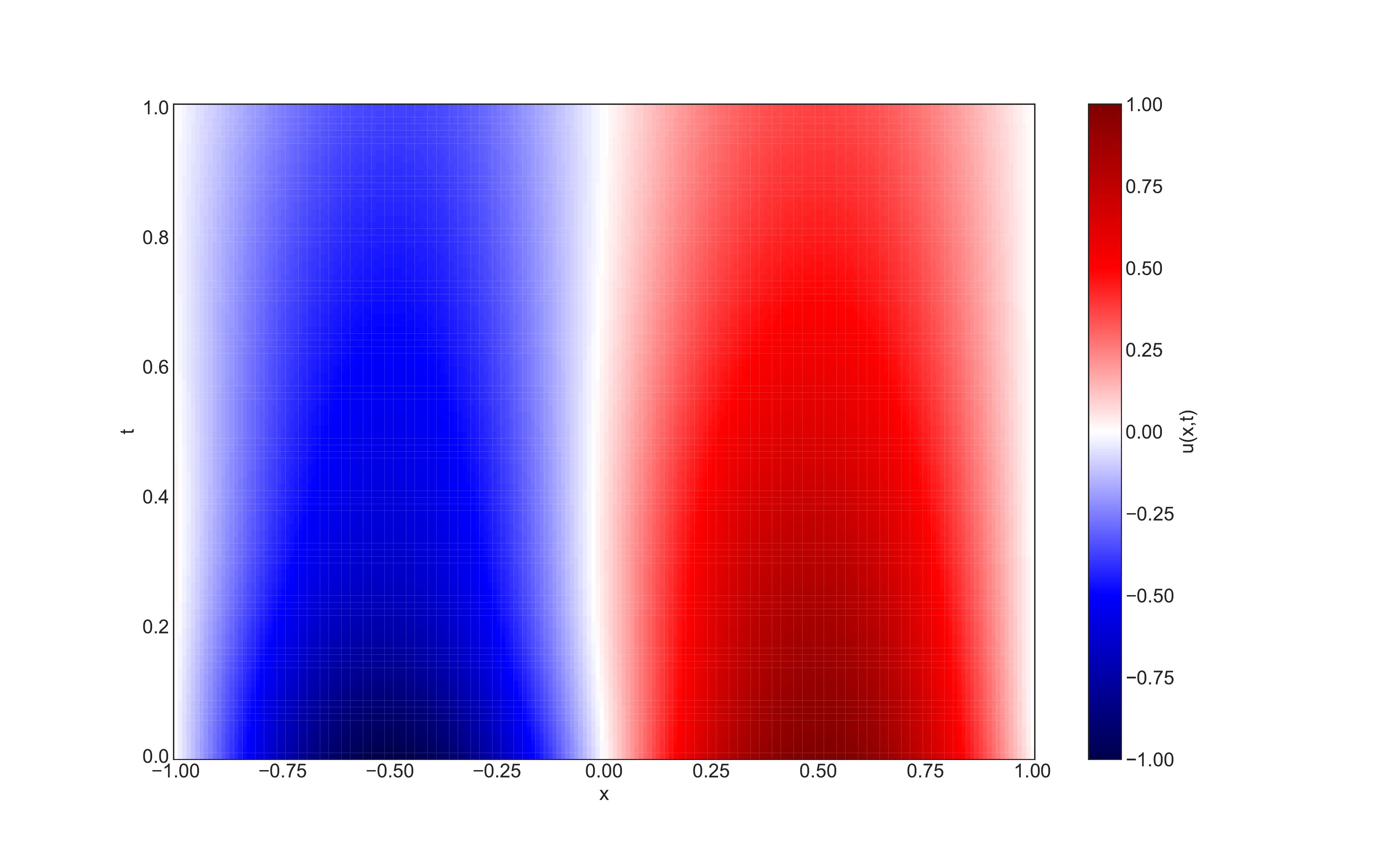}
\end{minipage}%
}%
\subfigure[]{
\begin{minipage}[t]{0.25\linewidth}
\centering
\includegraphics[width=40mm]{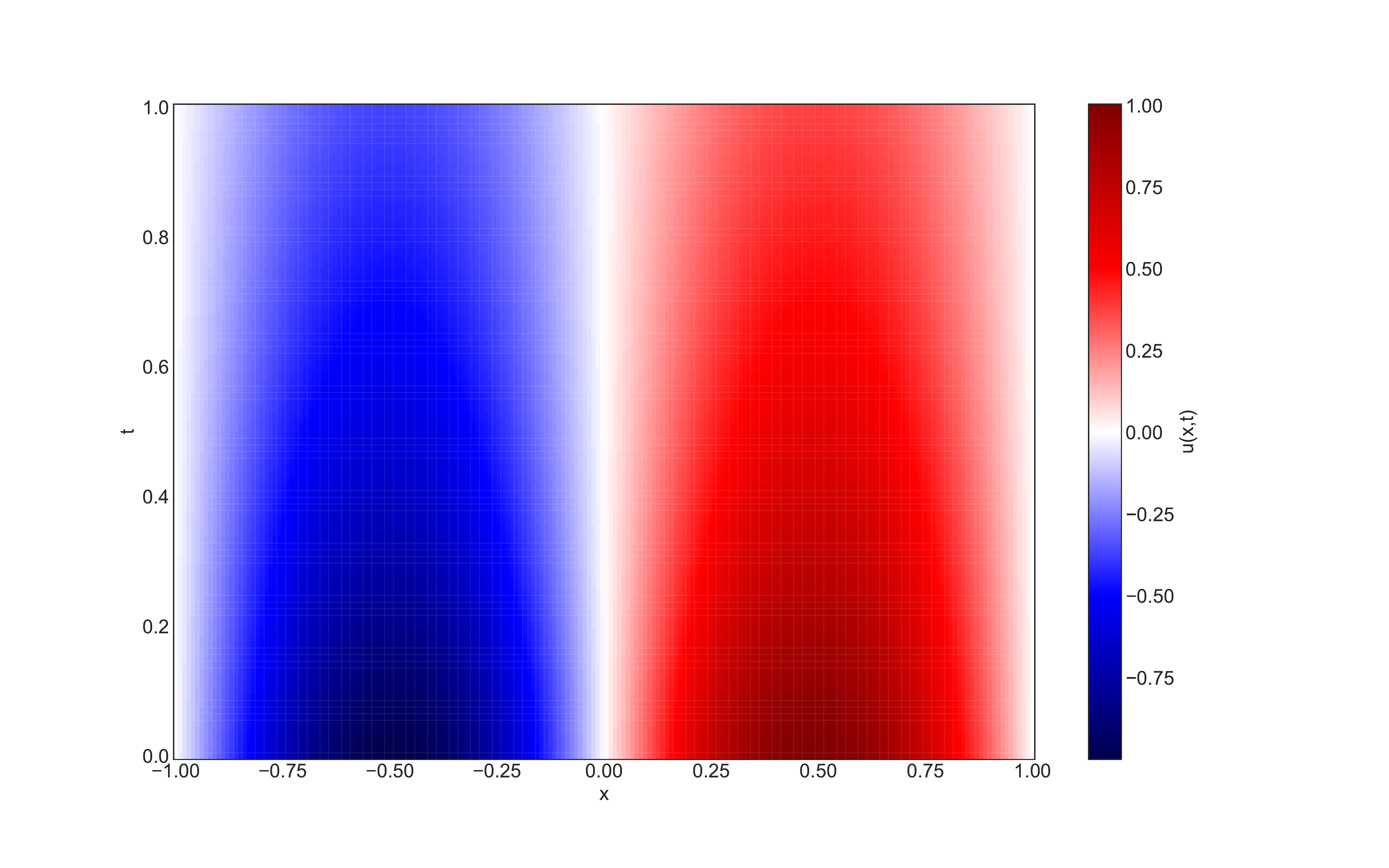}
\end{minipage}%
}%
\subfigure[]{
\begin{minipage}[t]{0.25\linewidth}
\centering
\includegraphics[width=40mm]{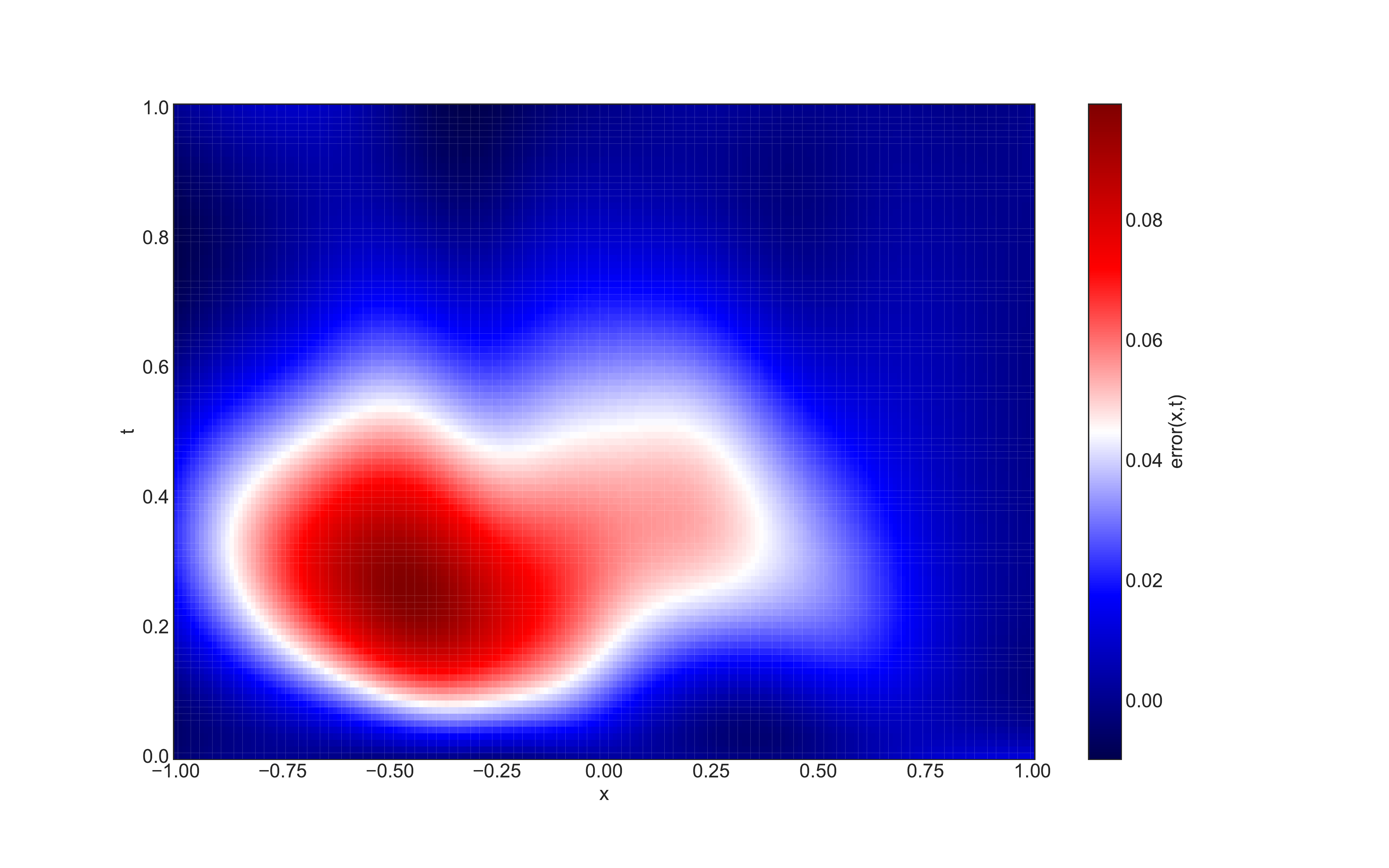}
\end{minipage}%
}%
\subfigure[]{
\begin{minipage}[t]{0.25\linewidth}
\centering
\includegraphics[width=40mm]{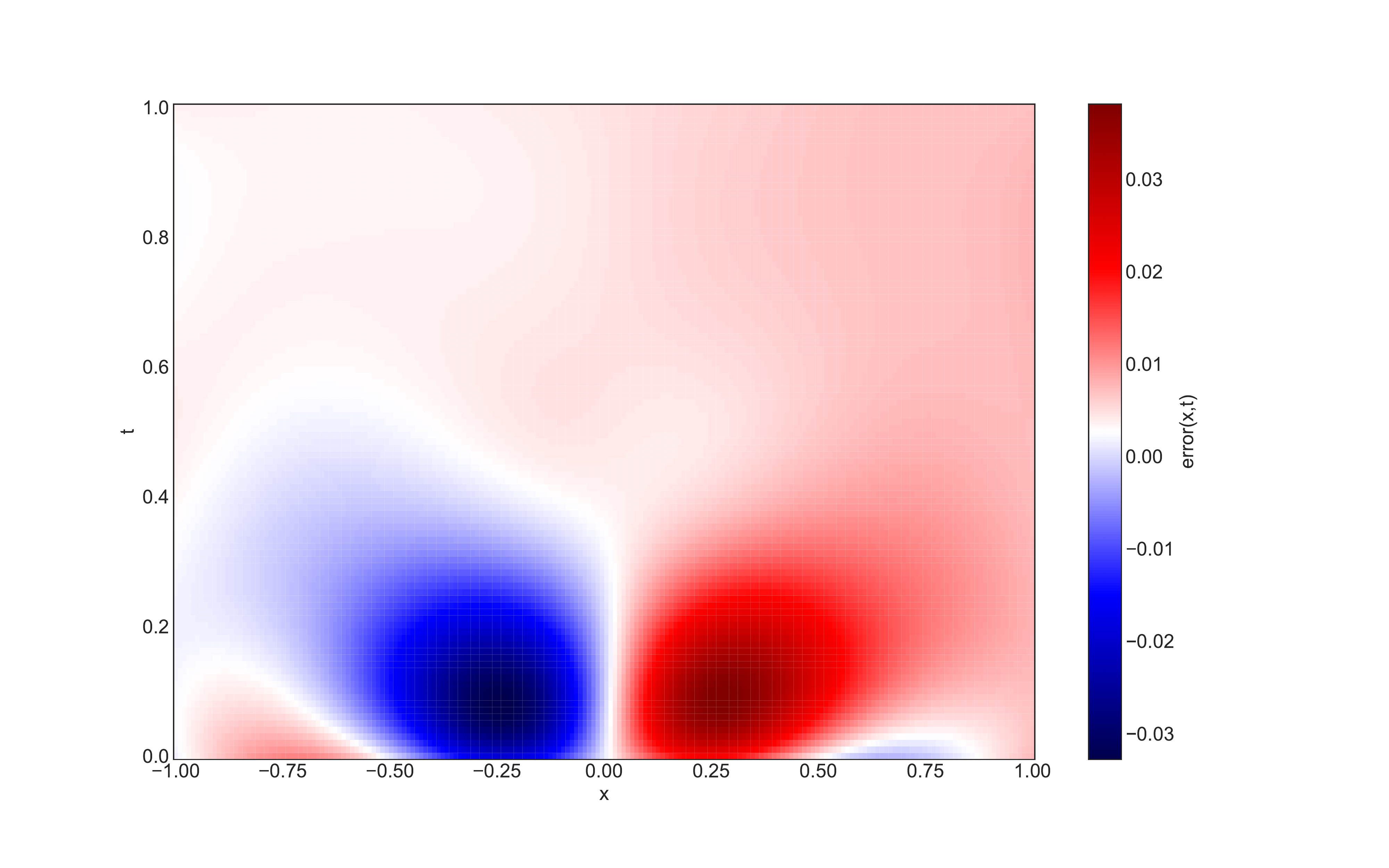}
\end{minipage}%
}%
\centering
\caption{Example in Section 3.2.2: comparison of the PINN and DaPINN with $x^2$ models. A network size of $\left [ I, 20, 20, 20, 1 \right ]$ and 10000 epochs with a learning rate of 0.001 are used, with $I=2$ for the PINN and $I=3$ for the DaPINN. (a) Analytic solution of the PDE; the black squares represent the initial measurement points at the boundary, and the white squares represent the measurement points in the final state. (b) Calculated relative error in the target parameter C. The dashed line '(Da)PINN*' is the case without using the end state measurement points, and 80 training points were used. (c) The relative error in the target parameter C. (d) The calculated $\mathcal{N}$ of the L2 error. (e, f) The predicted solutions $\mathcal{N}_{PINN}(x)$ and $\mathcal{N}_{DaPINN}(x,x^2)$ (g, h) The absolute error distributions of the predicted solutions. }
  \end{figure}

\subsection{Higher-order power series augmentation methods} 	
We demonstrated the validity of using dimension augmentation in the PINN model through several problems. It is worth noting that the DaPINN with second-order power series augmentation model achieves the best performance in all of the above examples, which demonstrates the generalizability of our method. However, whether better performance can be achieved using higher-order power series augmentation must be investigated. The following examples imply that third-order power series augmentation have better potential than second-order power series augmentation. The impact of introducing third-order power series augmentation is discussed in the following section.
\subsubsection{Heat}
Here, we discuss a 1D heat conduction problem:
\begin{equation}
\frac{\partial u}{\partial t}=a\frac{\partial^2u}{\partial x^2},\space x\epsilon\left[0,1\right],t\epsilon\left[0,1\right]
\end{equation}
The initial and boundary conditions are defined as follows:
\begin{equation}
u\left(0,t\right)=u\left(1,t\right)=0
\end{equation}
\begin{equation}
u\left(x,0\right)=sin{\pi x}
\end{equation}
Where, $a=0.4$ is the thermal diffusion coefficient.

The analytical solution is
\begin{equation}
u\left(x,y\right)=e^{-\left({a\pi}^2t\right)}sin\left(\pi x\right)
\end{equation}
In this experiment, we use higher-order power series augmentation in the DaPINN model and introduce $x^2$ and $x^3$ for input augmentation to discuss the effects of introducing more dimensions on the DaPINN model.

The second-order power series augmentation approach is similar to the method applied in Section 3.1, and the third-order power series augmentation is introduced below. And for this example, we have $\tau_1=x, \tau_2=x^2, \tau_3=x^3, \tau_4=t$.

We construct neural network with the loss function:
\begin{equation}
\mathcal{L}=\omega_{f}\mathcal{L}_{f}+\omega_{b}\mathcal{L}_{b}+\omega_{i}\mathcal{L}_{i}
\end{equation}

where $L_{f}$ and $L_{b}$ are the PDE residual term and boundary condition residual term, respectively:

\begin{multline}
\mathcal{L}_f=\frac{1}{\left|\mathcal{T}_f\right|}\sum_{\mathbf{x}\in \mathcal{T}_f}|(\frac{\partial\mathcal{N}\left(\tau_1,\tau_2,\tau_3,\tau_4\right)}{\partial \tau_4}\\
-a(\frac{\partial^2\mathcal{N}\left(\tau_1,\tau_2,\tau_3,\tau_4\right)}{{\partial \tau_1}^2}+4\tau_2\frac{\partial^2\mathcal{N}\left(\tau_1,\tau_2,\tau_3,\tau_4\right)}{{\partial \tau_2}^2}+9\tau_2^2\frac{\partial^2\mathcal{N}\left(\tau_1,\tau_2,\tau_3,\tau_4\right)}{{\partial \tau_3}^2}\\
+4\tau_1\frac{\partial^2\mathcal{N}\left(\tau_1,\tau_2,\tau_3,\tau_4\right)}{\partial \tau_1\partial \tau_2}+6\tau_2\frac{\partial^2\mathcal{N}\left(\tau_1,\tau_2,\tau_3,\tau_4\right)}{\partial \tau_1\partial \tau_3}+12\tau_3\frac{\partial^2\mathcal{N}\left(\tau_1,\tau_2,\tau_3,\tau_4\right)}{\partial \tau_2\partial \tau_3}\\
+2\frac{\partial\mathcal{N}\left(\tau_1,\tau_2,\tau_3,\tau_4\right)}{\partial \tau_2}+6x\frac{\partial\mathcal{N}\left(\tau_1,\tau_2,\tau_3,\tau_4\right)}{\partial \tau_3})|^2
\end{multline}

\begin{equation}
\mathcal{L}_b=\frac{1}{\left|\mathcal{T}_b\right|}\sum_{\mathbf{x}\in \mathcal{T}_b}\left|\mathcal{N}\left(\tau_1,\tau_2,\tau_3,\tau_4\right)\right|^2
\end{equation}
\begin{equation}
\mathcal{L}_i=\frac{1}{\left|\mathcal{T}_i\right|}\sum_{\mathbf{x}\in \mathcal{T}_i}\left|\mathcal{N}\left(\tau_1,\tau_2,\tau_3,\tau_4\right)-sin{\pi x}\right|^2
\end{equation}
Here, we set $\omega_{f}=\omega_{b}=\omega_{i}=1$

The results are shown in Fig. 6(a), and the errors of all three methods are greater than 100\% when 9 training points are used. When the number of training points is increased to 45, the error in the DaPINN with the third-order power series augmentation($x^3$) model decreases to 0.5\%, while the errors in the DaPINN with $x^2$ and PINN models are 0.9\% and 2.3\%, respectively. When the number of training points is increased to 117, the error in the DaPINN with $x^3$ model decreases to 0.04\%, while the errors in the DaPINN with $x^2$ and PINN models are 0.16\% and 0.32\%, respectively. As shown in Fig. 6(b), the error in the DaPINN with $x^3$ model decreases to 1.4\% when the number of training epochs reaches 4000, while the error in the DaPINN with $x^2$ and PINN models are 2.4\% and 4.3\%, respectively. When the number of training epochs is increased to 40,000, the error in the DaPINN with $x^3$ model decreases to 0.08\%, while the errors in the DaPINN with $x^2$ and PINN models are 0.12\% and 0.18\%.
We find that the DaPINN with $x^3$ model performs significantly better than the DaPINN with $x^2$ model. Taking into account the model conditions and the fact that the analytic solution contains a sine term, the power series augmentation of $\sin{x}$ is given by
\begin{equation}
sin\left(x\right)=\sum_{n\geq  0}{\frac{\left(-1\right)^{n}x^{2n+1}}{\left(2n+1\right)!}} 
\end{equation}
The solution to this problem is more sensitive to odd power terms of $x$; thus, the coefficient of $x^2$ is smaller, and the DaPINN model easily degrades to the PINN approach, while the DaPINN with higher-order power series terms has better generalizability and higher accuracy. Therefore, we infer that introducing higher-order power series terms can improve the DaPINN performance.
\begin{figure}[htbp]
\centering
\subfigure[]{
\begin{minipage}[t]{0.5\linewidth}
\centering
\includegraphics[width=60mm]{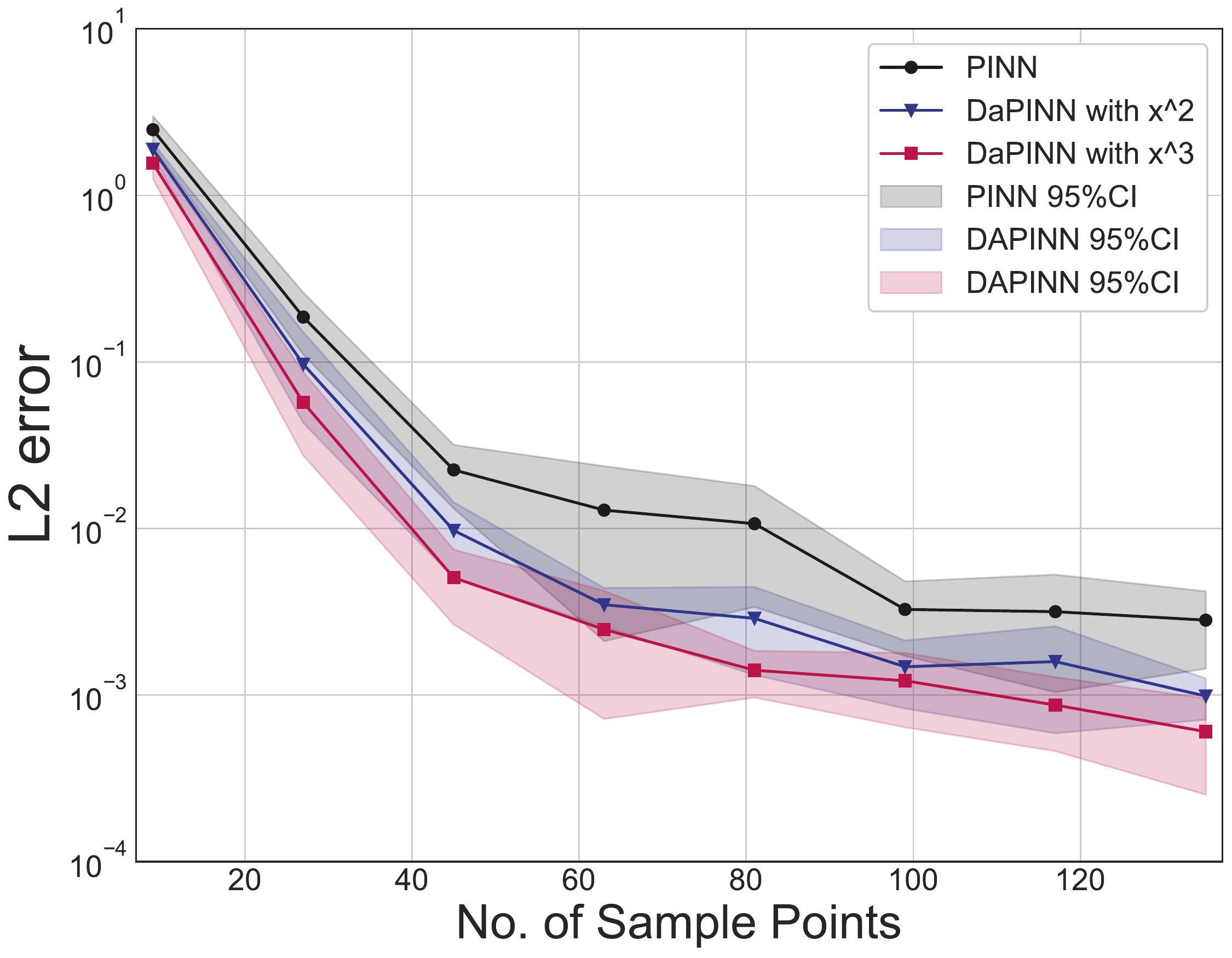}
\end{minipage}%
}%
\subfigure[]{
\begin{minipage}[t]{0.5\linewidth}
\centering
\includegraphics[width=60mm]{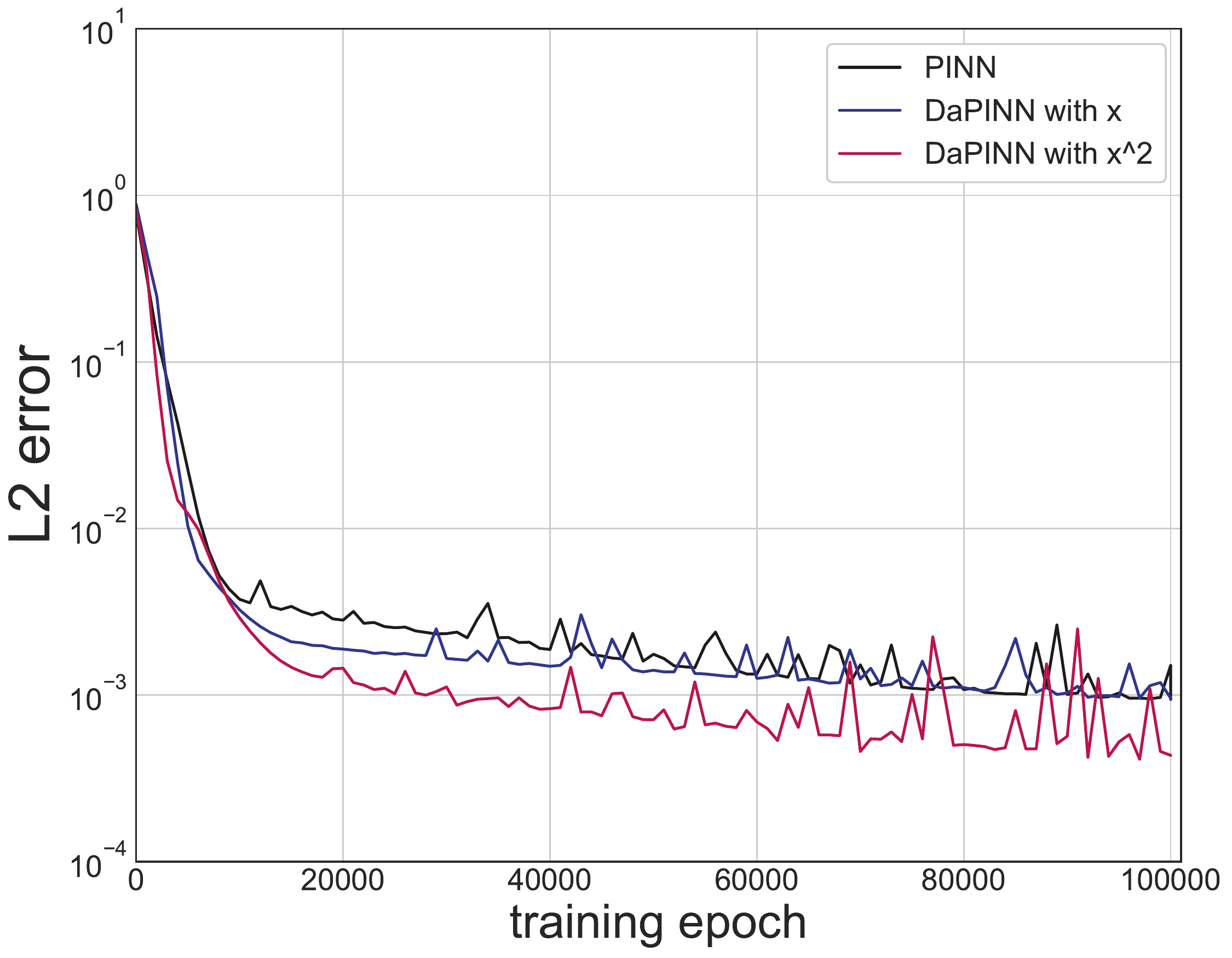}
\end{minipage}%
}%
\centering
\caption{Example in Section 3.3.1: comparison of the PINN, DaPINN with $x^2$ and DaPINN with $x^3$ models. A network size of $\left [ I, 20, 20, 20, 20, 1 \right ]$ and 10000 epochs with a learning rate of 0.001 are used, with $I=2$ for the PINN, $I=3$ for the DaPINN with $x^2$ and $I=4$ for the DaPINN with $x^3$. (a) Variation in the L2 relative error of the prediction function u with the number of samples for the PINN, DaPINN (second-order power series), and DaPINN (third-order power series) models. (b) Variation in the loss with the number of training epochs when the number of sample points is fixed at 135.}
\end{figure}
\subsubsection{Burgers' problem}
In this section, we discuss a Burgers' problem
We consider the 1D Burgers' equation:
\begin{equation}
\frac{\partial u}{\partial t}+u\frac{\partial u}{\partial x}-\frac{0.01}{\pi}\frac{\partial^2u}{\partial x^2}=0,\ \ x\epsilon\left[-1,1\right],\space t\epsilon\left[0,1\right]
\end{equation}
The initial boundary conditions are
\begin{equation}
u\left(x,0\right)=sin{\pi x}
\end{equation}
\begin{equation}
u\left(-1,t\right)=u\left(1,t\right)=0
\end{equation}
We also use the DaPINN with third-order power series augmentation, and the results are shown in Fig. 7(a). In the Burgers' problem, the advantage of the DaPINN model is still apparent. The error of the DaPINN model using the third-order power series augmentation is one order of magnitude less than that of the PINN model. Fig. 7(b)(c)(d) shows DaPINN prediction functions when the number of sample points is 3500. The PINN results are shown in Fig. 7 (e)(f)(g). The DaPINN still has high accuracy in regions with drastic changes, while the PINN fails. This result implies that the input dimension expansion makes it easier for the neural network to extract the target problem features; thus, the DaPINN uses required resources to obtain better results.
\begin{figure}[htbp]
\centering
\subfigure[]{
\begin{minipage}[t]{1\linewidth}
\centering
\includegraphics[width=80mm]{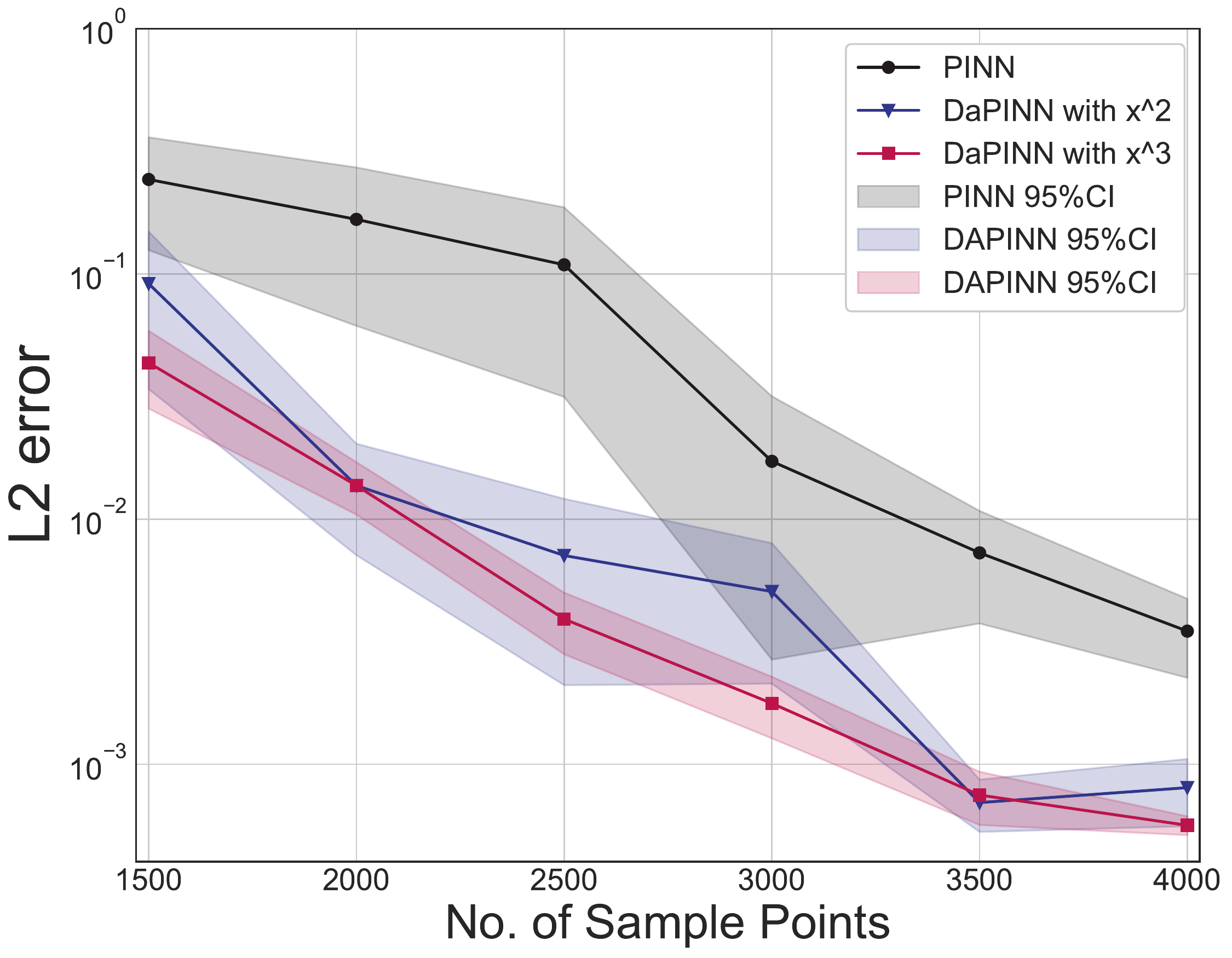}
\end{minipage}%
}%

\subfigure[]{
\begin{minipage}[t]{0.33\linewidth}
t=0.25
\centering
\includegraphics[width=50mm]{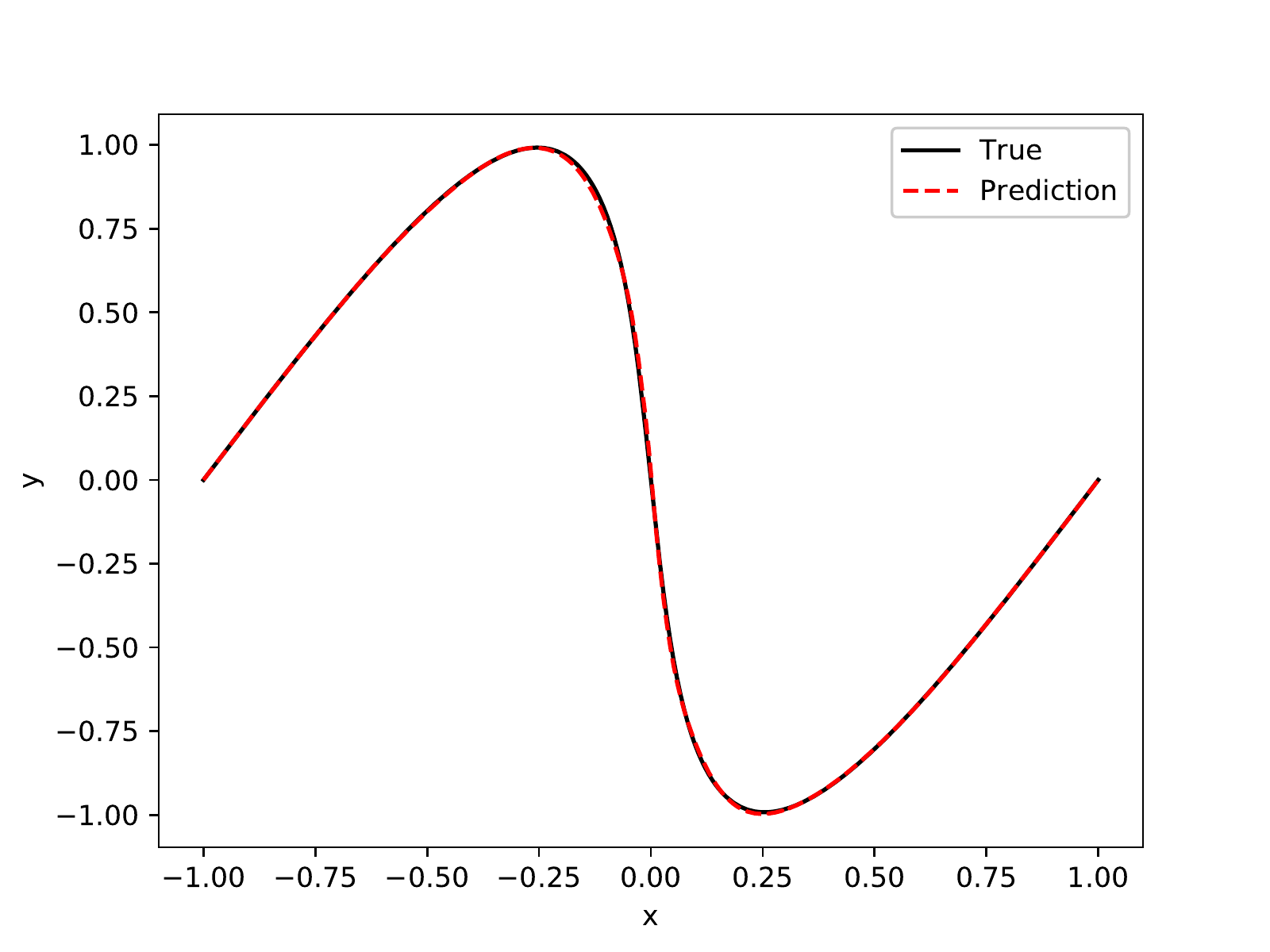}
\end{minipage}%
}%
\subfigure[]{
\begin{minipage}[t]{0.33\linewidth}
t=0.50
\centering
\includegraphics[width=50mm]{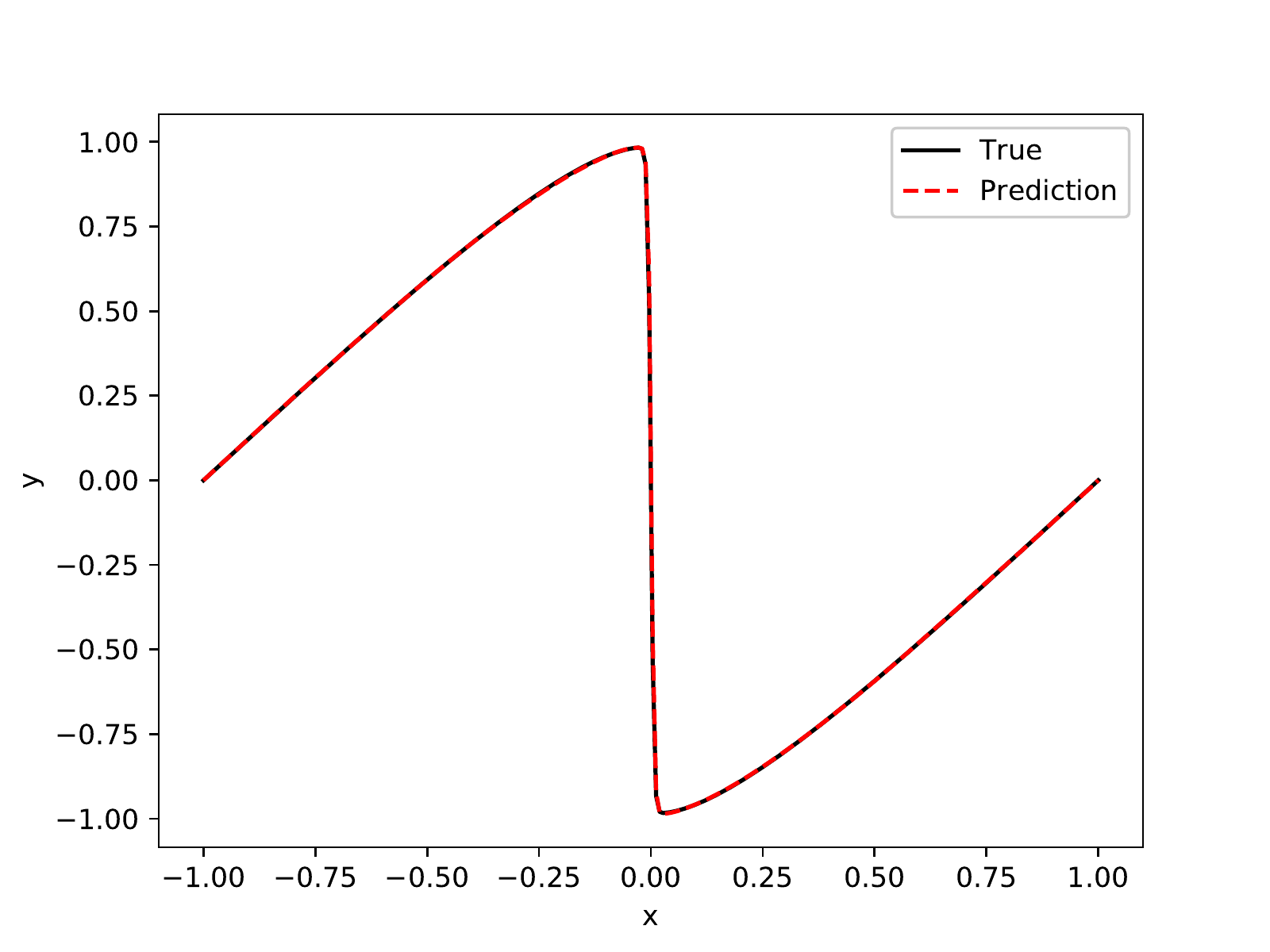}
\end{minipage}%
}%
\subfigure[]{
\begin{minipage}[t]{0.33\linewidth}
t=0.75
\centering
\includegraphics[width=50mm]{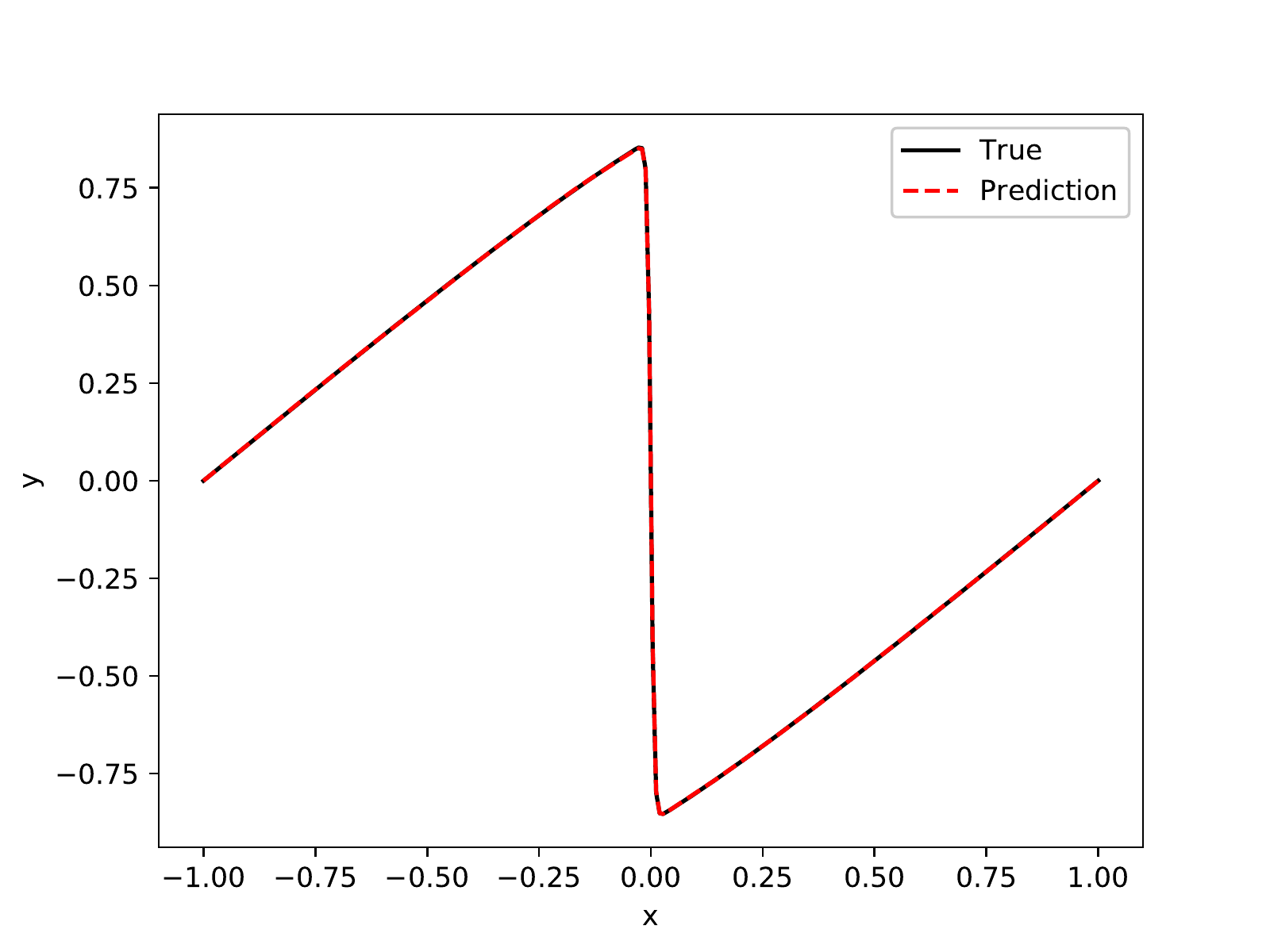}
\end{minipage}%
}%

\subfigure[]{
\begin{minipage}[t]{0.33\linewidth}
\centering
\includegraphics[width=50mm]{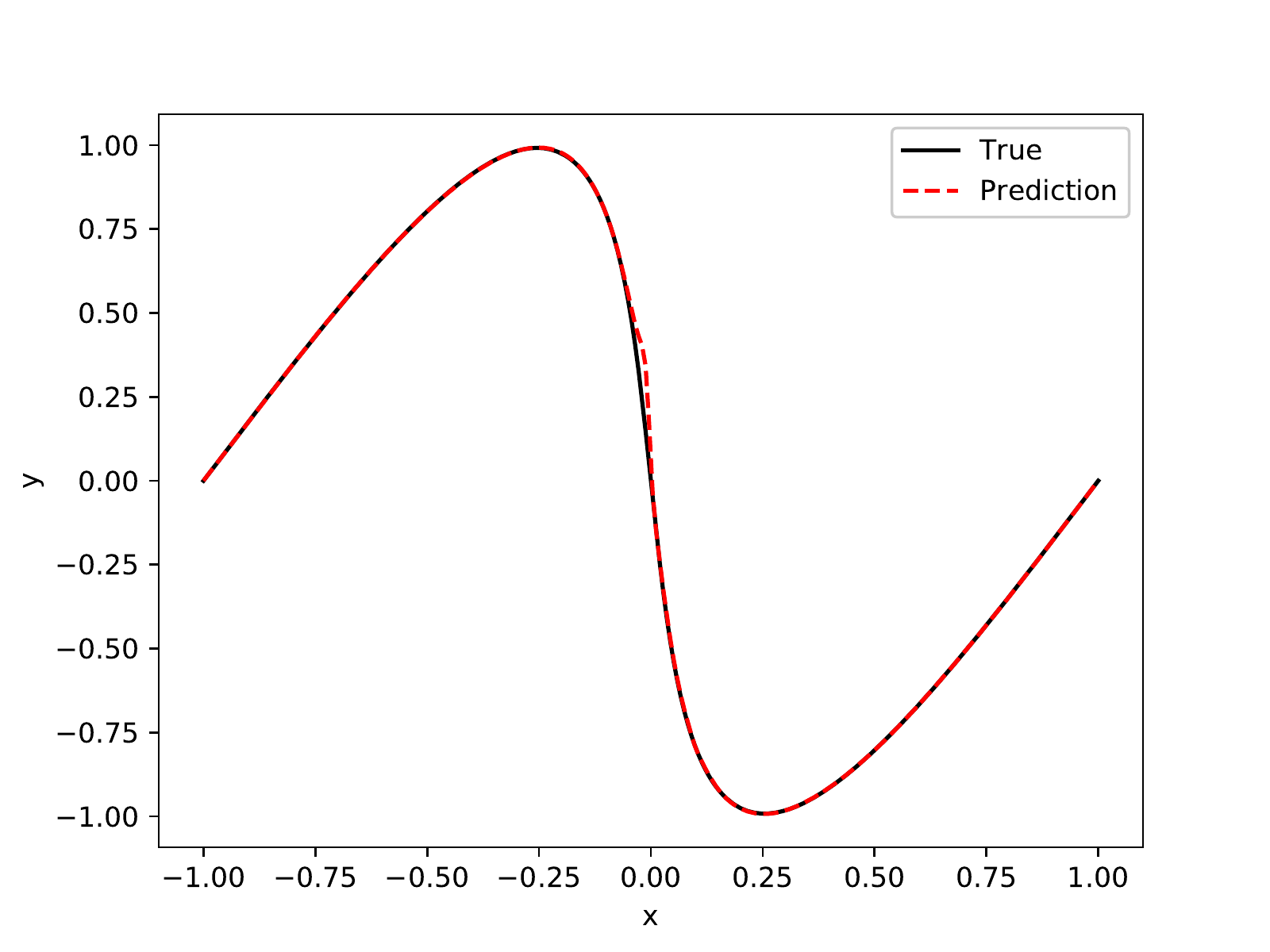}
\end{minipage}%
}%
\subfigure[]{
\begin{minipage}[t]{0.33\linewidth}
\centering
\includegraphics[width=50mm]{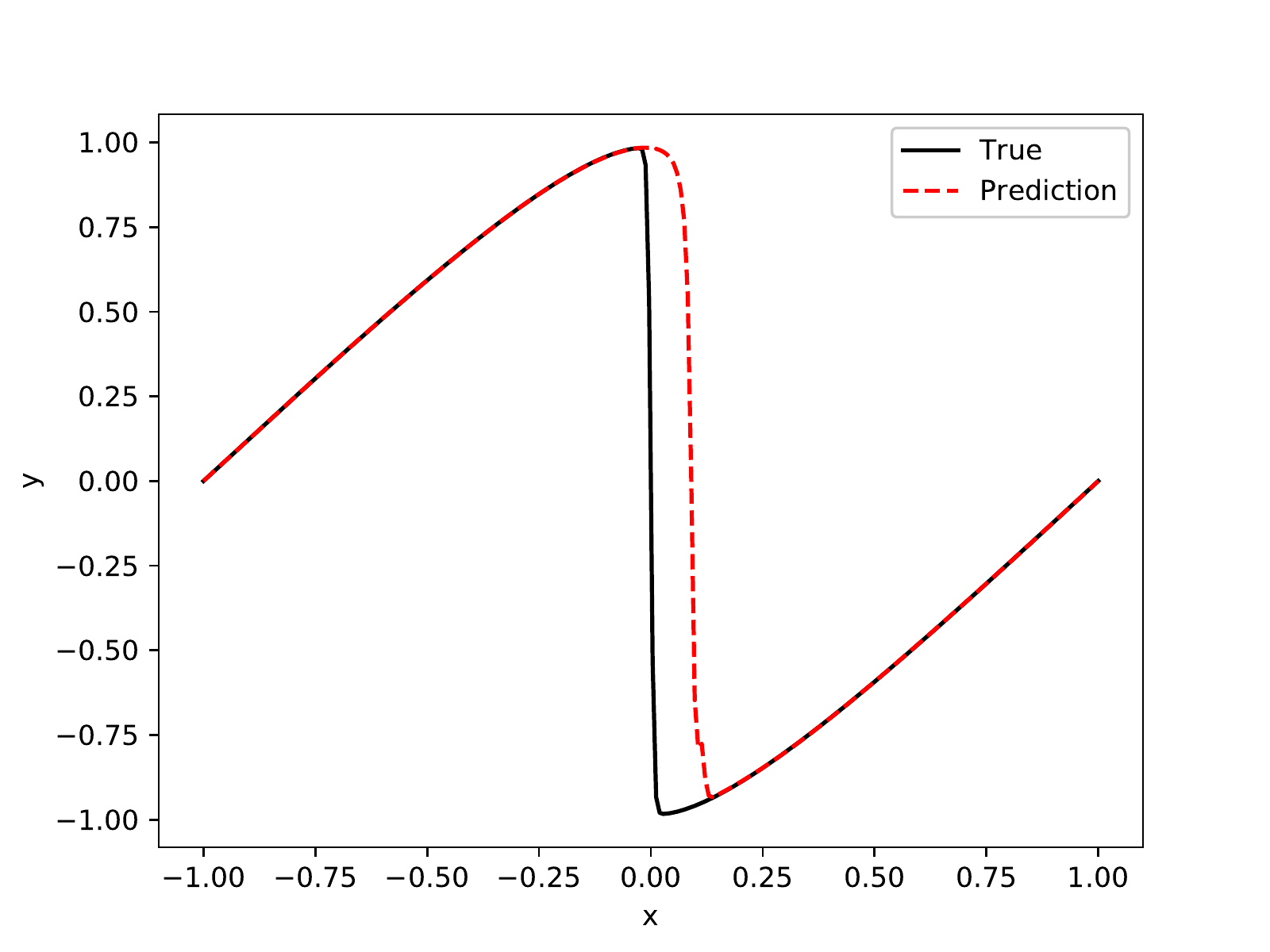}
\end{minipage}%
}%
\subfigure[]{
\begin{minipage}[t]{0.33\linewidth}
\centering
\includegraphics[width=50mm]{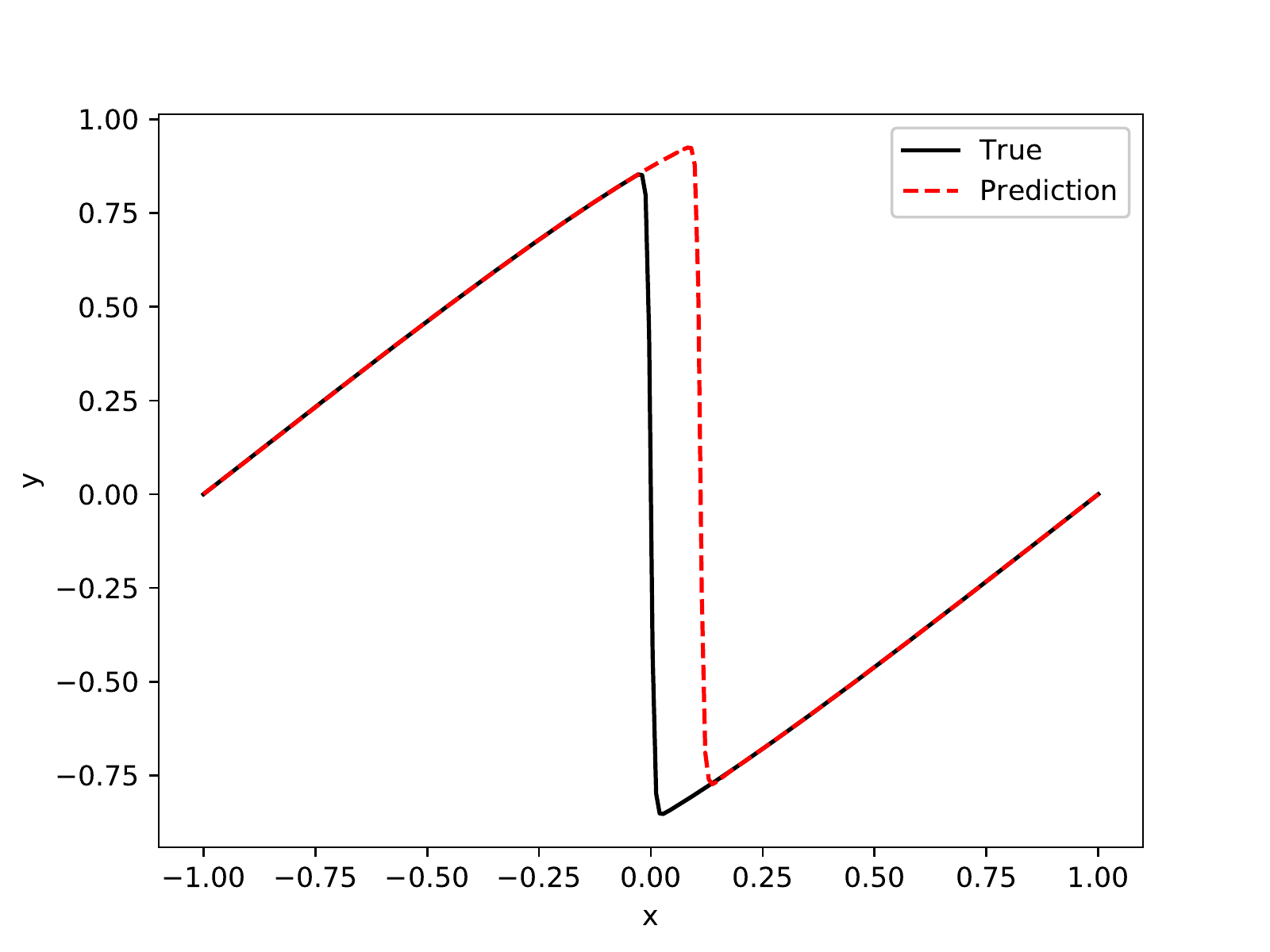}
\end{minipage}%
}%
\centering
\caption{Example in Section 3.3.2: comparison of the PINN, DaPINN with $x^2$ and DaPINN with $x^3$ models. A network size of $\left [ I, 32, 32, 32, 1 \right ]$ and 10000 epochs with a learning rate of 0.001 are used, with $I=2$ for the PINN, $I=3$ for the DaPINN with $x^2$ and $I=4$ for the DaPINN with $x^3$. (a) L2 relative error of the prediction function u for the PINN, DaPINN with $x^2$, and DaPINN with $x^3$ models versus the number of samples. (b)(c)(d) DaPINN prediction function when t=0.25, 0.5, and 0.75. (e)(f)(g) PINN prediction function when t=0.25, 0.5, and 0.75}
\end{figure}
\subsubsection{ Diffusion (inverse) }
We return to the problem discussed at the end of Section 3.2.2. We resolve this problem using the DaPINN with third-order power series augmentation.

\begin{figure}[htbp]
  \centering
\subfigure[]{
\begin{minipage}[t]{0.33\linewidth}
\centering
\includegraphics[width=55mm]{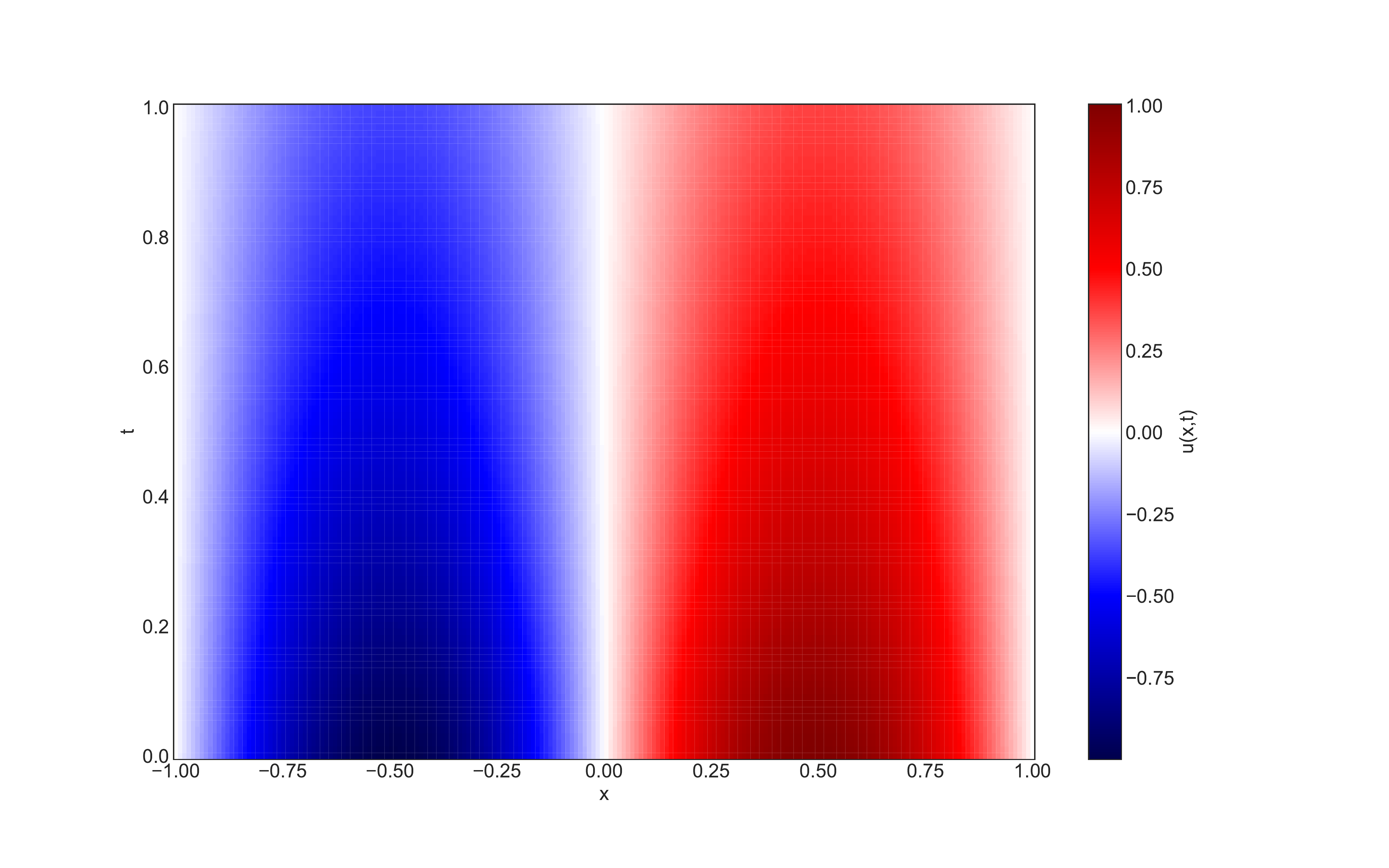}
\end{minipage}%
}%
\subfigure[]{
\begin{minipage}[t]{0.33\linewidth}
\centering
\includegraphics[width=55mm]{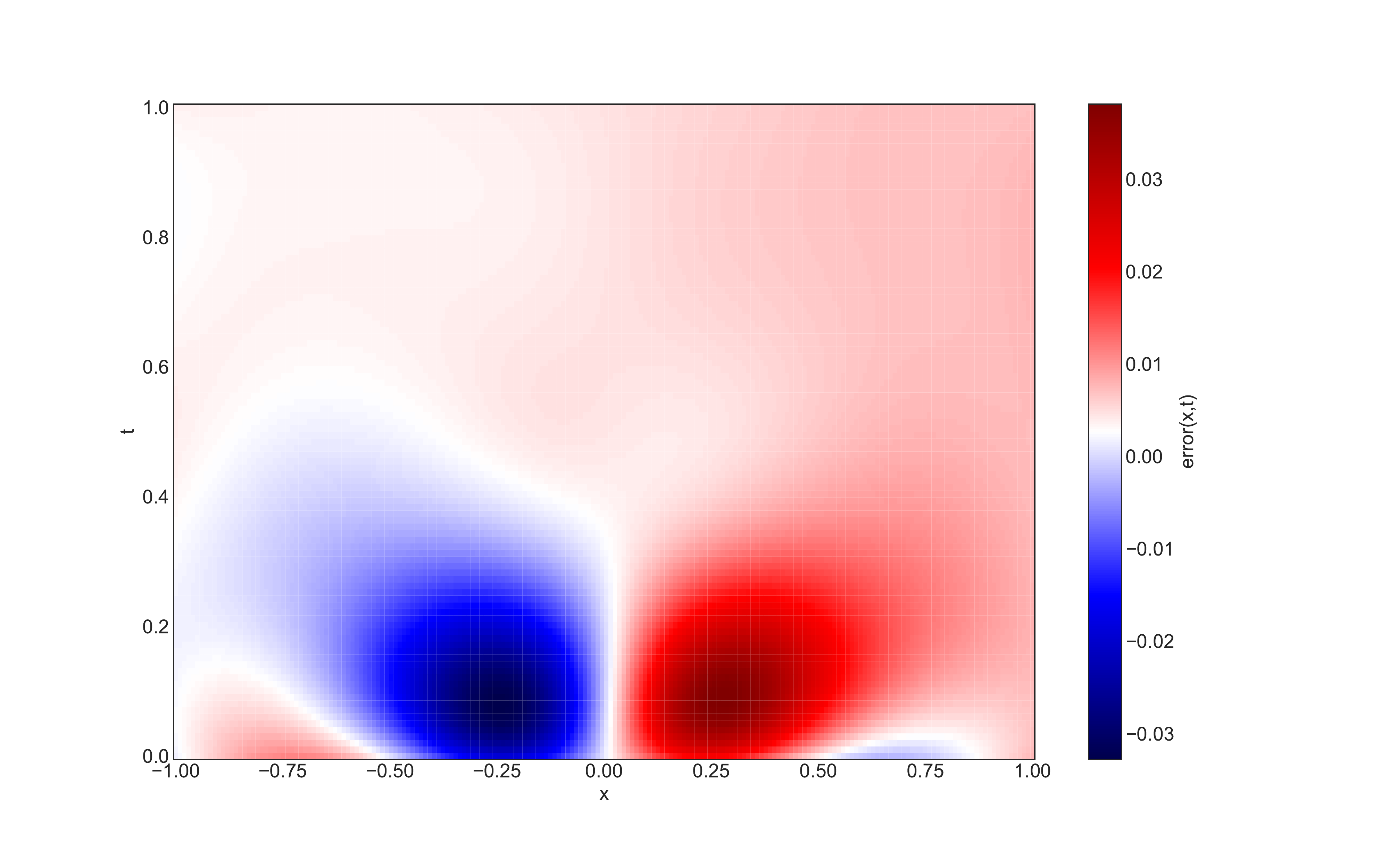}
\end{minipage}%
}%
\subfigure[]{
\begin{minipage}[t]{0.33\linewidth}
\centering
\includegraphics[width=55mm]{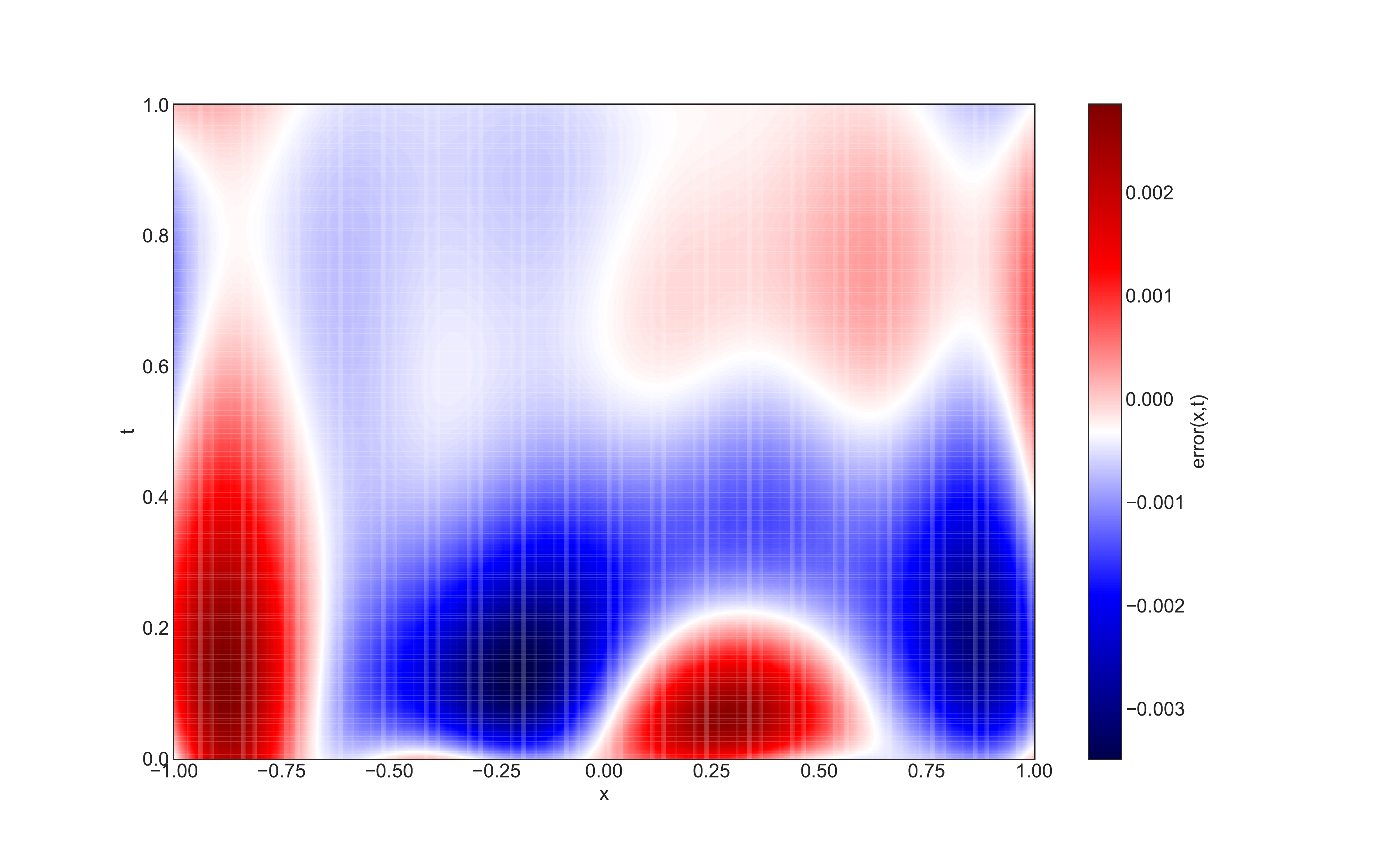}
\end{minipage}%
}%
  \centering
  \caption{Example in Section 3.3.3: (a) Absolute error distribution of the predicted solutions of the forward analytic solution of the objective function, using PINN model. (b, c) From left to right, the error distribution using DaPINN with $x^2$, DaPINN with $x^2\&x^3$.}
\end{figure}
Although the error characteristics due to dimension augmentation still exist, the introduction of the cubic term in the DaPINN model successfully resolves the concentration in the error distribution in the previous DaPINN model: the peak absolute error is reduced to averagely 10\% of its original value, and the error band near the steep region at $t=0$ is largely flattened (Fig. 8).

\subsection{Fourier series augmentation method} 	
In the previous experiments, we discussed the role of replica augmentation and power series augmentation and verified their effectiveness. In this section, we discuss the role of Fourier series augmentation and compare this approach with power series augmentation and the PINN model.
\subsubsection{Diffsion}

Here, we choose the following diffusion reaction equation:

\begin{equation}
\frac{\partial u}{\partial t}-\frac{\partial^2u}{\partial x^2}=f\left(x,t\right),\ \ x\epsilon\left[-\pi,\pi\right],t\epsilon\left[0,1\right]		
\end{equation}
where
\begin{equation}
f\left(x,t\right)=e^{-t}\left[\frac{3}{2}\sin{2x}+\frac{8}{3}\sin{3x}+\frac{15}{4}\sin{4x}+\frac{63}{8}\sin{8x}\right]
\end{equation}
The initial edge value condition is
\begin{equation}
u\left(x,0\right)=\sum_{i=1}^{4}\frac{\sin{x}}{i}+\frac{\sin{8x}}{8},\space u\left(-\pi,t\right)=u\left(\pi,t\right)=0
\end{equation}
The analytical solution of this equation is
\begin{equation}
u\left(x,t\right)=e^{-t}\sum_{i=1}^{4}\frac{\sin{x}}{i}+\frac{\sin{8x}}{8}
\end{equation}
We compare the DaPINN with third-order power series augmentation and the DaPINN with Fourier series augmentation with the PINN.

In this problem, the trigonometric function in the equation has a period of $2\pi$. As a Fourier series augmentation, we introduce $\tau_1=x, \tau_2=\sin{x}, \tau_3=\cos{x}, \tau_4=t$ as the augmented inputs with the following loss function:

\begin{multline}
\mathcal{L}_f=\frac{1}{\left|\mathcal{T}_f\right|}\sum_{\mathbf{x}\in \mathcal{T}_f}(\frac{\partial\mathcal{N}\left(\tau_1,\tau_2,\tau_3,\tau_4\right)}{\partial \tau_4}\\
-(\frac{\partial^2\mathcal{N}\left(\tau_1,\tau_2,\tau_3,\tau_4\right)}{{\partial \tau_1}^2}+{\tau_3}^2\frac{\partial^2\mathcal{N}\left(\tau_1,\tau_2,\tau_3,\tau_4\right)}{{\partial \tau_2}^2}+{\tau_2}^2\frac{\partial^2\mathcal{N}\left(\tau_1,\tau_2,\tau_3,\tau_4\right)}{{\partial \tau_3}^2}+2\tau_3\frac{\partial^2\mathcal{N}\left(\tau_1,\tau_2,\tau_3,\tau_4\right)}{\partial \tau_1\partial \tau_2}\\
-2\tau_2\frac{\partial^2\mathcal{N}\left(\tau_1,\tau_2,\tau_3,\tau_4\right)}{\partial \tau_1\partial \tau_3}-2\tau_2\tau_3\frac{\partial^2\mathcal{N}\left(\tau_1,\tau_2,\tau_3,\tau_4\right)}{\partial \tau_2\partial \tau_3}-\tau_2\frac{\partial\mathcal{N}\left(\tau_1,\tau_2,\tau_3,\tau_4\right)}{\partial \tau_2}\\
-\tau_3\frac{\partial\mathcal{N}\left(\tau_1,\tau_2,\tau_3,\tau_4\right)}{\partial \tau_3})-f\left(\tau_1,\tau_4\right))
\end{multline}

As shown in Fig. 9(a), when the number of training points is 65, the L2 relative error of the PINN is 103\%, the L2 relative error of both DaPINNs is about 1\%; thus, the errors of the two methods differ by two orders of magnitude. As shown in Fig. 9(b), when the number of training epochs is 5000, the L2 error of the DaPINN with $x^3$ is 0.9\%, the L2 error of the DaPINN with Fourier series augmentation is 1.5\%, and the L2 error of the PINN is 7.8\%. When the number of training epochs is 10000, the L2 errors of the DaPINN with $x^3$ and the DaPINN with Fourier series augmentation decrease to 0.2\% and 0.3\%, respectively, and the PINN L2 error decreases to 2.5\%.
Thus, we find that the Fourier series augmentation in this problem has similar properties to the third-order power series augmentation.

However, when Figs. 9(g), 9(i) and 9(h) are compared, we find that the DaPINN errors are not only smaller but also more uniformly distributed than the PINN errors, and the error peaks are not obvious. This result occurs because the DaPINN with higher-order power series augmentation better fits nonlinearities, thus making it possible to solve the problem of accuracy degradation in the PINN for complex functions.

Periodicity can be introduced by expanding the input with Fourier series, and the period in the expansion is determined based on the equation. In this diffusion problem, the equation contains $\sin{x}$, $\sin{2x}$ and other periodic functions; thus, the expansion uses $\sin{x}$ and $\cos{x}$ to ensure the validity of the period. This is equivalent to the period of the solution provided to the neural network. Moreover, the learning effect clearly improves when using Fourier series to expand the dimension of the DaPINN model.
\begin{figure}[htbp]
\centering
\subfigure[]{
\begin{minipage}[t]{0.5\linewidth}
\centering
\includegraphics[width=60mm]{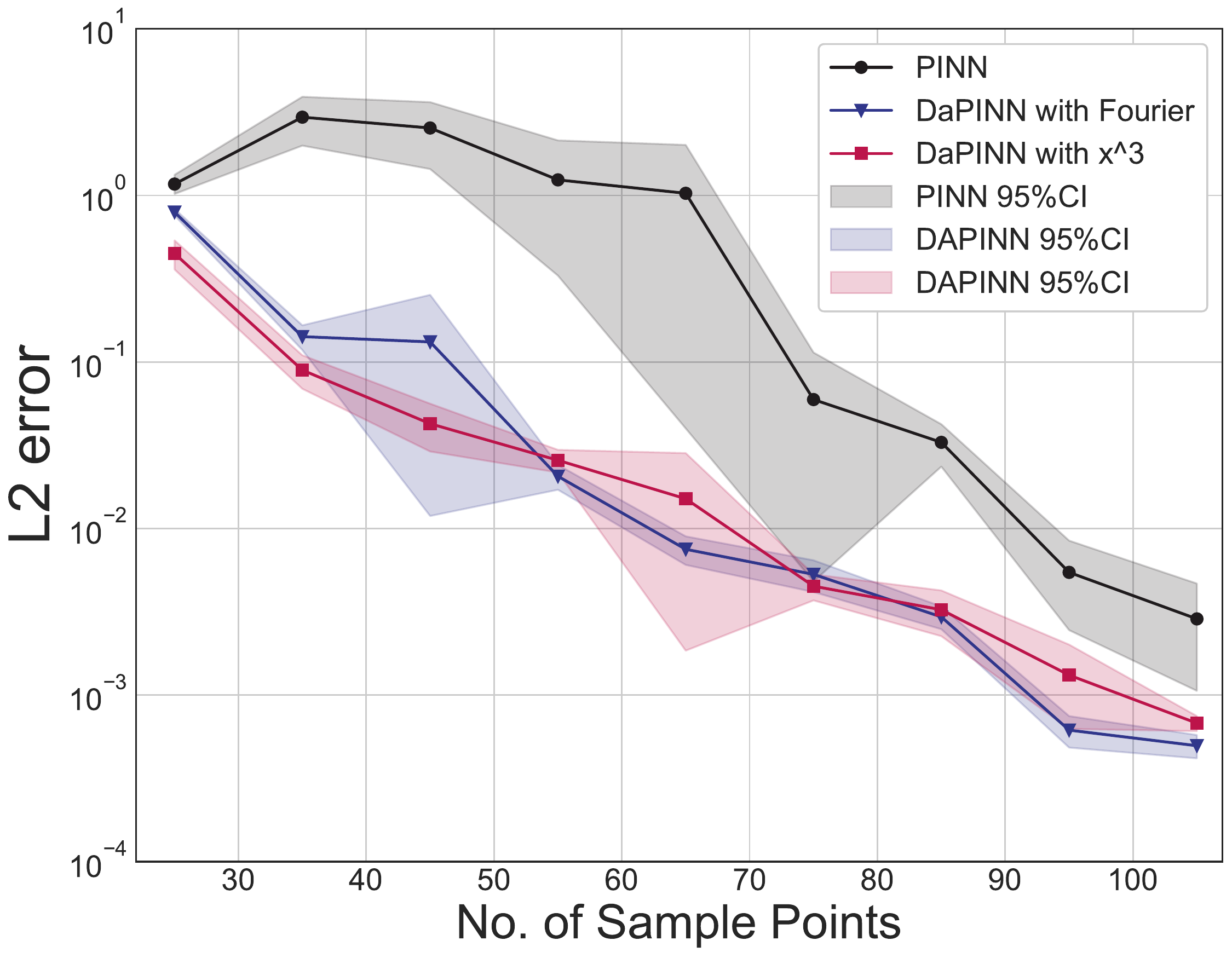}
\end{minipage}%
}%
\centering
\subfigure[]{
\begin{minipage}[t]{0.5\linewidth}
\centering
\includegraphics[width=60mm]{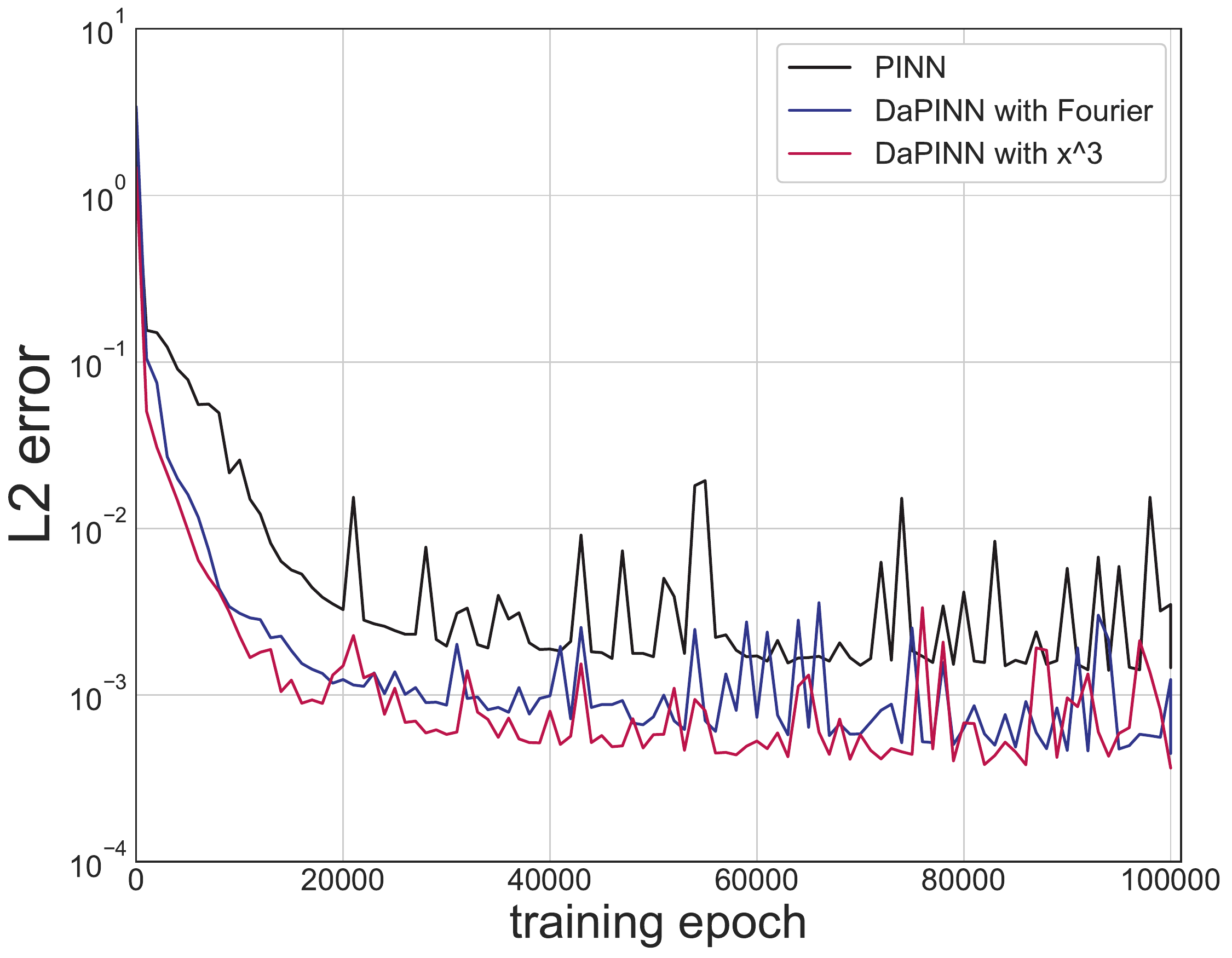}
\end{minipage}%
}%

\centering
\subfigure[]{
\begin{minipage}[t]{1\linewidth}
\centering
\includegraphics[width=80mm]{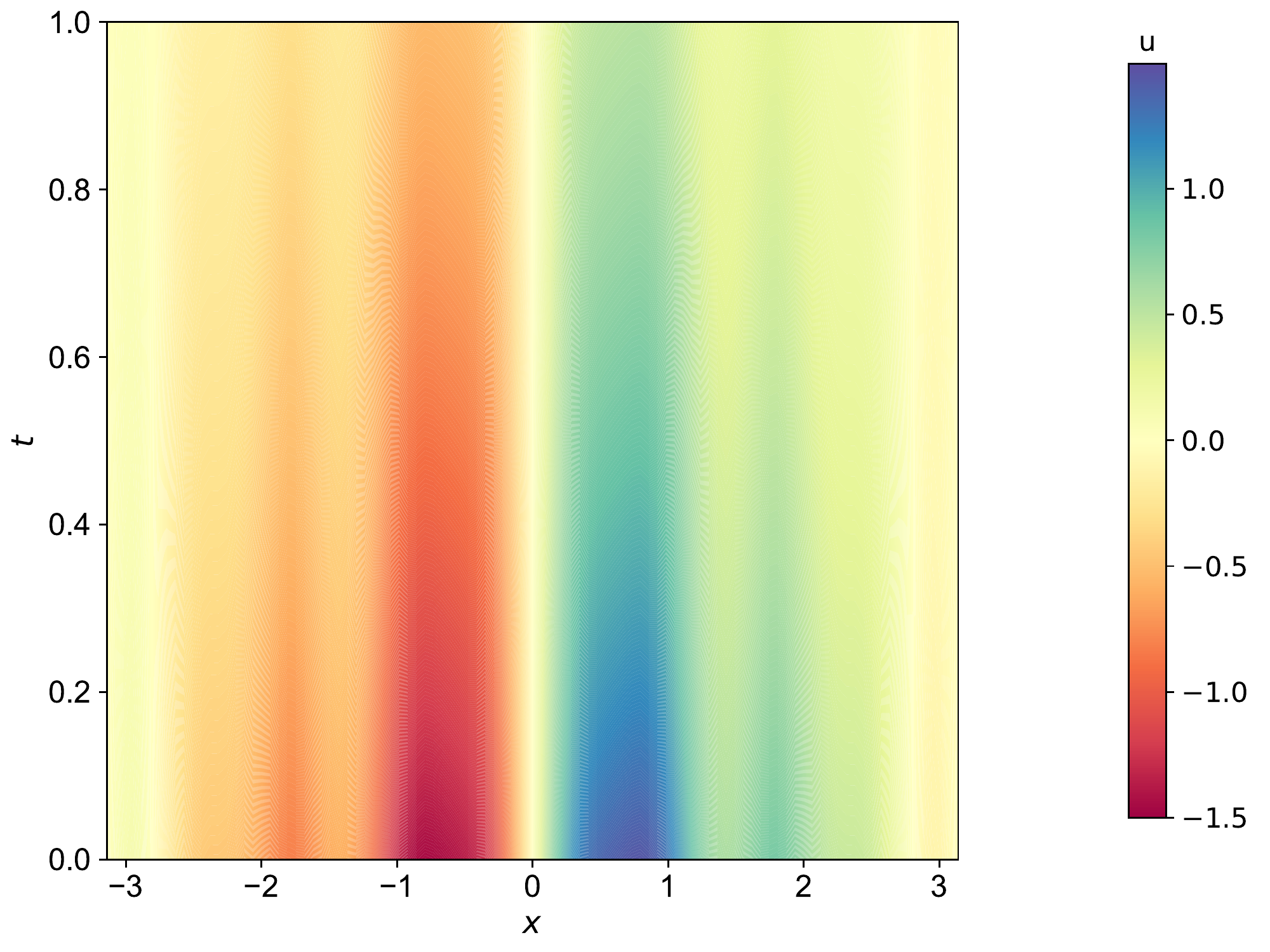}
\end{minipage}%
}%

\subfigure[]{
\begin{minipage}[t]{0.33\linewidth}
\centering
\includegraphics[width=50mm]{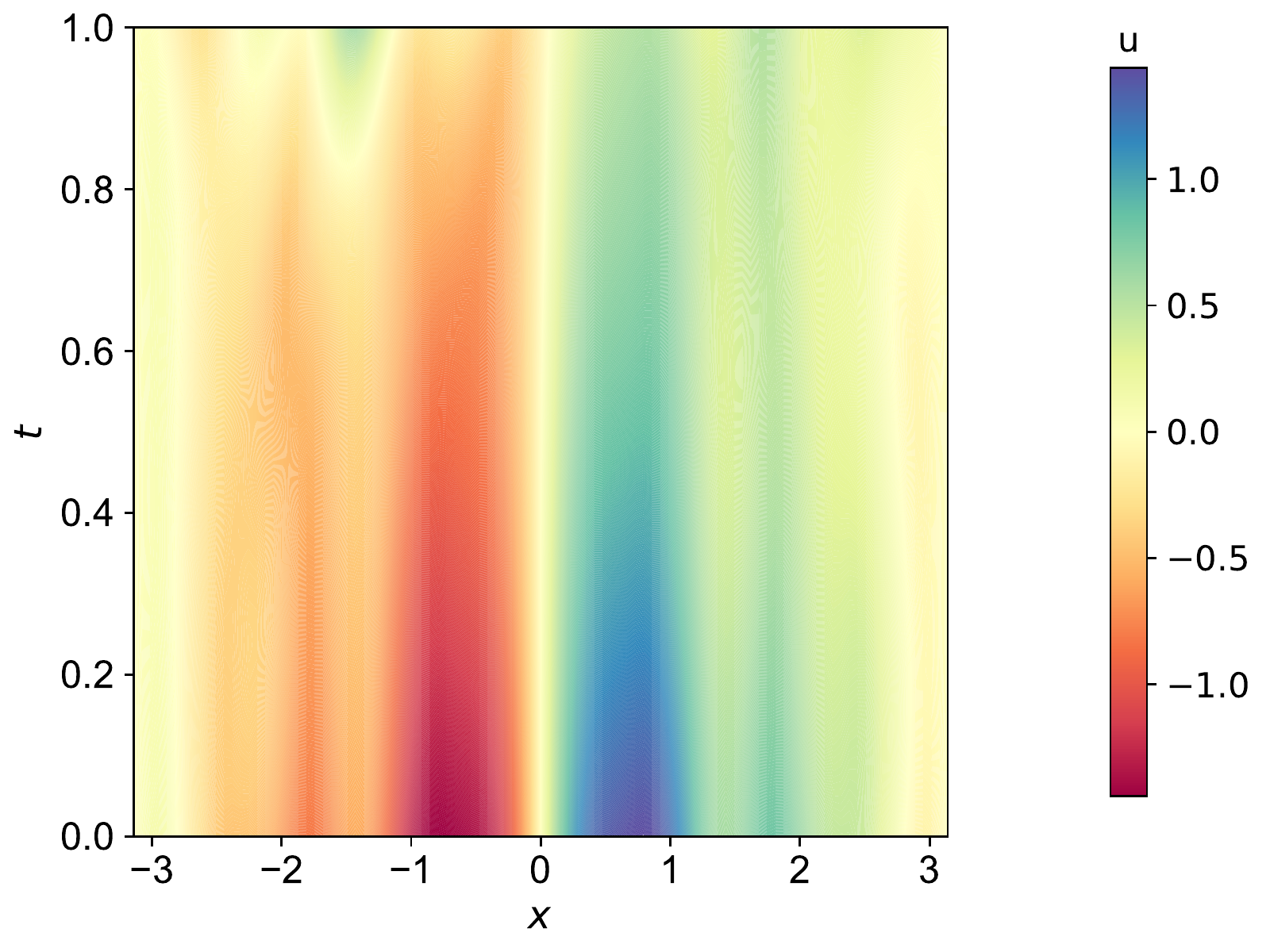}
\end{minipage}%
}%
\subfigure[]{
\begin{minipage}[t]{0.33\linewidth}
\centering
\includegraphics[width=50mm]{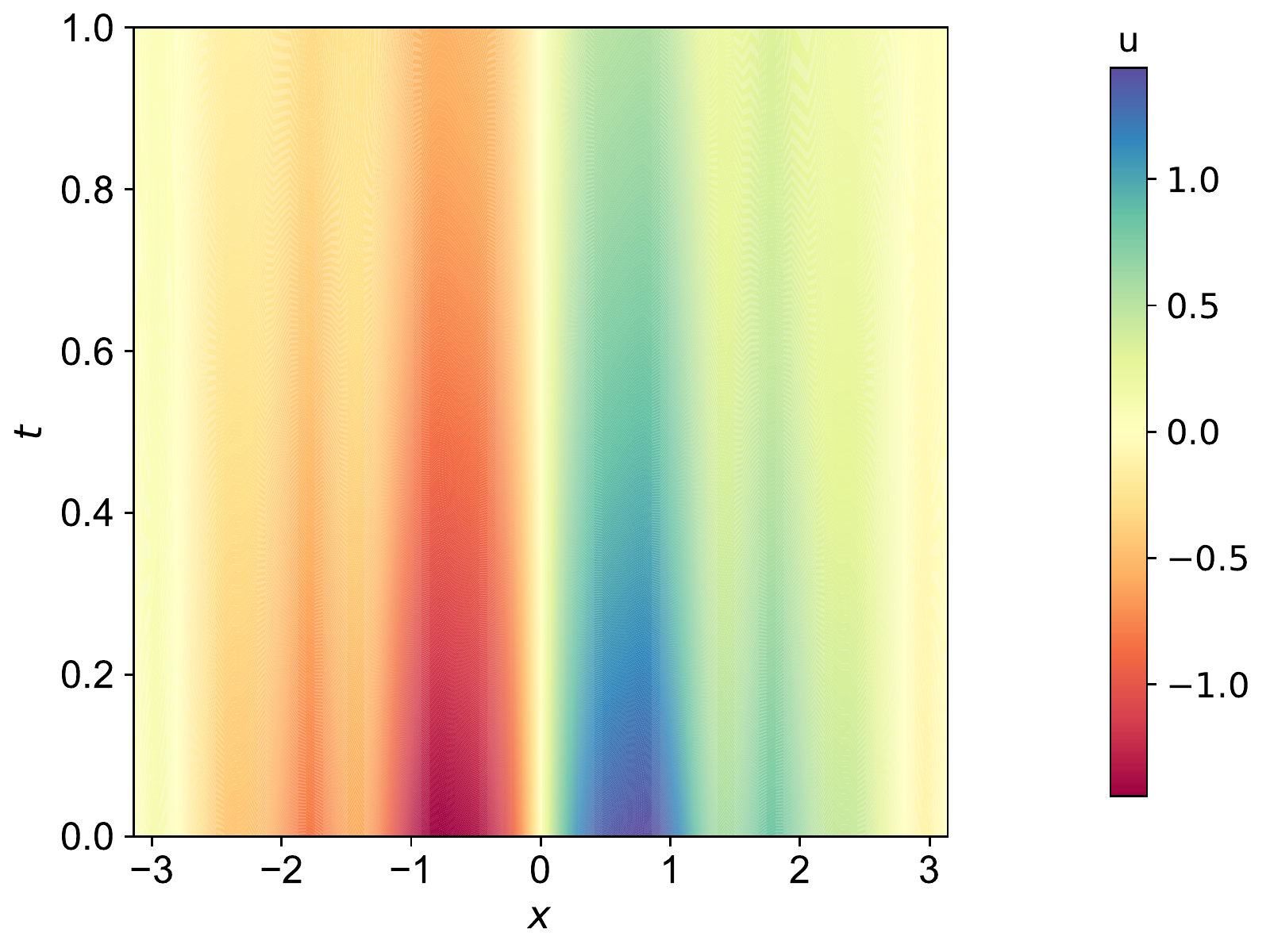}
\end{minipage}%
}%
\subfigure[]{
\begin{minipage}[t]{0.33\linewidth}
\centering
\includegraphics[width=50mm]{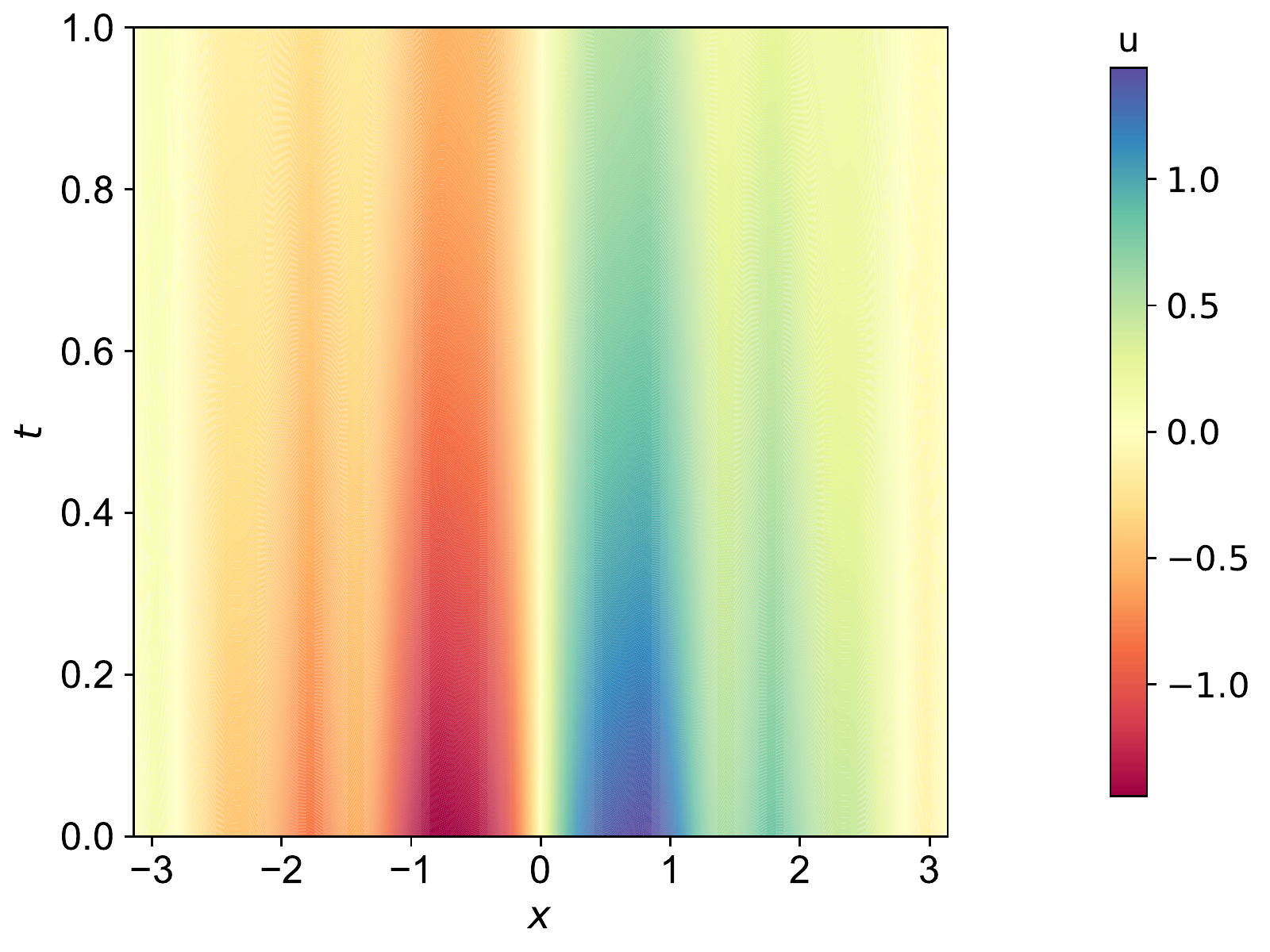}
\end{minipage}%
}%
\end{figure}
\begin{figure}
\subfigure[]{
\begin{minipage}[t]{0.333\linewidth}
\centering
\includegraphics[width=50mm]{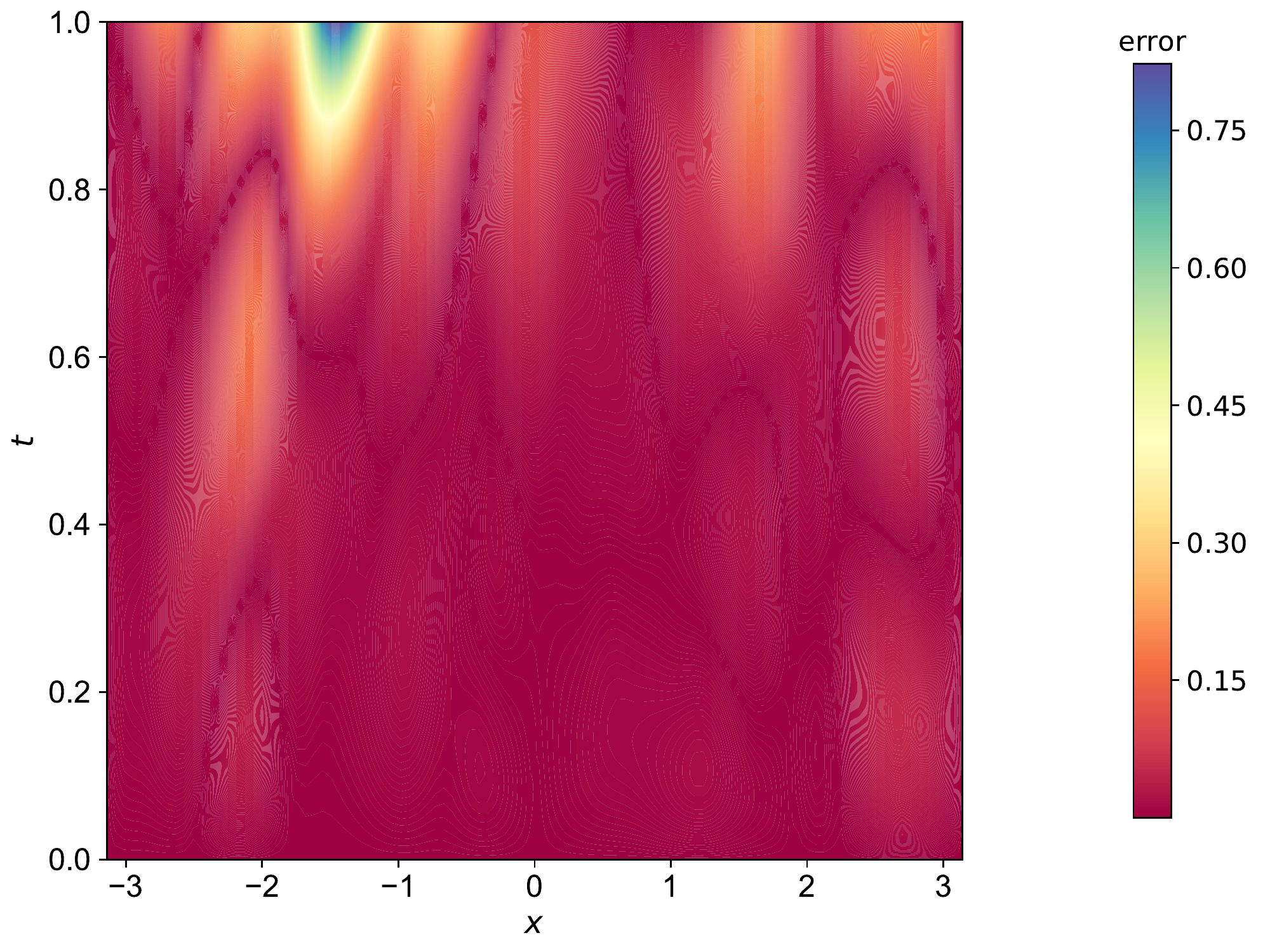}
\end{minipage}%
}%
\subfigure[]{
\begin{minipage}[t]{0.333\linewidth}
\centering
\includegraphics[width=50mm]{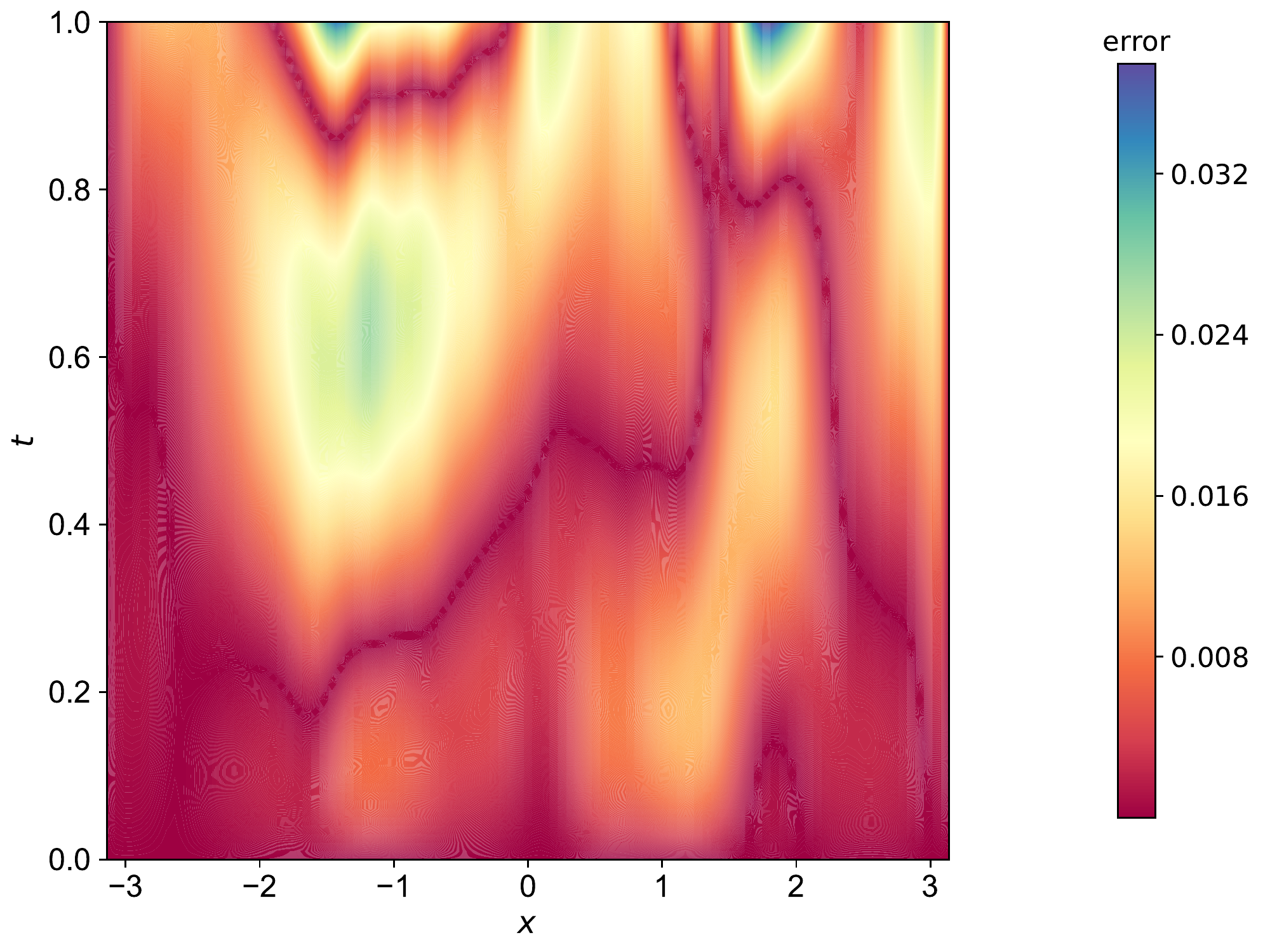}
\end{minipage}%
}%
\subfigure[]{
\begin{minipage}[t]{0.333\linewidth}
\centering
\includegraphics[width=50mm]{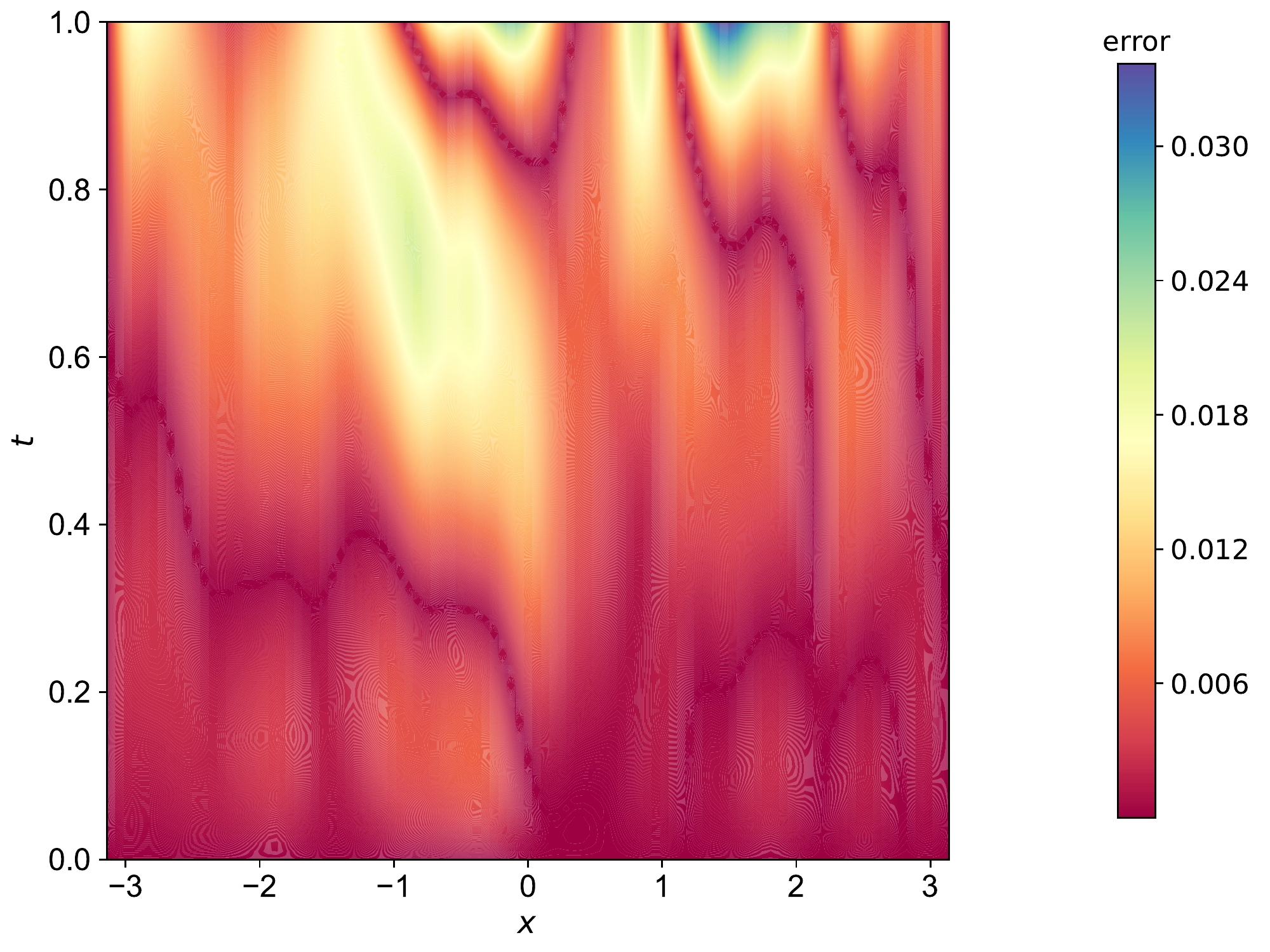}
\end{minipage}%
}%
\centering
\caption{Example in Section 3.3.1: comparison of the PINN, DaPINN with Fourier series augmentation and DaPINN with $x^3$ models. A network size of $\left [ I, 20, 20, 20, 1 \right ]$ and 20000 epochs with a learning rate of 0.001 are used, with $I=2$ for the PINN and $I=4$ for the DaPINNs. (a) Variation in the L2 relative error of the prediction functions u of the PINN and DaPINN models versus the number of samples. (b) L2 relative error of the prediction functions u of the PINN and DaPINN models with 115 sample points as a function of the number of training epochs. (c)(d)(e)(f)(g)(h)(i) Comparison between the PINN and DaPINN models trained using 60 residuals. (c) Analytic solution results. (d)(g) PINN results and their absolute errors. (e)(h) Power series augmentation DaPINN results and their absolute errors. (f)(i) Fourier series augmentation DaPINN results and their absolute errors.}
\end{figure}

\subsubsection{Allen Cahnn equation}
Here, we investigate the Allen Cahn equation:
\begin{equation}
\frac{\partial u}{\partial t}=d\frac{\partial^2u}{{\partial x}^2}+5(u-u^3),x\in[-1,1],t\in[0,1]
\end{equation}
The initial edge value condition is
\begin{equation}
u(x,0)=x^2\cos{\pi x}
\end{equation}
\begin{equation}
u(-1,t)=u(1,t)=-1
\end{equation}
We compare DaPINNs with third-order power series augmentation and Fourier series augmentation with the PINN. In the Fourier series augmentation, we introduce $\sin{\pi x}$ and $\cos{\pi x}$ because we know the period information from the initial condition.

The results are shown in Fig. 10. The PINN error is 11.2\% when the number of training points reaches 540, the L2 error is 1.5\% for the DaPINN with $x^3$, and the L2 error is 2.1\% for the DaPINN with Fourier series augmentation.
\begin{figure}[htbp]
  \centering
  \includegraphics[width=100mm]{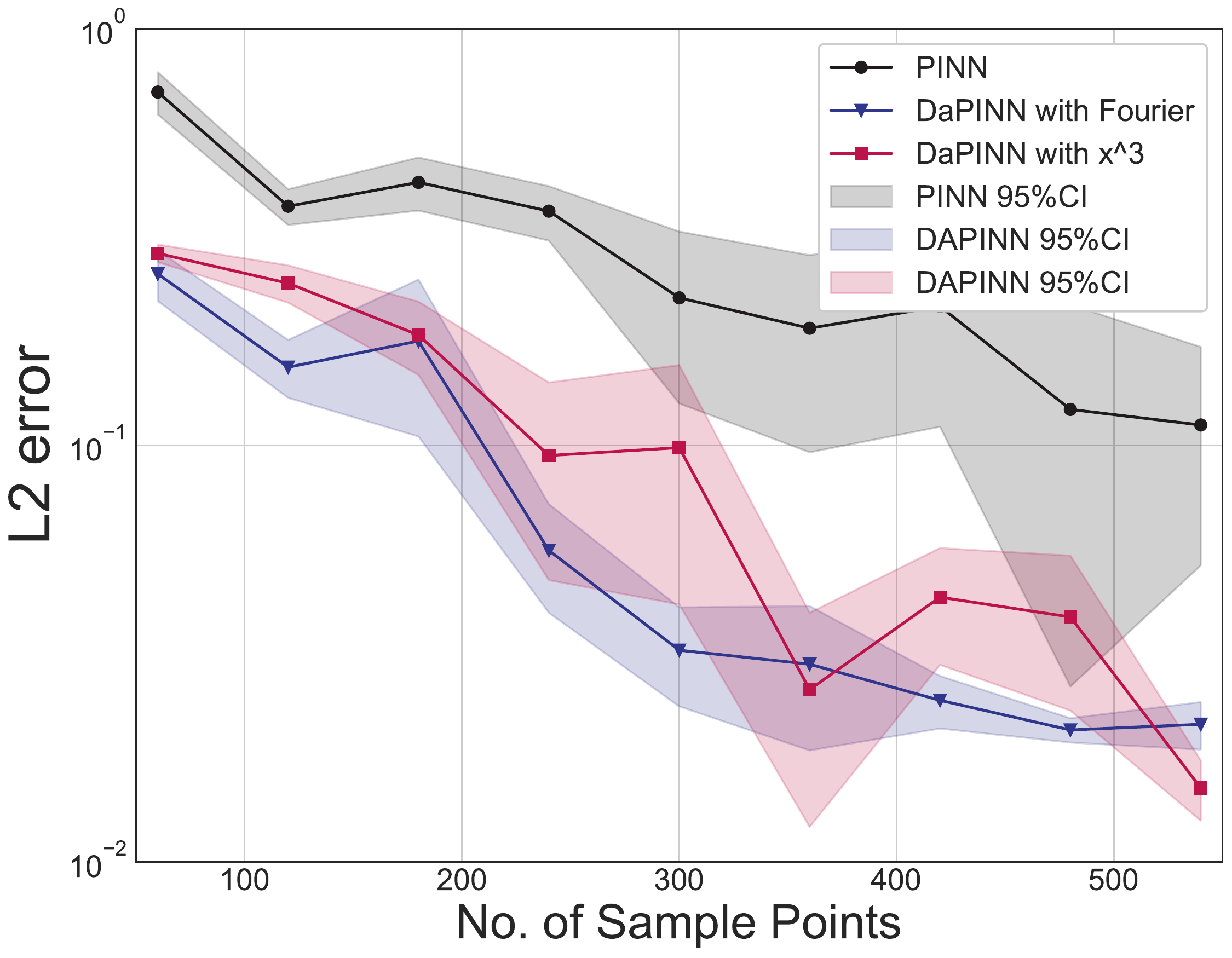}
  \caption{Example in Section 3.4.2: comparison of the PINN, DaPINN with Fourier series augmentation and DaPINN with $x^3$ Models. A network size of $\left [ I, 20, 20, 20, 1 \right ]$ and 20000 epochs with a learning rate of 0.001 are used, with $I=2$ for the PINN and $I=4$ for the DaPINNs. The L2 relative errors of the prediction functions u of the PINN and DaPINN models with respect to the number of samples are shown.}
\end{figure}
\section{Discussion}
We showed that the DaPINN with power series augmentation outperforms the PINN in solving partial differential equations under the same training conditions (number of training points and number of epochs), especially for complex equations with steep solutions. In the following section, we discuss the improvement in the computational complexity with the DaPINN approach and the compatibility of the proposed network with other optimization methods.
\subsection{Computational complexity} 	
We investigated the increase in the computational complexity of the DaPINN. While the PINN and DaPINN models in this work both use the FNN architecture, the DaPINN changes the network, mainly by increasing the network input width.

The number of model parameters is an important measure of space complexity. Consider an FNN with an input dimension of i, an output dimension of o and n hidden layers with widths of $m_j$. The number of model parameters is
\begin{equation}
Param=\left ( i+1 \right )m_1+ \sum_{j=1}^{n-1} (m_j+1)m_{j+1}+\left ( m_n+1 \right )o
\end{equation}

It is clear that the number of model parameters is more sensitive to m and n than to i. In Section 3.3, we solved the Burgers' equation with a PINN with i=2, m=32, and n=4; thus, the number of model parameters was 3297. After the DaPINN doubles the input width, the number of model parameters increases to 3361. Thus, the number of model parameters increases by approximately 1.9\%.

The number of floating point operations (FLOPs) in a model is an important measure of the time complexity. The number of FLOPs in an FNN can be calculated with the following equation, where $\alpha$ is the number of FLOPs required by the activation function. For tanh, $\alpha$ varies from 2 to 20 depending on the input value.

\begin{equation}
Flop=2\left (i m_1+ \sum_{j=1}^{n-1} m_jm_{j+1}+ m_no \right ) +\bar{\alpha } \sum_{j=1}^{n} m_j
\end{equation}

In Section 3.3, the number of FLOPs was 7616 for the PINN and 7747 for the DaPINN; thus, the number of FLOPs increased by approximately 1.6\%. Here, $\bar{\alpha}$ is roughly estimated to be 10.

The results indicate that the DaPINN model has essentially the same complexity as the PINN model. Moreover, the larger the number of neurons in the network, the closer the complexity of the two methods.

The above discussion of the time complexity does not take training into account. Since the time complexity is difficult to quantify, the time complexity is shown in terms of the actual operation time in Table 1. The test platform CPU is an Intel Core i9-10900K, and the GPU is an NVIDIA Ge RTX 3090. The computational cost of the DaPINN is approximately 40\% higher than that of PINN after completing the same number of training epochs.
\begin{table*}[h]
	\centering
\begin{threeparttable}
  \centering
  \scalebox{1.2}{
  \centering
  \begin{tabular}{cccc}
  \toprule 

      Question & PINN(s) & DaPINN(s) & Relative Cost  \\\hline 
      \midrule 
      1dpossion & 27.82 & 37.49 & 1.35  \\ 
      2dpossion & 32.43 & 44.4 & 1.41  \\ 
      heat & 137.95 & 191.33 & 1.39  \\ 
      diffusion & 137.16 & 188.05 & 1.37  \\ 
      burgers & 58.74 & 85.31 & 1.45  \\ 
      \bottomrule 
  \end{tabular}}
  \\
  Table1.Training time and relative relationship between PINN and DaPINN with the same hyperparameters
\end{threeparttable}
\end{table*}

However, as demonstrated above, the DaPINN achieves the same accuracy with fewer training sessions than the PINN. Table 2 compares the time consumed by the DaPINN and PINN models to achieve the same accuracy. The DaPINN requires significantly less time than the PINN to achieve the same precision, although the DaPINN takes more time to complete each epoch. Therefore, even considering the additional computational cost of the DaPINN, the training cost of the DaPINN is significantly smaller than that of the PINN.

\begin{table*}[h]
\resizebox{0.85\textwidth}{!}{
\begin{threeparttable}
\scalebox{1.1}{
  \centering
  \begin{tabular}{ccccccc}
\hline
\toprule 
\multirow{2}{*}{Error} &  \multicolumn{3}{c}{2d Poison} &\multicolumn{3}{c}{Diffusion} \\ \cmidrule(lr){2-4}\cmidrule(lr){5-7}\morecmidrules\cmidrule(lr){2-4}\cmidrule(lr){5-7}
 
 & PINN (s) & DaPINN (s) & Relative Time & PINN (s) & DaPINN (s) & Relative Time  \\ \hline
\midrule 
10\% & 8.0 & 8.8 & 110\% & 5.8 & 1.1 & 18\%  \\ 
5\% & 9.3 & 9.7 & 104\% & 11.6 & 2.1 & 18\%  \\ 
1\% & 12.0 & 13.2 & 110\% & 18.9 & 10.7 & 57\%  \\ 
0.5\% & 22.6 & 15.8 & 70\% & 23.9 & 15.0 & 63\%  \\ 
0.1\% & $\backslash$ & 82.7 & ~ & 145.1 & 34.3 & 24\%  \\ 
0.05\% & $\backslash$ & $\backslash$ & ~ & $\backslash$ & 88.0 &   \\ \hline
\bottomrule 
\end{tabular}}
\\
\centering
Table2.The training time required for the DaPINN and PINN models to reach the same precision. $"\setminus"$ indicates that the model cannot reach that level of precision. 
\end{threeparttable}}
\end{table*}

\subsection{Effect of the width and depth of the neural network on the DaPINN performance} 
Because the hyperparameters in a neural network affect the performance of the network, it is important to determine the effect of the hyperparameters on the model. Previously, we investigated the effects of the number of training points and number of epochs on the DaPINN and PINN performance. Next, we discuss the effect of the width (number of neurons in each hidden layer) and depth (number of hidden layers) of the neural network on the network performance. In the PINN, small and large networks may lead to poor reproducibility or overfitting \cite{HE2020103610}. We used 400 sample points and 10,000 epochs to test DaPINN and PINN models with different widths and depths in the example discussed in Section 3.4.1, as shown in FIG. 11. The PINN and DaPINN both achieved the best results with a depth of 4 and a width of 10. Thus, the optimal network is moderate in size. We also found that the DaPINN is insensitive to the network size, which indicates that the DaPINN can use a smaller network and achieve the same accuracy, thereby reducing computational costs.
\begin{figure}[htbp]
  \centering
  \includegraphics[width=100mm]{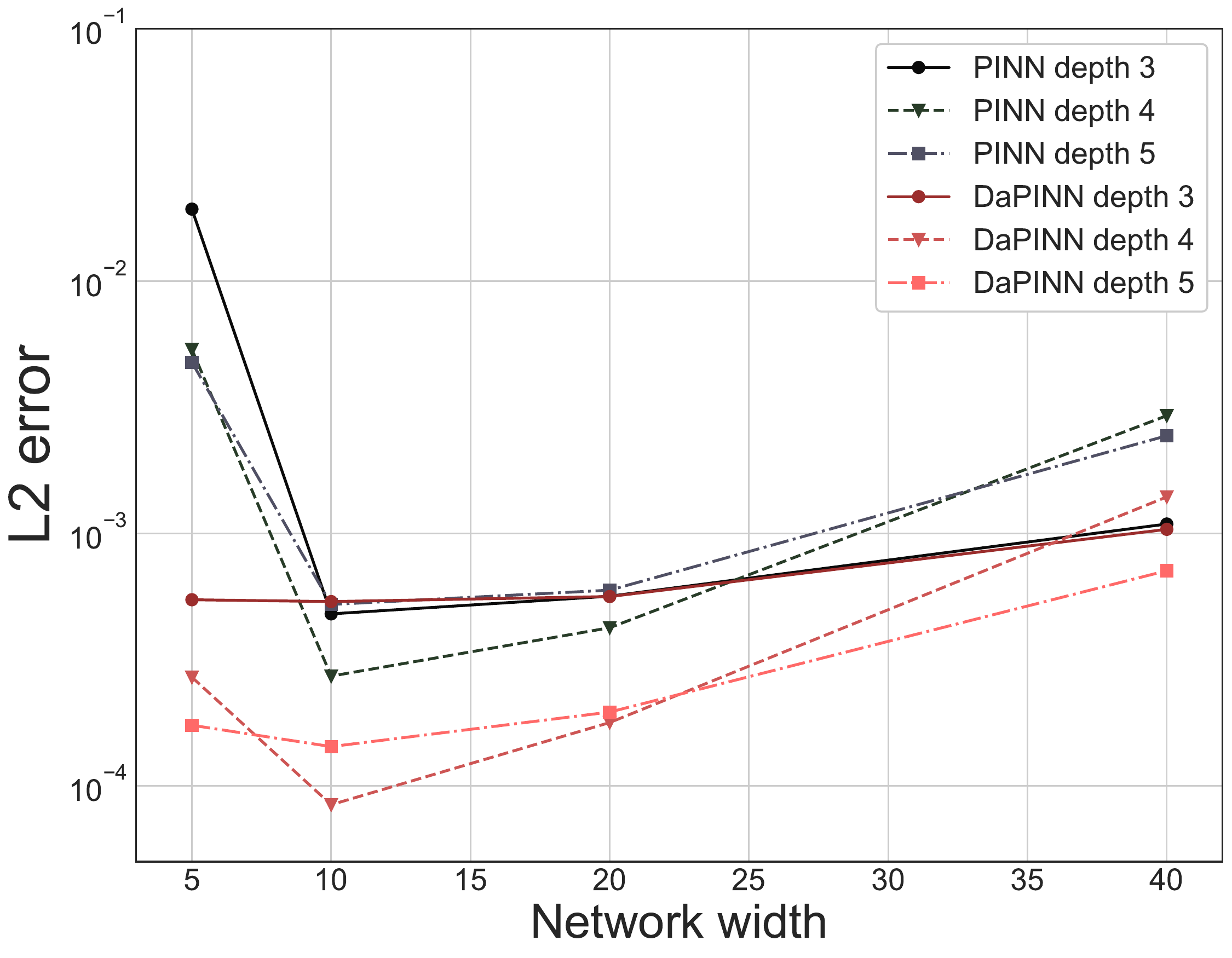}
  \caption{Problems in Section 3.4.1 with 400 sample points and 10,000 epochs: L2 relative errors of DaPINN and PINN models with different widths and depths.}
\end{figure}
\\
  
\subsection{Compatibility with other methods} 
The DaPINN modifies only the input layer and loss function of the network; thus, many PINN optimization methods are also applicable to the DaPINN.

We improved the accuracy and training efficiency of the DaPINN for solving PDEs by introducing the residual-based adaptive refinement (RAR) method, and we apply the RAR method to adaptively improve the distributions of the residuals during the training process. This method has a significant effect when solving problems with drastic changes in the function. Considering that the solution of the 1D Burgers' equation is very steep near $x = 0$ and that the error is concentrated in this region, the additive adaptive residual method is suitable in this problem.

In Section 3.3.2, we solved the problem using the PINN, DaPINN with $x^2$ and DaPINN with $x^3$ models. Here, we apply the RAR method to the PINN and DaPINN with $x^3$ models and compare the results. For the RAR method, we use 1500 training points in the first training epoch and add 20 residual points in each subsequent training epoch.

The results are shown as the green and yellow lines in Fig. 12(a). To reduce the L2 error to 0.5\%, the original PINN and DaPINN models need 3600 and 2400 points, respectively. With the RAR method, the PINN and DaPINN methods require only 1960 and 1660 points to achieve the same accuracy. In particular, the DaPINN uses only a quarter of the training points used by the PINN.

The solution to this problem has dramatic changes near $x = 0$ (Fig. 12(b)); thus, the PINN error is mainly concentrated near this area (Fig. 12(g)(i)). After adding 300 residual points, the maximum error in the PINN model decreases from 125\% to 75\%, while the DaPINN's maximum error decreases from 100\% to 2.5\%. In contrast, the DaPINN error (Fig. 12(h)) is more scattered than the PINN error (Fig. 12(j)).

\begin{figure}[htbp]
  \centering
  \subfigure[]{
  \begin{minipage}[t]{0.5\linewidth}
  \centering
  \includegraphics[width=70mm]{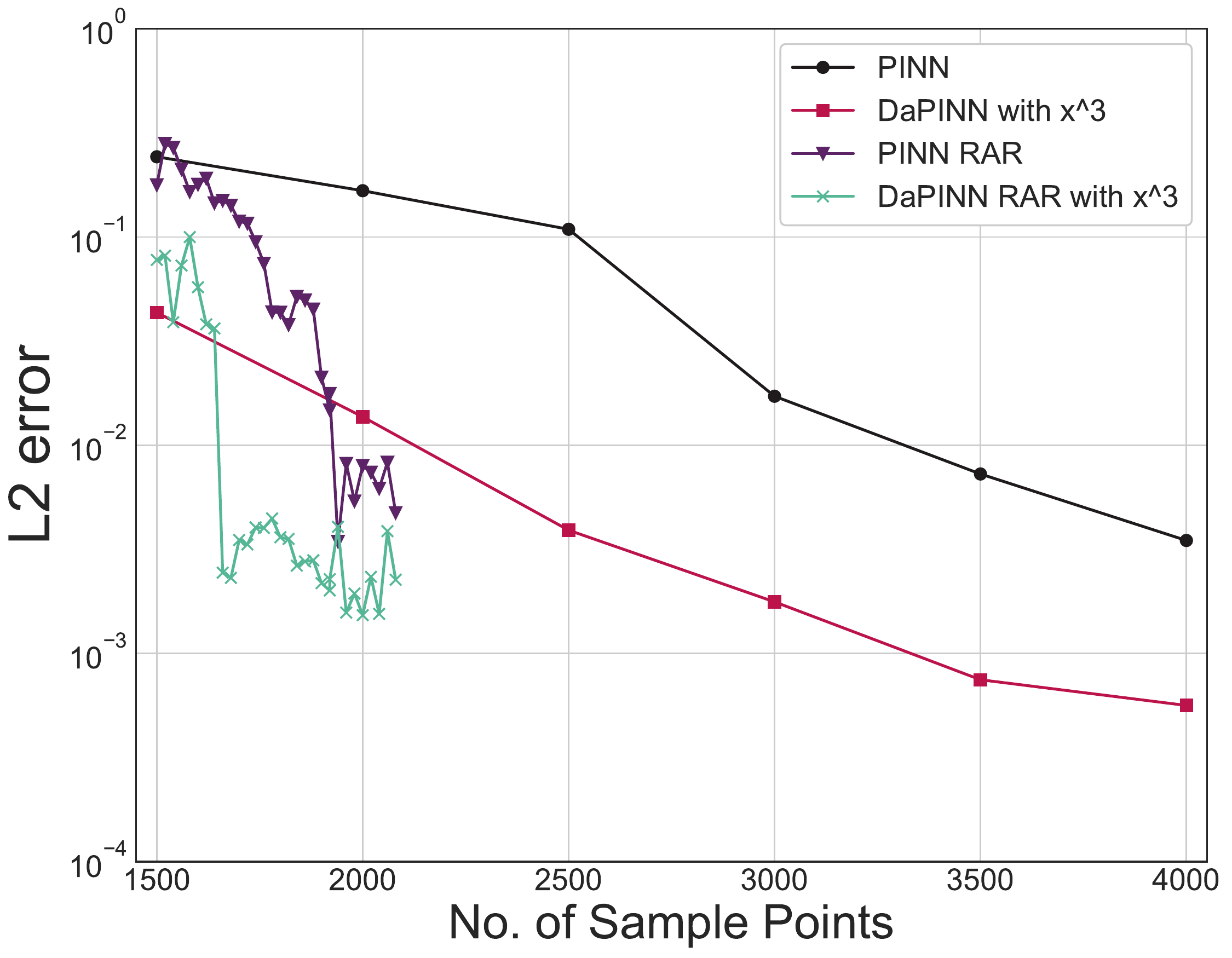}
  \end{minipage}%
  }%
  \subfigure[]{
  \begin{minipage}[t]{0.5\linewidth}
  \centering
  \includegraphics[width=70mm]{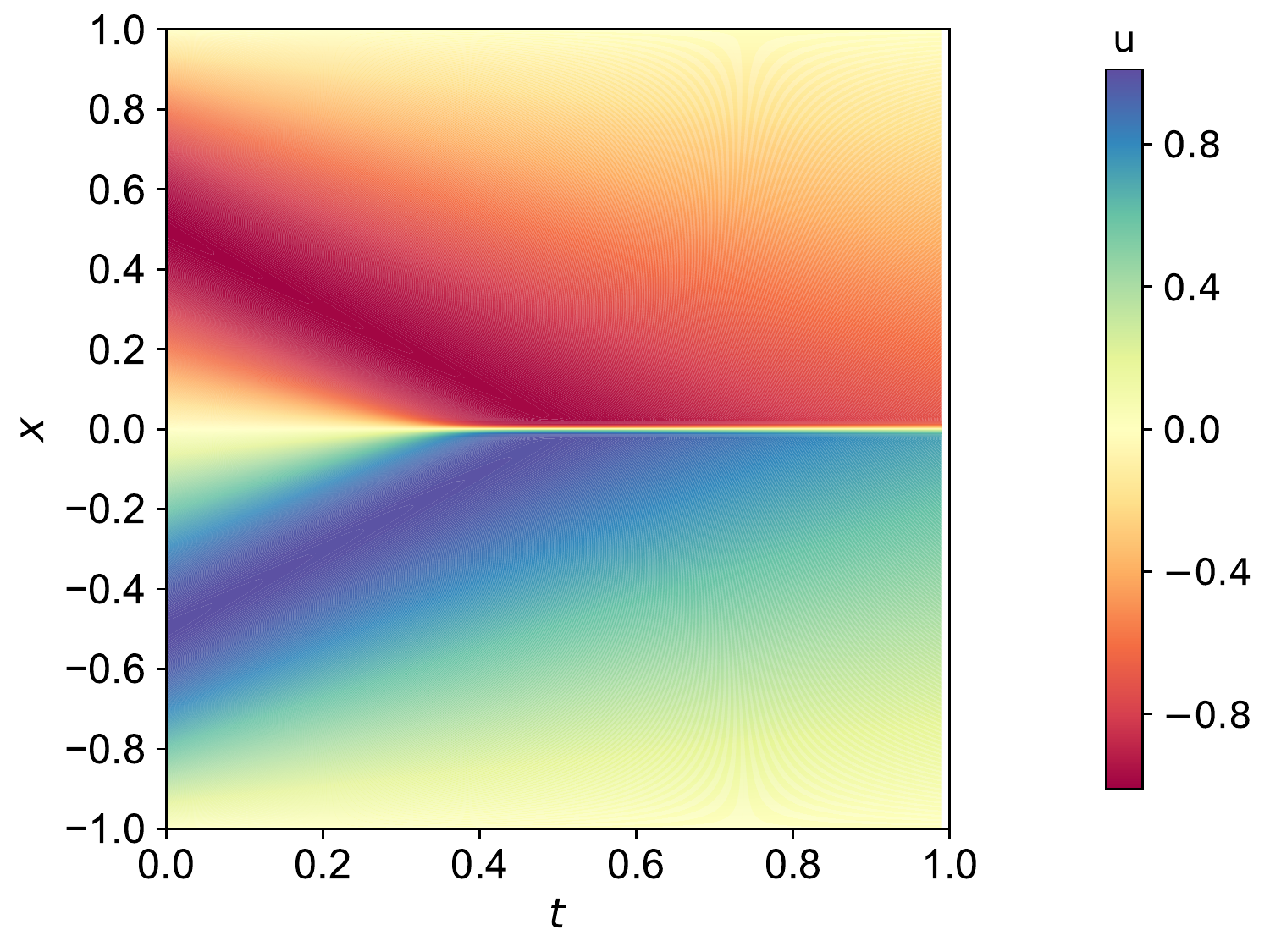}
  \end{minipage}%
  }%

  \subfigure[]{
  \begin{minipage}[t]{0.25\linewidth}
  \centering
  \includegraphics[width=40mm]{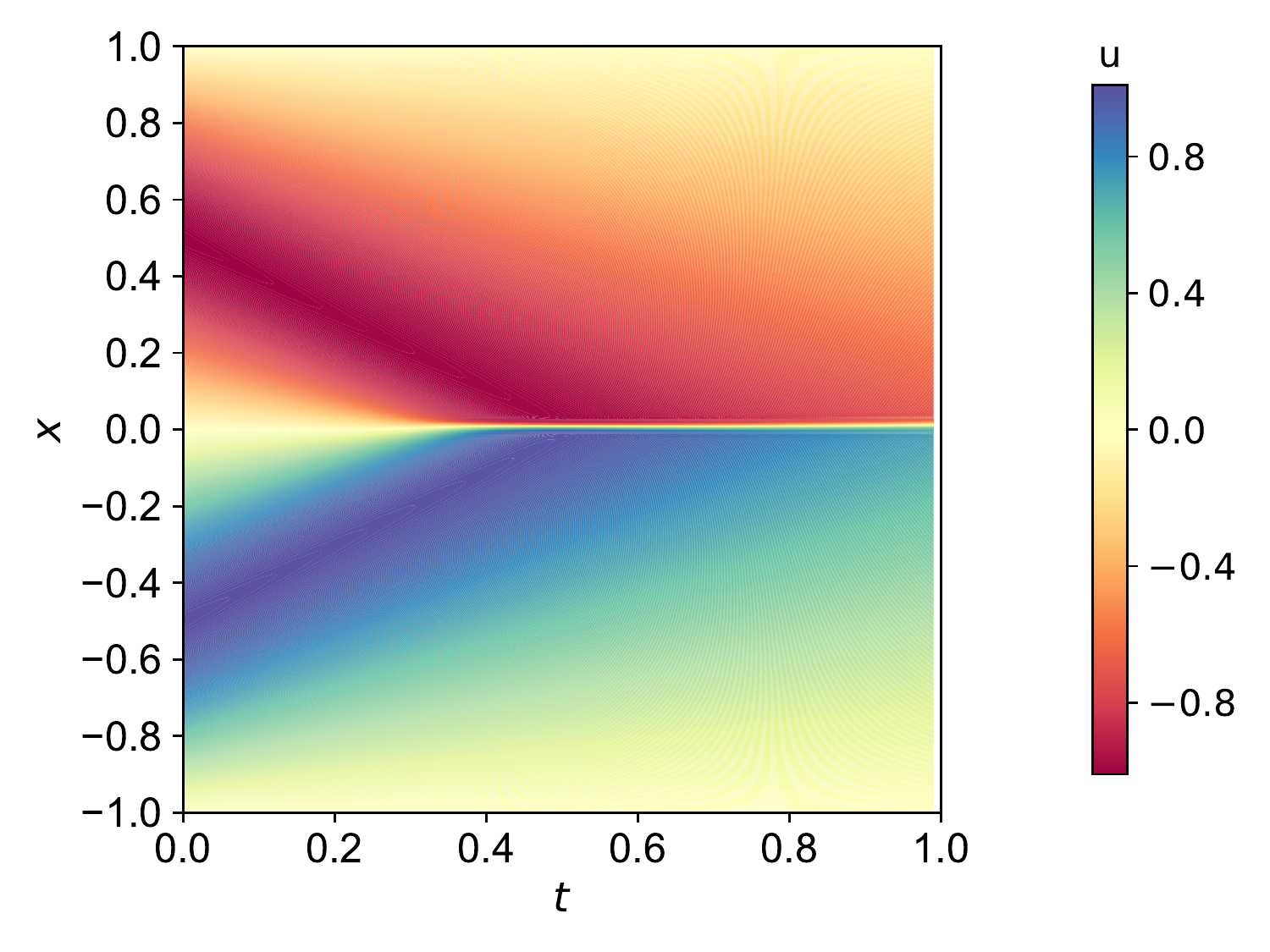}
  \end{minipage}%
  }%
  \subfigure[]{
  \begin{minipage}[t]{0.25\linewidth}
  \centering
  \includegraphics[width=40mm]{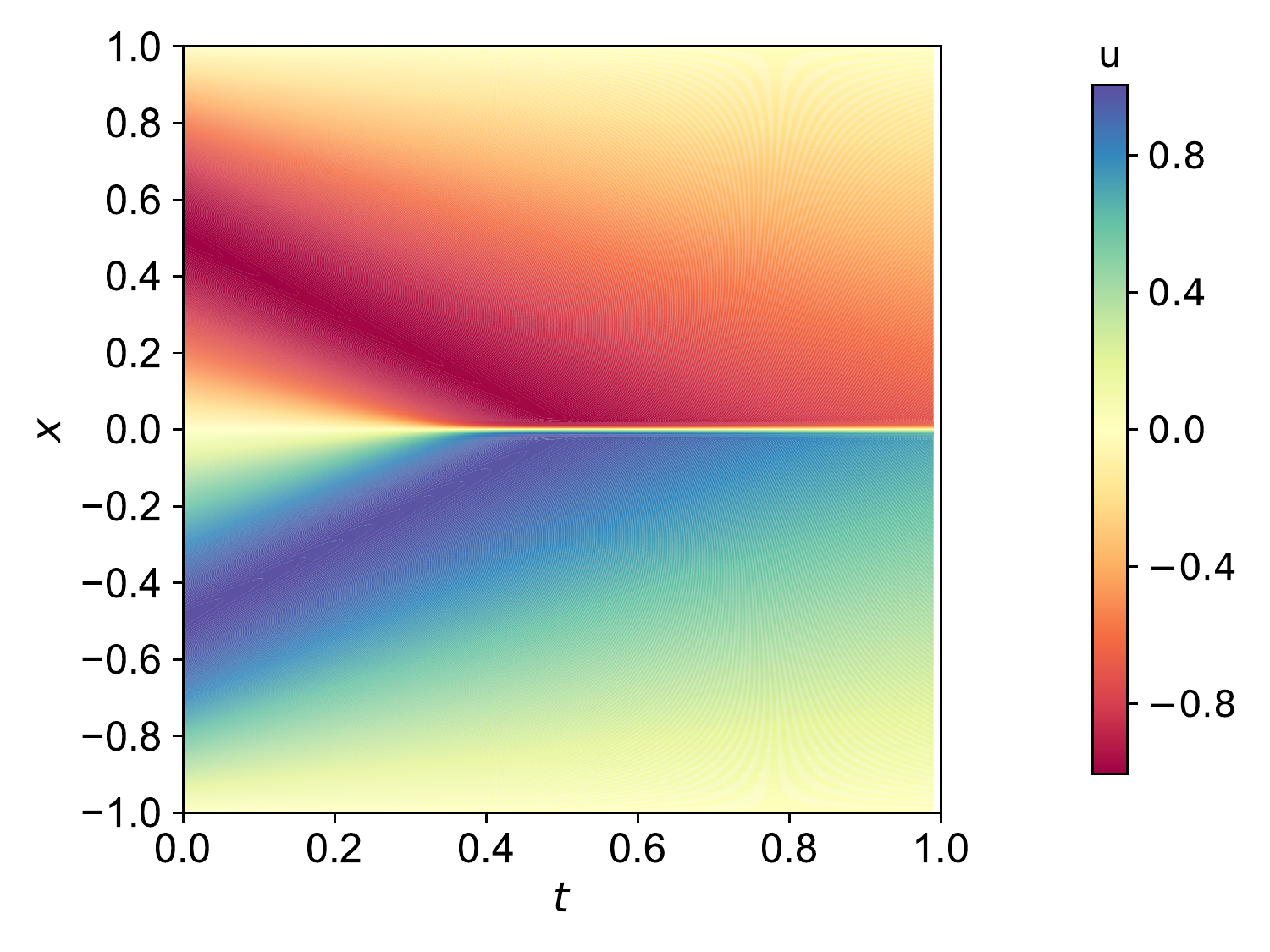}
  \end{minipage}%
  }%
  \subfigure[]{
  \begin{minipage}[t]{0.25\linewidth}
  \centering
  \includegraphics[width=40mm]{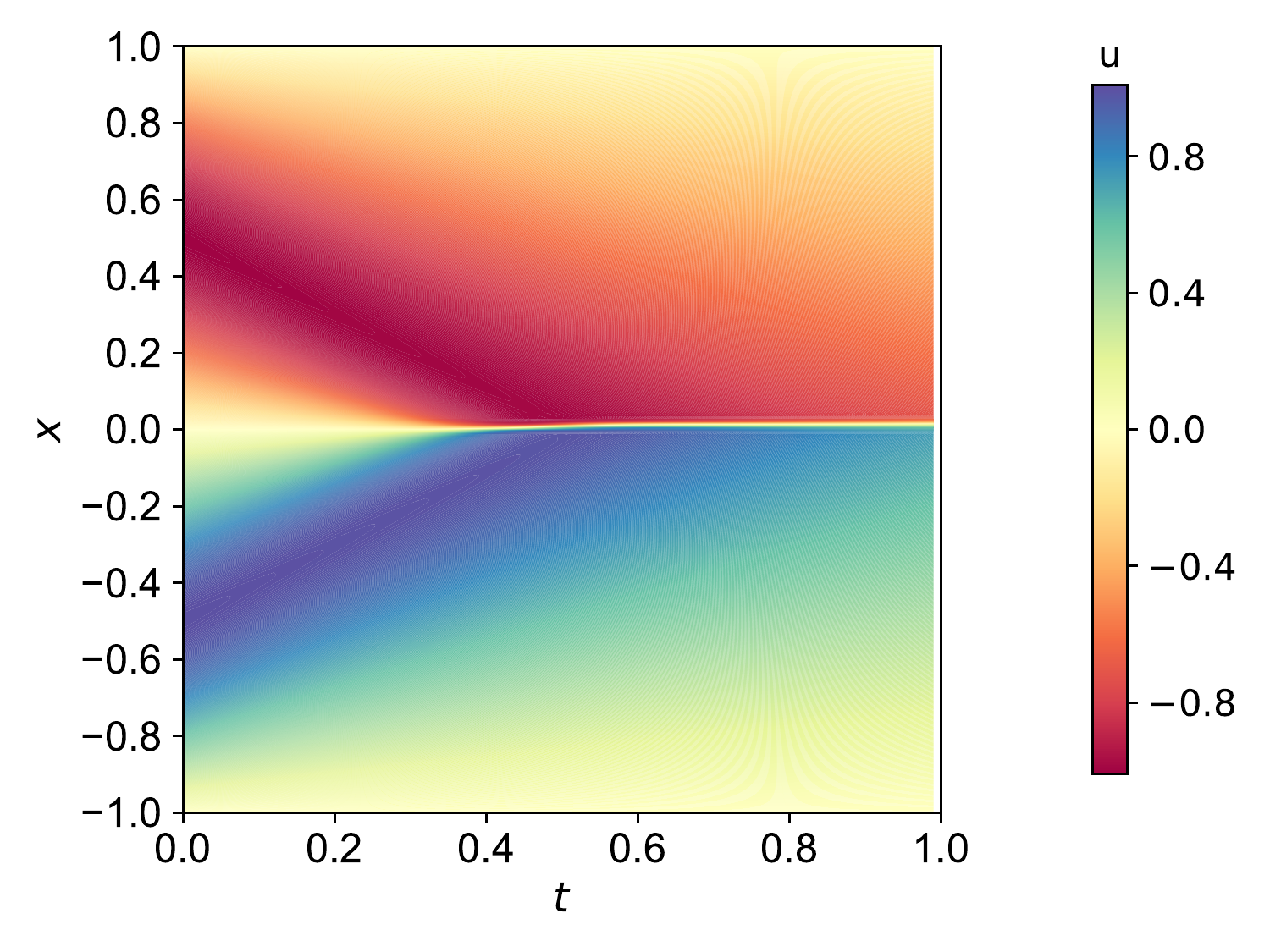}
  \end{minipage}
  }%
  \subfigure[]{
  \begin{minipage}[t]{0.25\linewidth}
  \centering
  \includegraphics[width=40mm]{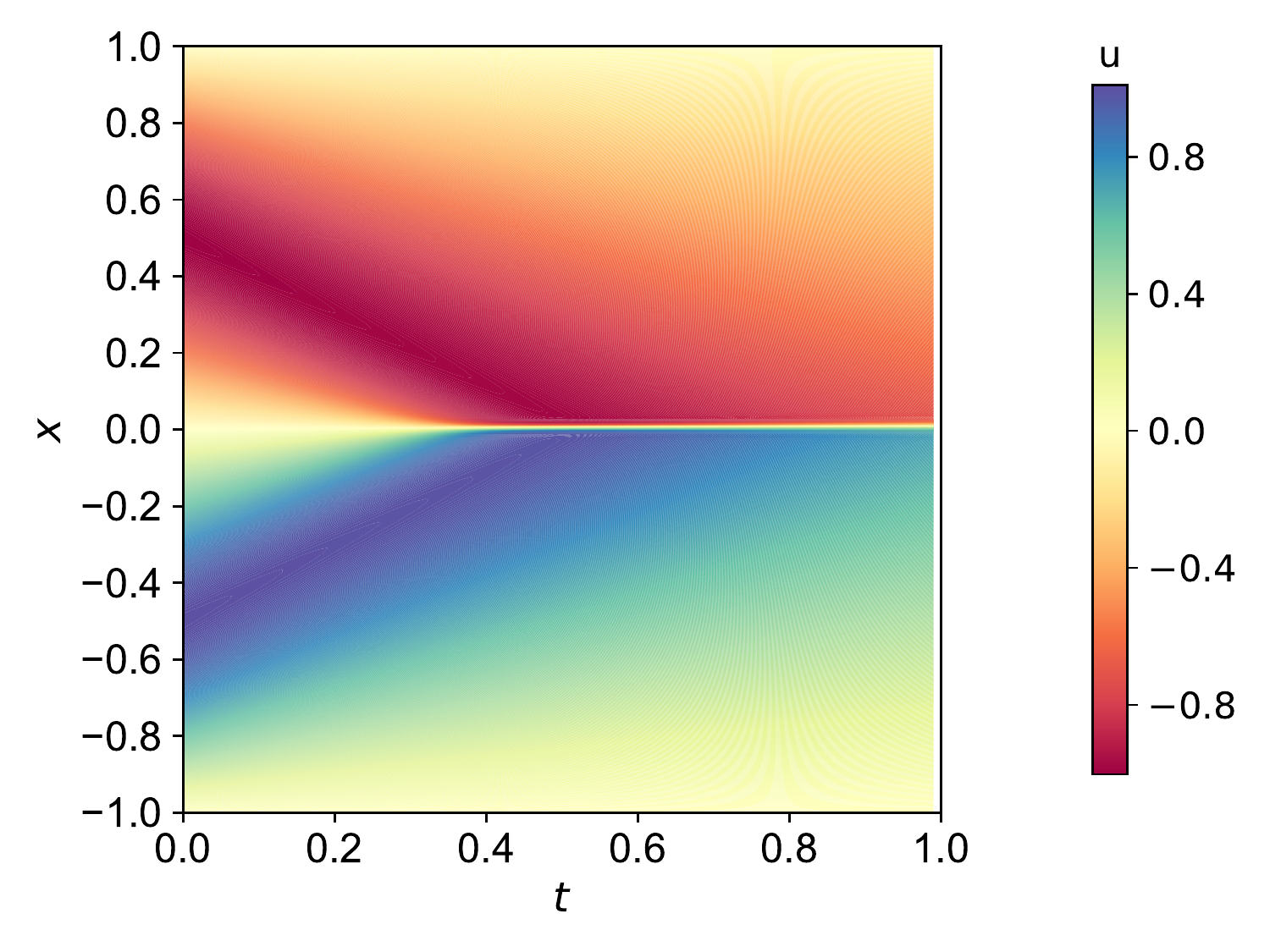}
  \end{minipage}
  }%

  \subfigure[]{
  \begin{minipage}[t]{0.25\linewidth}
  \centering
  \includegraphics[width=40mm]{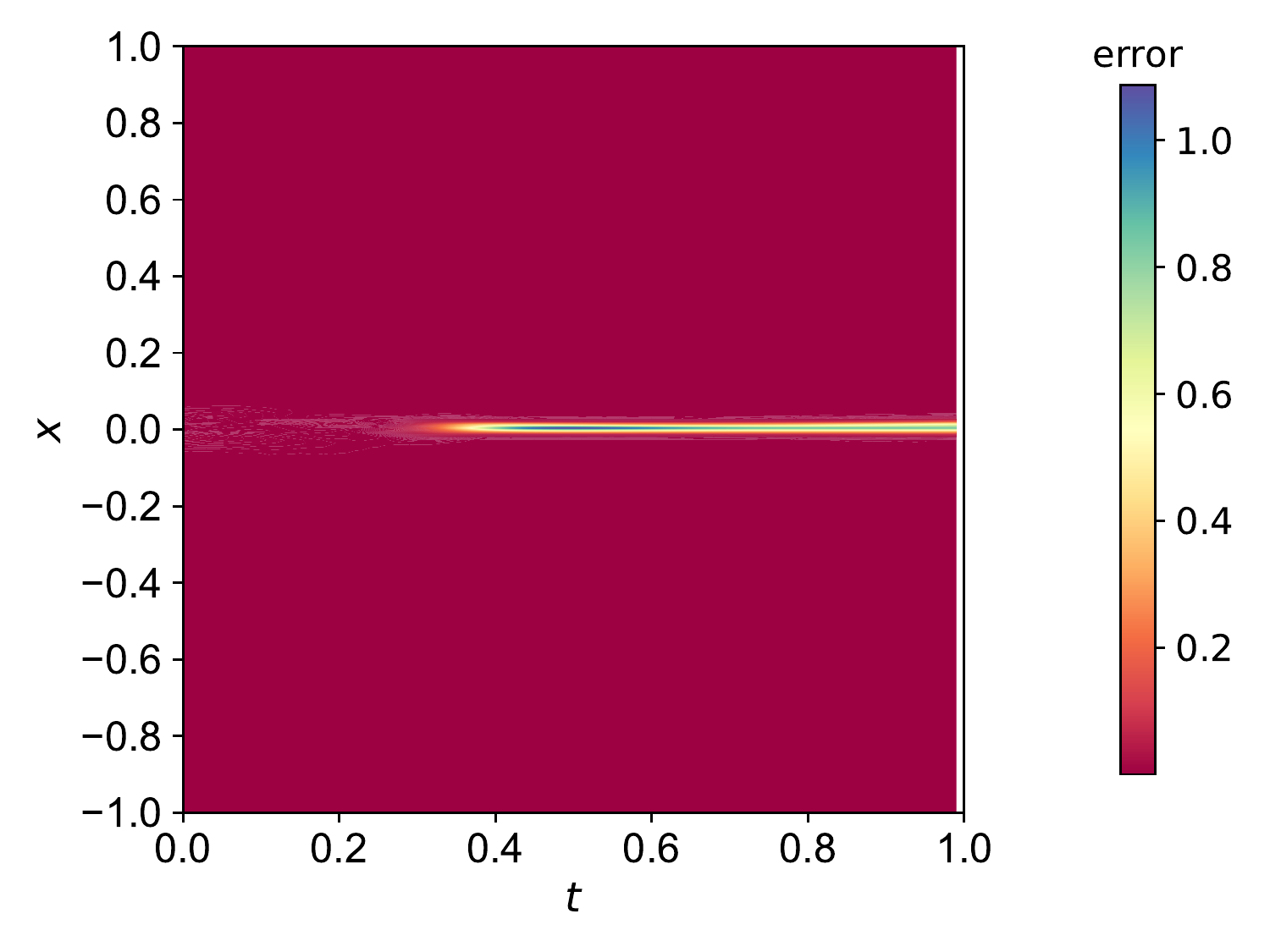}
  \end{minipage}%
  }%
  \subfigure[]{
  \begin{minipage}[t]{0.25\linewidth}
  \centering
  \includegraphics[width=40mm]{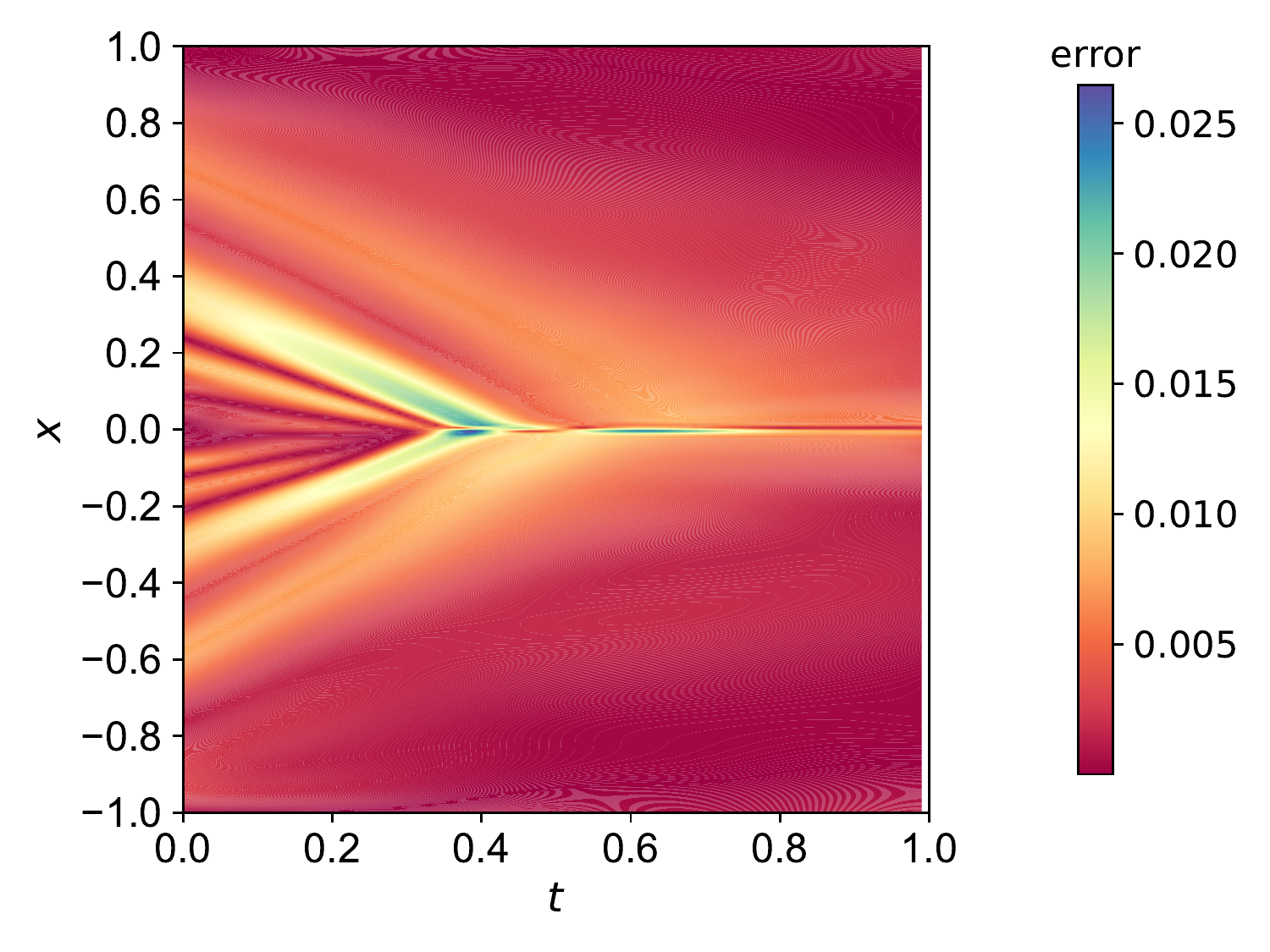}
  \end{minipage}%
  }%
  \subfigure[]{
  \begin{minipage}[t]{0.25\linewidth}
  \centering
  \includegraphics[width=40mm]{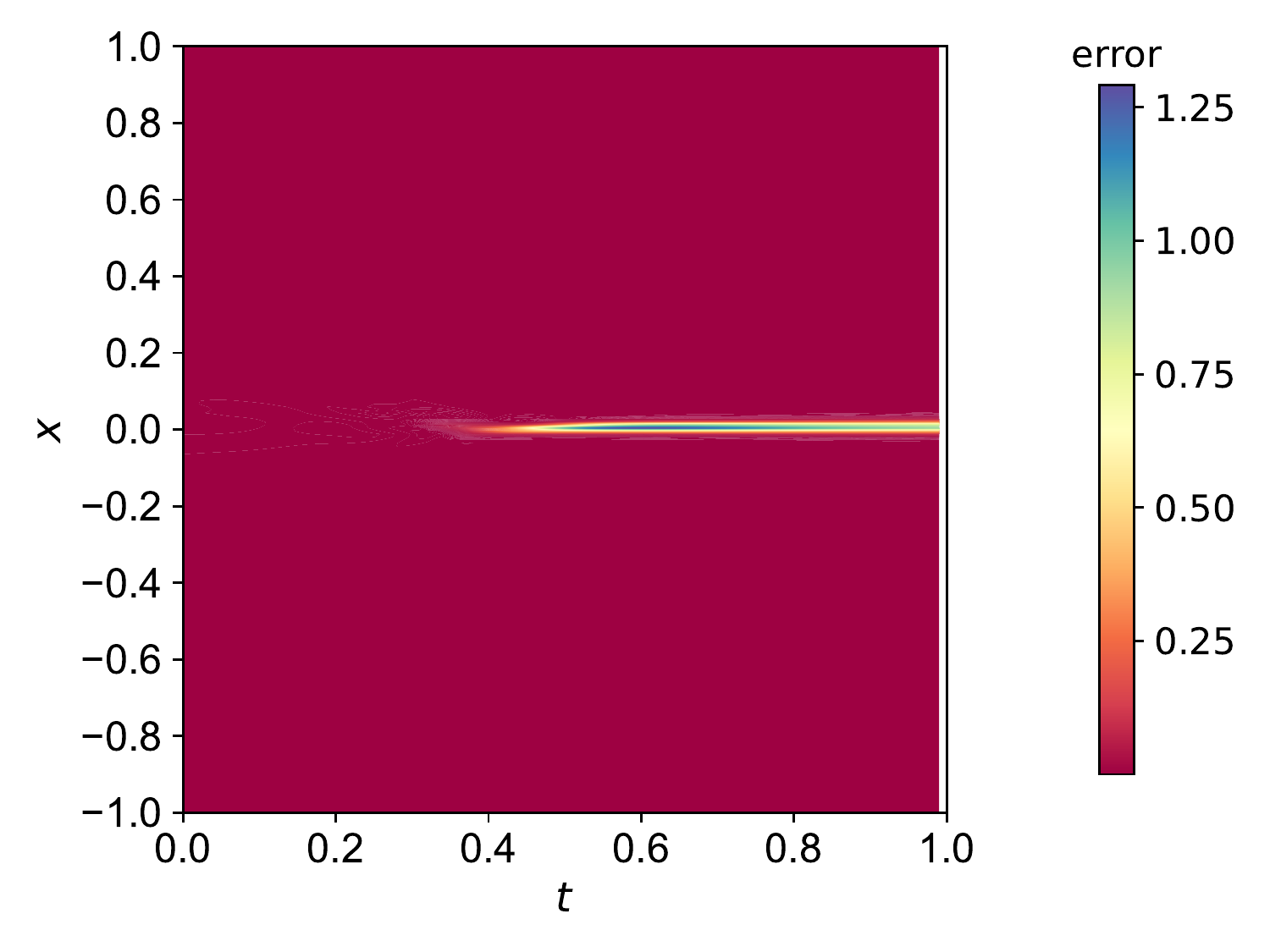}
  \end{minipage}
  }%
  \subfigure[]{
  \begin{minipage}[t]{0.25\linewidth}
  \centering
  \includegraphics[width=40mm]{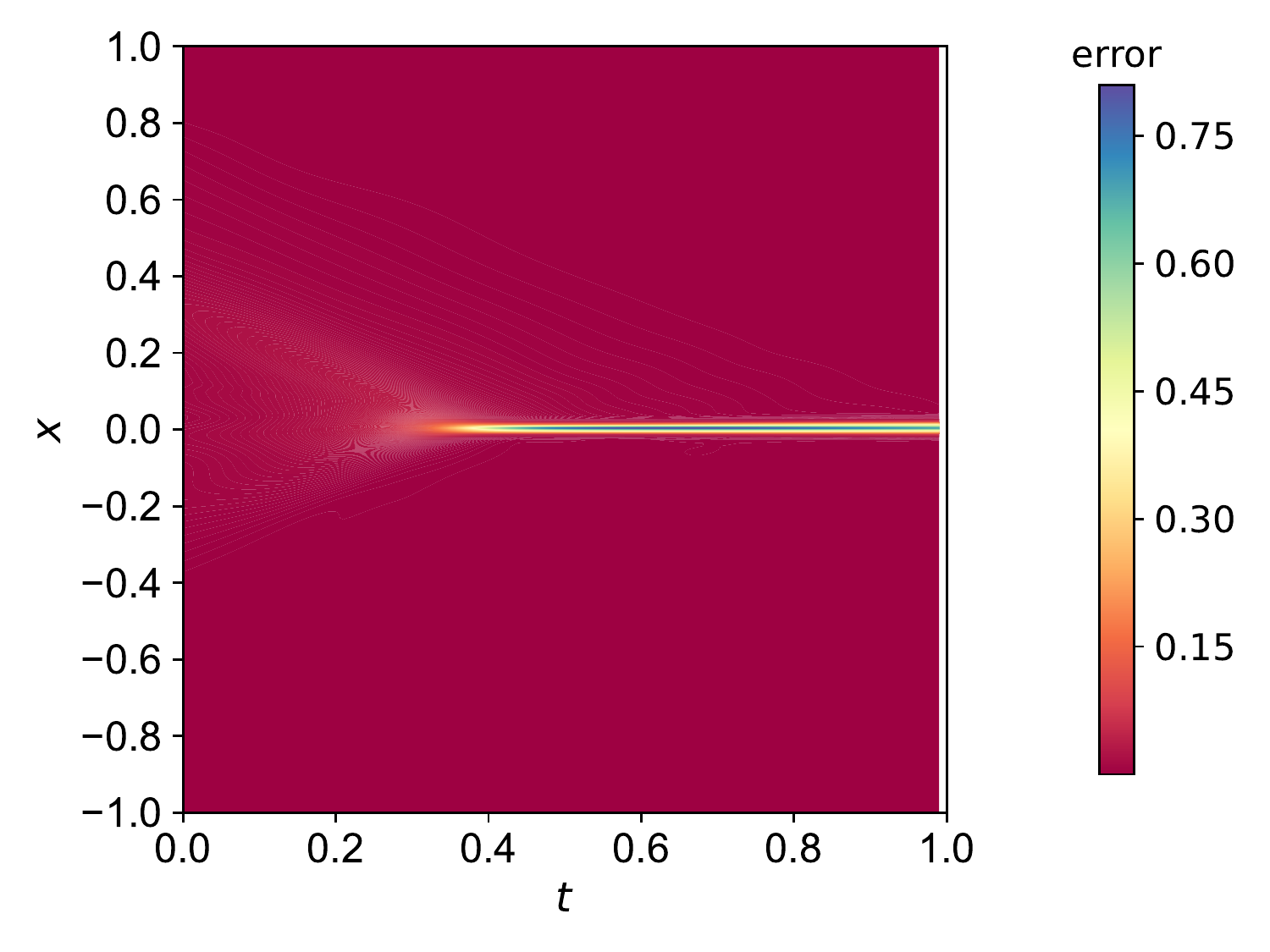}
  \end{minipage}
  }%
  \centering
  \caption{Example in Section 4.2: (a) L2 relative errors in the prediction functions u of the PINN and DaPINN models with respect to the number of samples. (b) Analytical solution results. (c) (g) DaPINN results before adding sample points and their absolute error. (d) (h) DaPINN results before adding 300 sample points using the RAR method. (e) (i) PINN results before adding sample points and their absolute error results. (f) (j) PINN results before adding 300 sample points using the RAR method and their absolute errors.}
  \end{figure}

When we combine the DaPINN with the RAR method, we obtain excellent results, showing that the DaPINN approach is versatile and can be used with other methods without conflicts to achieve better accuracy.

\section{Summary and Conclusion}
In this paper, we systematically propose the DaPINN, a new scheme for enhancing the original PINN by introducing an inductive bias such as an augmented input dimension to the neural network. Moreover, we demonstrate the effectiveness of our scheme, showing that the power series, Fourier series and replica augmentation methods dramatically improve the network performance. The results show that the DaPINN significantly outperforms the PINN for the same number of training points and epochs and that the DaPINN requires less time to achieve the same accuracy. We also prove that the introduction of higher-order terms continuously increases the DaPINN accuracy by using the power series augmentation approach. Furthermore, we discuss the effect of the network size on the DaPINN and conclude that while excessively large and small networks deteriorate network performance, smaller networks tend to have better results. In addition, we combine the DaPINN with the residual-based adaptive refinement (RAR) method to further improve model performance. We achieve excellent results with fewer sample points in solving the drastically varying Burgers' equation, demonstrating the compatibility of the DaPINN with other methods.

Although the results suggest that our proposed scheme is effective, detailed implementations do not work equally well for different scenarios. In general, the power series augmentation method is simple and effective and does not require the introduction of additional parameters, which are considerable advantages. However, the replica augmentation method can be used for problems with very few sample points because its mirror input feature reduces overfitting. For PDEs containing periodic functions (either in the equation or definite condition), better results will be obtained using Fourier series augmentation.
\section{Further Work}
The DaPINN can be implemented with other methods where further studies could be performed to reveal more general conclusions. Also in this paper, we do not test the performance of DaPINN for high-dimensional problems, which remains a meaningful area to discover.
\\
\\
\\
\\
\\
\\
\\
\\
\\
\bibliographystyle{unsrt}
\bibliography{A_Dimension-Augmented_Physics-Informed_Neural_Network}

\end{document}